\RequirePackage[2020-02-02]{latexrelease}
\documentclass[oneside,table,dvipsnames]{arxiv}

\usepackage{tikz}
\usepackage{xr-hyper} %
\usepackage{hyperref}
\usepackage{cleveref}

\newcommand{\newparagraph}[1]{\noindent\textbf{#1} }

\NewDocumentCommand{\codeword}{v}{%
\frenchspacing\texttt{\textcolor{gray}{#1}}%
}
\hypersetup{
    pdftitle={CEBRA: Learnable latent embeddings for joint behavioral and neural analysis},
    pdfauthor={Steffen Schneider, Jin H Lee, Mackenzie W Mathis},
    pdfsubject={CEBRA: Learnable latent embeddings for joint behavioral and neural analysis},
    colorlinks=true, allcolors=violet
}

\usepackage{graphicx}
\usepackage{epstopdf}
\usepackage{booktabs}
\usepackage{siunitx}
\usepackage{enumitem}
\usepackage{verbatim}
\usepackage{dblfloatfix}
\usepackage{algpseudocode}
\usepackage{bbding}
\usepackage{balance}

\usepackage{amsthm}
\usepackage{amsmath}
\usepackage{amsfonts}
\usepackage{amssymb} %
\usepackage{mathtools}

\DeclareMathOperator{\rank}{rank}
\renewcommand{\vec}[1]{\mathbf{#1}}
\renewcommand{\aa}{\vec{a}}
\newcommand{\bb}{\vec{b}}
\newcommand{\cc}{\vec{c}}
\newcommand{\dd}{\vec{d}}

\newcommand{\ff}{\vec{f}}
\renewcommand{\gg}{\vec{g}}
\newcommand{\hh}{\vec{h}}

\newcommand{\qq}{\vec{q}}

\renewcommand{\ss}{\vec{s}}
\newcommand{\uu}{\vec{u}}
\newcommand{\vv}{\vec{v}}

\newcommand{\xx}{\vec{x}}
\newcommand{\yy}{\vec{y}}
\newcommand{\zz}{\vec{z}}

\newcommand{\mat}[1]{\mathbf{#1}}
\renewcommand{\AA}{\mat{A}}
\newcommand{\BB}{\mat{B}}

\newcommand{\II}{\mat{I}}
\newcommand{\JJ}{\mat{J}}

\newcommand{\LL}{\mat{L}}
\newcommand{\MM}{\mat{M}}

\newcommand{\PP}{\mat{P}}
\newcommand{\QQ}{\mat{Q}}

\newcommand{\R}{\mathbb{R}}

\newcommand{\cX}{X}
\newcommand{\cY}{Y}
\newcommand{\cZ}{Z}
\newcommand{\cint}[1]{\int\limits_{\mathclap{#1}}}

\newcommand{\ggx}{\gg}
\newcommand{\ggy}{\gg'}
\newcommand{\ffx}{\ff}
\newcommand{\ffy}{\ff'}
\newcommand{\hhx}{\hh}
\newcommand{\hhy}{\hh'}

\newcommand{\tggx}{\tilde \gg}
\newcommand{\tggy}{\tilde \gg'}
\newcommand{\tffx}{\tilde \ff}
\newcommand{\tffy}{\tilde \ff'}
\newcommand{\thhx}{\tilde \hh}
\newcommand{\thhy}{\tilde \hh'}

\newcommand{\fx}{f}
\newcommand{\fy}{f'}

\newcommand{\hy}{h'}

\newcommand{\dimu}{d}
\newcommand{\dimv}{d'}
\newcommand{\dimx}{D}
\newcommand{\dimy}{D'}
\newcommand{\dime}{E}

\newcommand{\Ru}{\R^{\dimu}}
\newcommand{\Rv}{\R^{\dimv}}
\newcommand{\Rx}{\R^{\dimx}}
\newcommand{\Ry}{\R^{\dimy}}
\newcommand{\RE}{\R^{\dime}}

\newcommand{\Loss}[3]{{\mathcal{L}[#1]{#2}}_{#3}}
\newcommand{\Linf}[1][]{\Loss{\psi}{\detokenize{#1}}{\mathrm{asympt}}}
\newcommand{\Ln}[1][]{\Loss{\psi}{#1}{n}}
\newcommand{\Linfopt}[1][]{\Loss{\psi^{*}}{\detokenize{#1}}{\mathrm{asympt}}}
\newcommand{\DKL}{\mathcal{D}_{\mathrm{KL}}}

\newcommand{\bmu}{\boldsymbol{\mu}}
\newcommand{\suffstat}{\tilde{q}}
\newcommand{\marginal}{p_{\mathcal{D}}}

\newtheorem{defi}{Definition}
\newtheorem{prop}{Proposition}
\newtheorem{corol}{Corollary}

\newtheorem{innercustomgeneric}{\customgenericname}
\providecommand{\customgenericname}{}
\newcommand{\newcustomtheorem}[2]{%
  \newenvironment{#1}[1]
  {%
   \renewcommand\customgenericname{#2}%
   \renewcommand\theinnercustomgeneric{##1}%
   \innercustomgeneric
  }
  {\endinnercustomgeneric}
}
\newcustomtheorem{customdef}{Definition}
\newcustomtheorem{customprop}{Theorem}

\newcommand{\beginsupplement}{%
        \setcounter{table}{0}
        \renewcommand{\thetable}{S\arabic{table}}%
        \setcounter{figure}{0}
        \renewcommand{\thefigure}{S\arabic{figure}}%
     }

\definecolor{light-gray}{gray}{0.95}

\newcommand{\cebra}{CEBRA}

\usepackage[
    frozencache,
    cachedir=Listings
]{minted}
\setminted{fontsize=\footnotesize}
\usepackage[
    framemethod=TikZ,
    roundcorner=5pt,
    leftmargin =-1cm,
    middlelinewidth=0pt,
    backgroundcolor=gray!10,
]{mdframed}

\usepackage{nimbusmononarrow}
\usepackage{orcidlink}
\usepackage{fontawesome}
\usepackage{multicol}

\makeatletter
\BeforeBeginEnvironment{minted}{\begin{mdframed}}
\AfterEndEnvironment{minted}{\end{mdframed}}
\makeatother

\title{Learnable latent embeddings for joint behavioral and neural analysis}

\shorttitle{\textbf{\textcolor{Blue}{C}\textcolor{BlueViolet}{E}\textcolor{Plum}{B}\textcolor{RedViolet}{R}\textcolor{red}{A}}}

\author[1\,*\,\orcidlink{0000-0003-2327-6459}]{Steffen Schneider}
\author[1\,*\,\orcidlink{0000-0001-7018-1941}]{Jin Hwa Lee}
\author[1\,\Envelope\, \orcidlink{00000-0001-7368-4456}]{Mackenzie Weygandt Mathis}

\affil[1]{École Polytechnique Fédérale de Lausanne (EPFL), Brain Mind Institute \& Neuro-X. Geneva, Switzerland} 
\affil[$*$]{co-first authors, \Envelope  mackenzie@post.harvard.edu} 
\leadauthor{Schneider, Lee, Mathis}

\begin{document}

\onecolumn
\maketitle

\vspace{10pt}

\textbf{Mapping behavioral actions to neural activity is a fundamental goal of neuroscience. As our ability to record large neural and behavioral data increases, there is growing interest in modeling neural dynamics during adaptive behaviors to probe neural representations. In particular, neural latent embeddings can reveal underlying correlates of behavior, yet, we lack non-linear techniques that can explicitly and flexibly leverage joint behavior and neural data.
Here, we fill this gap with a novel method, CEBRA, that jointly uses behavioral and neural data in a hypothesis- or discovery-driven manner to produce consistent, high-performance latent spaces. We validate its accuracy and demonstrate our tool's utility for both calcium and electrophysiology datasets, across sensory and motor tasks, and in simple or complex behaviors across species. It allows for single and multi-session datasets to be leveraged for hypothesis testing or can be used label-free. Lastly, we show that CEBRA can be used for the mapping of space, uncovering complex kinematic features, and rapid, high-accuracy decoding of natural movies from visual cortex.}

\begin{multicols}{2}

 {\large A} central quest in neuroscience is the neural origin of behavior~\cite{Urai2022LargescaleNR,Krakauer2017NeuroscienceNB}. Yet, we are still limited in both the number of neurons and length of time we can record from behaving animals in a session. Therefore, we need new methods that can combine data across animals and sessions with minimal assumptions, and generate interpretable neural embedding spaces~\cite{Urai2022LargescaleNR, Jazayeri2021InterpretingNC}. Current tools for representation learning are either linear, or if non-linear they typically rely on generative models, and they do not yield consistent embeddings across animals (or repeated runs of the algorithm). Here, we combine recent advances in non-linear disentangled representation learning and self-supervised learning to develop a new dimensionality reduction method that can be applied jointly to behavioral and neural recordings to reveal meaningful lower dimensional neural population dynamics~\cite{humphries2021strong, Jazayeri2021InterpretingNC, zhou2020learning}. 
 \medskip
 
 From data visualization (clustering) to discovering latent spaces that explain neural variance, dimensionality reduction of behavior or neural data has been impactful in neuroscience. For example, complex 3D forelimb reaching can be reduced to only 8--12 dimensions~\cite{vargas2010decoding, Okorokova2020DecodingHK}, and the low dimensional embeddings reveal some robust aspects of movements (i.e., PCA-based manifolds where the neural state space can easily be constrained and is stable across time~\cite{Yu2008GaussianprocessFA, Churchland2012NeuralPD,Gallego2018CorticalPA}). Linear methods such as PCA are often used to increase interpretability, but this comes at the cost of performance~\cite{Urai2022LargescaleNR}. UMAP~\cite{mcinnes2018umap} and tSNE~\cite{van2009dimensionality} are excellent non-linear methods, but they lack the ability to explicitly use time information, which is always available in neural recordings, and they are not directly as interpretable as PCA.
 \medskip
 
 Non-linear methods are desirable to use for high performance decoding, but often lack identifiability: the desirable property that true model parameters can be determined, up to a known indeterminacy~\cite{Roeder2020, Hyvrinen2019NonlinearIU}. This is critical as it ensures that the learned representations are uniquely determined and thus facilitates consistency across animals and/or sessions. 
 \medskip
 
 There is recent evidence that label-guided VAEs could improve interpretability~\cite{zhou2020learning, Sani2020ModelingBR, klindt2021towards}. Namely, by using behavioral variables, such algorithms can learn to project future behavior onto past neural activity~\cite{Sani2020ModelingBR}, or explicitly use label-priors to shape the embedding~\cite{zhou2020learning}. However, these methods still have restrictive explicit assumptions on the underlying statistics of the data, and they do not guarantee consistent neural embeddings across animals~\cite{Pandarinath2018InferringSN, Prince2021ParallelIO, zhou2020learning}, which limits their generalizability as well as interpretability (and thereby affects accurate decoding across animals).
 \medskip

We address these open challenges with \textcolor{Blue}{C}\textcolor{BlueViolet}{E}\textcolor{Plum}{B}\textcolor{RedViolet}{R}\textcolor{red}{A}, a new self-supervised learning algorithm for obtaining interpretable, \textcolor{Blue}{\textbf{C}}onsistent \textcolor{BlueViolet}{\textbf{E}}m\textcolor{Plum}{\textbf{B}}eddings of high-dimensional \textcolor{RedViolet}{\textbf{R}}ecordings using \textcolor{red}{\textbf{A}}uxiliary variables.
 Our method combines ideas from non-linear independent component analysis (ICA) with contrastive learning~\cite{Gutmann12JMLR,oord2018representation,Hyvrinen2019NonlinearIU,khosla2020supervised}, a powerful self-supervised learning scheme, to generate latent embeddings conditioned on behavior (auxiliary variables) and/or time. CEBRA uses a novel data sampling scheme to train a neural network encoder with a contrastive optimization objective to shape the embedding space.
 It also can generate embeddings across multiple subjects, and cope with distribution shifts between experimental sessions, subjects, and recording modalities.
 Importantly, our method neither relies on data augmentation (as does SimCLR~\cite{Chen2020ASF}), nor on a specific generative model that would limit its range of use, and can be used in a hypothesis-driven (behaviorally-guided), data-driven (time-only), or hybrid manner. 
 We can uncover latent embeddings that are consistent and informative: we can decode behaviors with high accuracy, which we demonstrate for datasets from vision, sensorimotor, and memory systems.

\section*{Results}

    \begin{figure*}[t]
    \begin{center}
    \includegraphics[width=.97\textwidth]{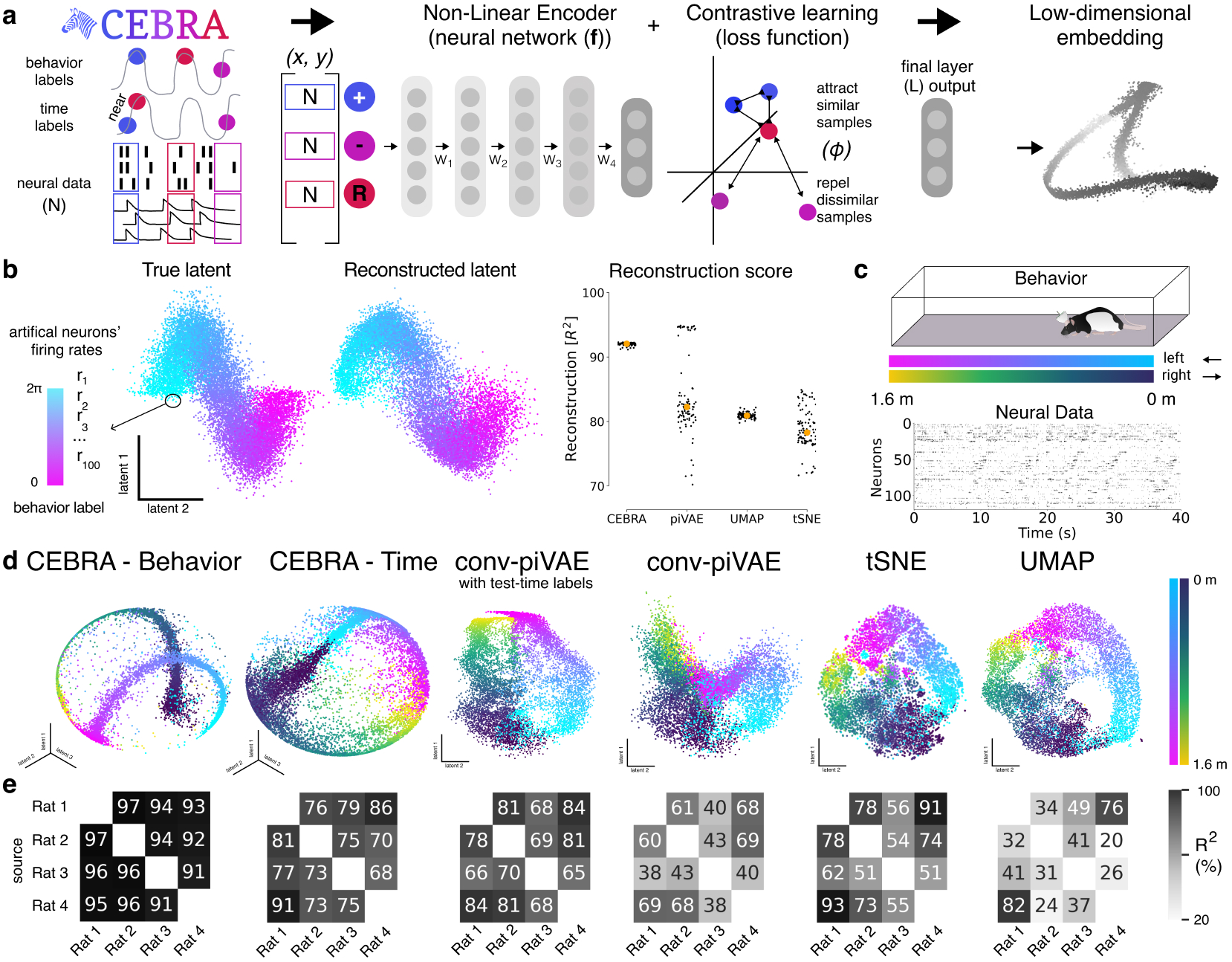}
    \end{center}
    \vspace{-5pt}
    \caption{{\bf \cebra \space for consistent and interpretable embeddings} {\bf (a)}: CEBRA allows for self-supervised, supervised, and hybrid approaches for hypothesis-driven and discovery-driven analysis. Overview of pipeline: collect data (e.g., pairs of behavior (or time) and neural data (x,y)), determine positive and negative pairs, train CEBRA, and produce embeddings.
    {\bf (b)}: Left: True 2D latent, where each point is mapped to spiking rate of 100 neurons. (Middle): \cebra \space embedding after linear regression to the true latent. Right: Reconstruction score is $R^2$ of linear regression between the true latent and resulting embedding from each method. The ``behavior label'' is a 1D random variable sampled from uniform distribution of [0, 2$\pi$] that is assigned to each time bin of synthetic neural data, visualized by the color map.
    The orange line is the median,and each black dot is an individual run (n=100). CEBRA-Behavior shows significantly higher reconstruction score compare to pi-VAE, tSNE and UMAP (one-way ANOVA, F(3, 396)=278.31, p=3.95e-97 with post hoc Tukey HSD p<0.001). 
    {\bf (c)}: Rat hippocampus data from~\cite{grosmark2016diversity}. Electrophysiology data collected during a task where the animal transverse a 1.6m linear track ``leftwards'' or ``rightwards''.
    {\bf (d)}: We benchmarked CEBRA against conv-pi-VAE (both with labels and without (self-supervised mode)), tSNE, and unsupervised UMAP. Note, for performance against the original pi-VAE see Extended Data Fig.~\ref{fig:CEBRAintroData}. We plot the 3 latents (note, all CEBRA embedding figures show the first 3 latents).%
    The dimensionality (D) of the latent space is set to the minimum and equivalent dimension per method (3D for CEBRA and 2D for others) to fairly compare. Note, higher dimensions for CEBRA can give higher consistency values (see Extended Data Fig.~\ref{fig:multiSession}).
    {\bf (e)}: Correlation matrices depict the $R^2$ after fitting a linear model between behavior-aligned embeddings of two animals, one as the target one as the source (mean, n=10 runs). Parameters were picked by optimizing average run consistency across rats. %
    }
    \label{fig:CEBRAvsAll}
    \end{figure*}

\subsection*{Joint behavioral and neural embeddings}

    \justify We propose a framework for jointly trained latent embeddings. %
    \cebra \space leverages user-defined labels (hypothesis-driven), or time-only labels (discovery-driven; Fig.~\ref{fig:CEBRAvsAll}a, Suppl. Note 1) to obtain consistent embeddings of neural activity that can be used for both visualization of data and downstream tasks like decoding.
    Specifically, it is an instantiation of non-linear ICA based on contrastive learning~\cite{Hyvrinen2019NonlinearIU}. Contrastive learning is a technique that leverages contrasting samples (positive and negative) against each other to find attributes in common and those that separate them. We can use discrete and continuous variables and/or time to shape the distribution of positive and negative pairs, and then use a non-linear encoder (here, a convolutional neural network (CNN), but can be another type of model) trained with a novel contrastive learning objective. The encoder features form a low-dimensional embedding of the data (Fig.~\ref{fig:CEBRAvsAll}a).
    Generating consistent embeddings is highly desirable and closely linked to identifiability in non-linear ICA~\cite{Hlv2021DisentanglingIF}. Theoretical work has shown that using contrastive learning with auxiliary variables is identifiable for bijective neural networks using a noise contrastive estimation (NCE) loss~\cite{Hyvrinen2019NonlinearIU}, and that with an InfoNCE loss this bijectivity assumption can sometimes be removed~\cite{Zimmermann2021ContrastiveLI} (see also Suppl. Note 2). InfoNCE minimization can be viewed as a classification problem where given a reference sample, the correct positive pair needs to be distinguished from multiple negative pairs.
    \medskip

    CEBRA optimizes a neural network $\ff$ that maps neural activity to an embedding space of a defined dimension (Fig.~\ref{fig:CEBRAvsAll}a).
    Pairs of data $(\xx, \yy)$ are mapped to this embedding space, and then compared with a similarity measure $\phi(\cdot, \cdot)$. Abbreviating this process with $\psi(\xx, \yy) = \phi(\ff(\xx), \ff(\yy)) / \tau$ with a temperature hyperparameter $\tau$, the full criterion to optimize is
    \begin{equation*}
      \mathop{\mathbb{E}}\limits_{{\substack{
            \xx \sim p(\xx),\ 
            \yy_+ \sim p(\yy|\xx)\\
            \yy_1,\dots,\yy_n \sim q(\yy|\xx)\\
        }}}
        \left[ - \psi(\xx, \yy_+) + \log \sum_{i=1}^{n} e^{\psi(\xx, \yy_i)}
        \right],
    \end{equation*}
    which, depending on the dataset size, can be optimized with algorithms for either batch or stochastic gradient descent.
    \medskip 
    
    In contrast to other contrastive learning algorithms, the positive pair distribution $p$ and the negative pair distribution $q$ can be systematically designed and allows the use of time, behavior, and other auxiliary information to shape the geometry of the embedding space. If only discrete labels are used, this training scheme is conceptually similar to supervised contrastive learning~\cite{khosla2020supervised}.
    \medskip
     
    \cebra \space can leverage continuous behavioral (kinematics, actions) as well as other discrete variables (trial ID, rewards, brain-area ID, etc.). Additionally, user-defined information about desired invariances in the embedding is used (across animals, sessions, etc.), allowing flexible ways of analyzing data. We group this information into task-irrelevant and task-relevant variables, and these can be leveraged in different contexts. For example, to investigate trial-to-trial variability or learning across trials, information like a trial ID would be considered a task-relevant variable. On the contrary, if we aim to build a robust brain machine interface that should be invariant to such short-term changes, we would include trial information as a task-irrelevant variable and obtain an embedding space which no longer carries this information.
    Crucially, this allows for inferring latent embeddings without explicitly modeling the data generating process (as done in pi-VAE~\cite{zhou2020learning} and LFADS~\cite{Pandarinath2018InferringSN}). Omitting the generative model and replacing it by a contrastive learning algorithm facilitates broader applicability without modifications.

    \begin{figure*}[ht!]
    \begin{center}
    \includegraphics[width=\textwidth]{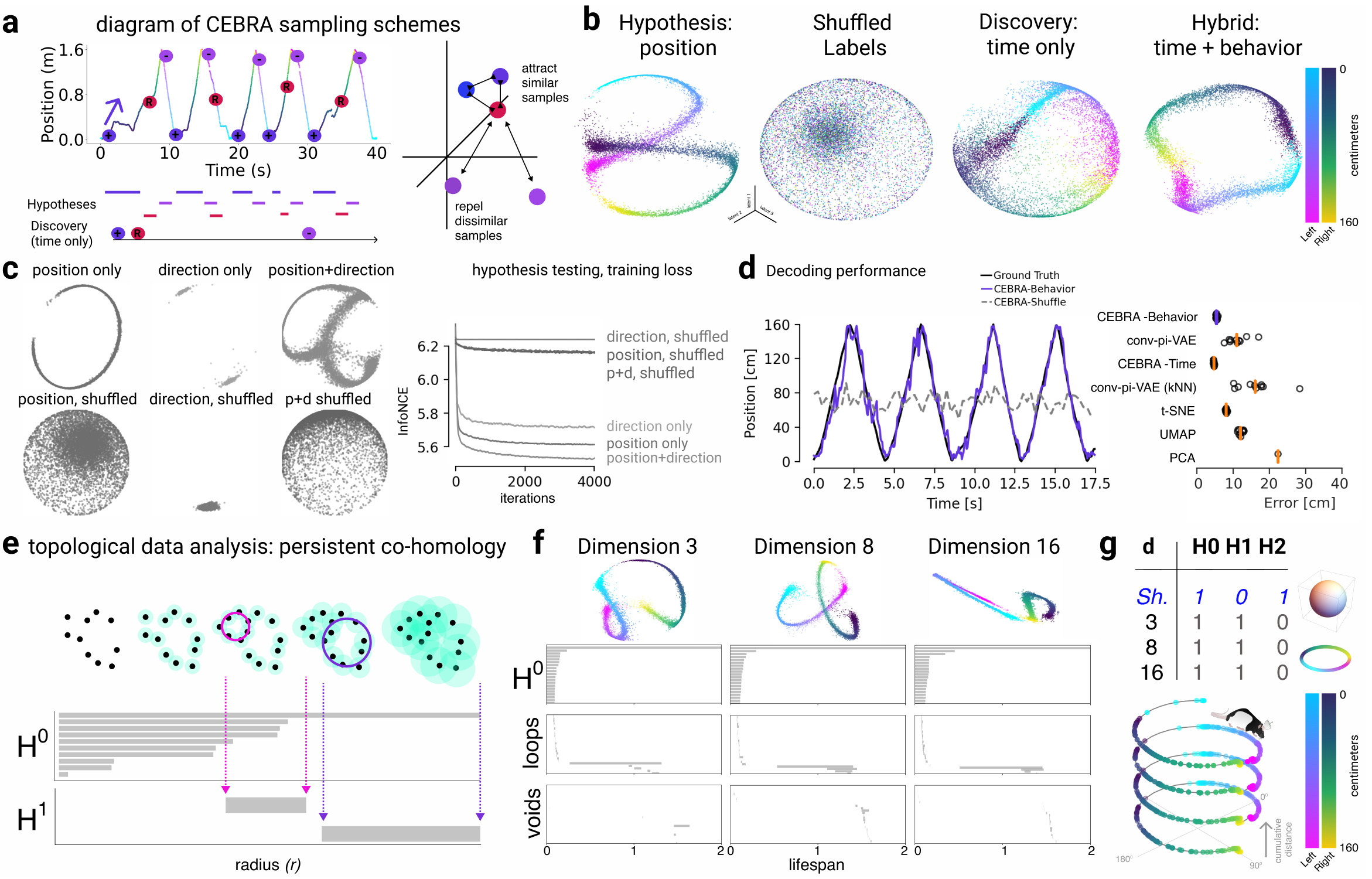}
    \end{center}
    \vspace{-10pt}
    \caption{{\bf Hypothesis-driven and discovery-driven analysis with CEBRA} {\bf (a)}: CEBRA can be used in three modes: hypothesis-driven, discovery-driven, or in a hybrid mode, which allows for weaker priors on the latent embedding. 
    {\bf (b)}: CEBRA with position-hypothesis derived embedding, shuffled (erroneous), time-only, and Time+Behavior (hybrid; here, a 5D space was used, where first 3D is guided by both behavior+time, and last 2D is guided only by time, and the first 3 latents are plotted). 
    {\bf (c)}: Embeddings with position-only, direction-only, and shuffled position-only, direction-only for hypothesis testing. The loss function can be used as a metric for embedding quality.
    {\bf (d)}: We utilized the hypothesis-driven (position+direction) or the shuffle (erroneous) to decode the position of the rat, which produces a large difference in decoding performance: position+direction $R^2$ is 73.35\% vs. -49.90\% shuffled and median absolute error 5.8 cm vs 44.7 cm.  Purple line is decoding from the 32-dimensional hypothesis-based latent space, dashed line is shuffled. Right is the performance across additional methods (The orange line indicates the median of the individual runs (n=10) that are indicated by black circles. Each run is averaged over 3 splits of the dataset). 
    {\bf (e)}: Schematic of how persistent co-homology is computed. Each data point is thickened to a ball of gradually expanding radius $r$, while tracking birth and death of ``cycles'' in each dimension ($H^0$ counts number of connected components or 0-dim cycles, $H^1$ counts the number of loops (1-dim cycles), $H^2$ counts the number of voids (2-dim cycles)). The prominent lifespans, indicated as pink and purple arrows, are considered to determine Betti numbers.
    {\bf (f)}: Visualization of the neural embeddings computed with different input dimensions, and the related persistent co-homology lifespan diagrams below.
    {\bf (g)}: Betti numbers from shuffled embeddings (Sh.) and across increasing dimensions (d) of CEBRA, and the topology preserving circular coordinates using the first co-cycle from persistent co-homology analysis (see Methods).}
    \label{fig:Topology}
    \end{figure*}

\subsection*{Robust and decodable latent embeddings}

    \justify We first demonstrate that CEBRA significantly outperforms tSNE, UMAP, and pi-VAE (the latter was shown to outperform PCA, LFADS, demixed-PCA, and pfDS~\cite{zhou2020learning}) at reconstructing ground truth synthetic data (one-way ANOVA, F(3, 396)=278.31, p=3.95e-97; Fig.~\ref{fig:CEBRAvsAll}b, Extended Data Fig.~\ref{fig:CEBRAintroData}a, b).
    \medskip
    
    We then turned to a hippocampus dataset that was used to benchmark neural embedding algorithms~\cite{grosmark2016diversity, zhou2020learning} (Extended Data Fig.~\ref{fig:CEBRAintroData}c, Suppl. Note 1). To note, we first significantly improved pi-VAE by adding a CNN thereby allowing this model to leverage multiple time steps, and used this for further benchmarking (Extended Data Fig.~\ref{fig:CEBRAintroData}d-e).
    To test our methods, we first consider the correlation of the resulting embedding space across subjects (does it produce similar latent spaces?), and the correlation across repeated runs of the algorithm (how consistent are the results?).
    We found that \cebra \space significantly outperformed other algorithms at producing consistent embeddings, and it produced visually informative embeddings (Fig.~\ref{fig:CEBRAvsAll}c-e, Extended Data Figs.~\ref{fig:CEBRAvsAlltemp}, ~\ref{fig:CEBRAvsAllSUPPL}; for each embedding a single point represents the neural population activity over a specified time bin). 
    \medskip 
    
    When using \cebra-Behavior the correlation of the resulting embedding space across subjects is significantly higher compared to conv-pi-VAE with, or without test-time labels (one-way ANOVA F(5, 67)=28, p=\num{3.4e-15}, Table~\ref{tab:consistency}; Fig.~\ref{fig:CEBRAvsAll}d, e)---note, \cebra \space does not require test-time labels. 
    Qualitatively, it can be appreciated that both \cebra-Behavior and -Time have similar output embeddings, while the latents from conv-pi-VAE with label priors or without labels are not consistent: namely, conv-pi-VAE without label priors results in a more entangled latent, suggesting that the label prior strongly shapes the output embedding structure of conv-pi-VAE.
    We also considered correlations across repeated runs of the algorithm and found higher consistency and lower variability with CEBRA (Fig.~\ref{fig:CEBRAvsAll}b, Extended Data Fig.~\ref{fig:CEBRAvsAllSUPPLrats}).

\subsection*{Hypothesis- and discovery-driven analyses}

    \justify One of the advantages of \cebra \space is its flexibility, limited assumptions, and ability to test hypotheses. For the hippocampus, one can hypothesize that these neurons represent space~\cite{Huxter2003IndependentRA, Moser2008PlaceCG}, and therefore the behavioral label could be position, or velocity (Figure~\ref{fig:Topology}a). Conversely, for the sake of argument, we could have an alternative hypothesis; i.e., hippocampus does not map space, just the direction of travel, or some other feature. Using the same model, but hypothesis-free and using time for selecting the contrastive pairs is also possible, and/or a hybrid thereof (Fig.~\ref{fig:Topology}a). 
    \medskip
    
    We trained hypothesis-guided, time-only, or hybrid models across a range of input dimensions and embedded the neural latents into a 3D space for visualization. Qualitatively, we find that position-based model produces a highly smooth embedding that reveals the position of the animal---namely, there is a continuous ``loop'' of neural latent activity around the track (Fig.~\ref{fig:Topology}b). 
    This is consistent with what is known about the hippocampus~\cite{grosmark2016diversity} and in particular reveals the topology of the linear track with direction specificity.  Whereas shuffling the labels, which breaks the correlation between neural activity and direction and position, produces an unstructured embedding (Fig.~\ref{fig:Topology}b).
    \medskip
    
    \cebra-Time produces an embedding that more closely resembles that of position (Fig.~\ref{fig:Topology}b). This also suggests that time contrastive learning captured the major latent space structure, independent of any label input, reinforcing that \cebra \space can serve both discovery and hypothesis-driven questions (and running both variants can be informative). The hybrid design, whose goal is to disentangle the latent to subspaces that are relevant to the given behavioral and the residual temporal variance and noise, showed a similarly structured embedding space as behavior (Fig.~\ref{fig:Topology}b).
     \medskip
    
    To quantify how \cebra \space can disentangle which variable had the largest influence on the embedding, we tested for encoding position, direction, and combinations thereof (Fig.~\ref{fig:Topology}c). We find that position plus direction is the most informative label~\cite{Dombeck2010FunctionalIO} (Fig.~\ref{fig:Topology}c, and Extended Data Fig.~\ref{fig:SFigure3_hypoTesting}a-d). This is evident in the embedding and the value of the loss function upon convergence, which serves as an additional ``goodness of fit'' metric to select the best labels; i.e., which label(s) produce the lowest loss at the same point in training (Extended Data Fig.~\ref{fig:SFigure3_hypoTesting}e). Note that erroneous (shuffled) labels converge to considerably higher loss values.
    \medskip
    
    To measure performance we consider how well can we decode behavior from the embeddings. As an additional baseline we performed linear dimensionality reduction with PCA. We used a k-nearest-neighbor (kNN) decoder for position and direction and measured the reconstruction error. We find \cebra-Behavior has significantly better decoding performance (Fig.~\ref{fig:Topology}d, and Suppl. Video 1), compared to pi-VAE and our conv-pi-VAE (one-way ANOVA F=131, p=\num{3.6e-24}), and \cebra-Time compared to unsupervised methods (tSNE, UMAP, PCA; one-way ANOVA F=\num{12091} p=\num{6.95e-42}; see also Table~\ref{tab:decoding}). \citet{zhou2020learning} reported a median absolute decoding error of 12 cm error, while we can achieve $\approx$5 cm (Fig.~\ref{fig:Topology}d). CEBRA therefore allows for high performance decoding while ensuring consistent embeddings.

\begin{figure*}[t]
\begin{center}
\includegraphics[width=\textwidth]{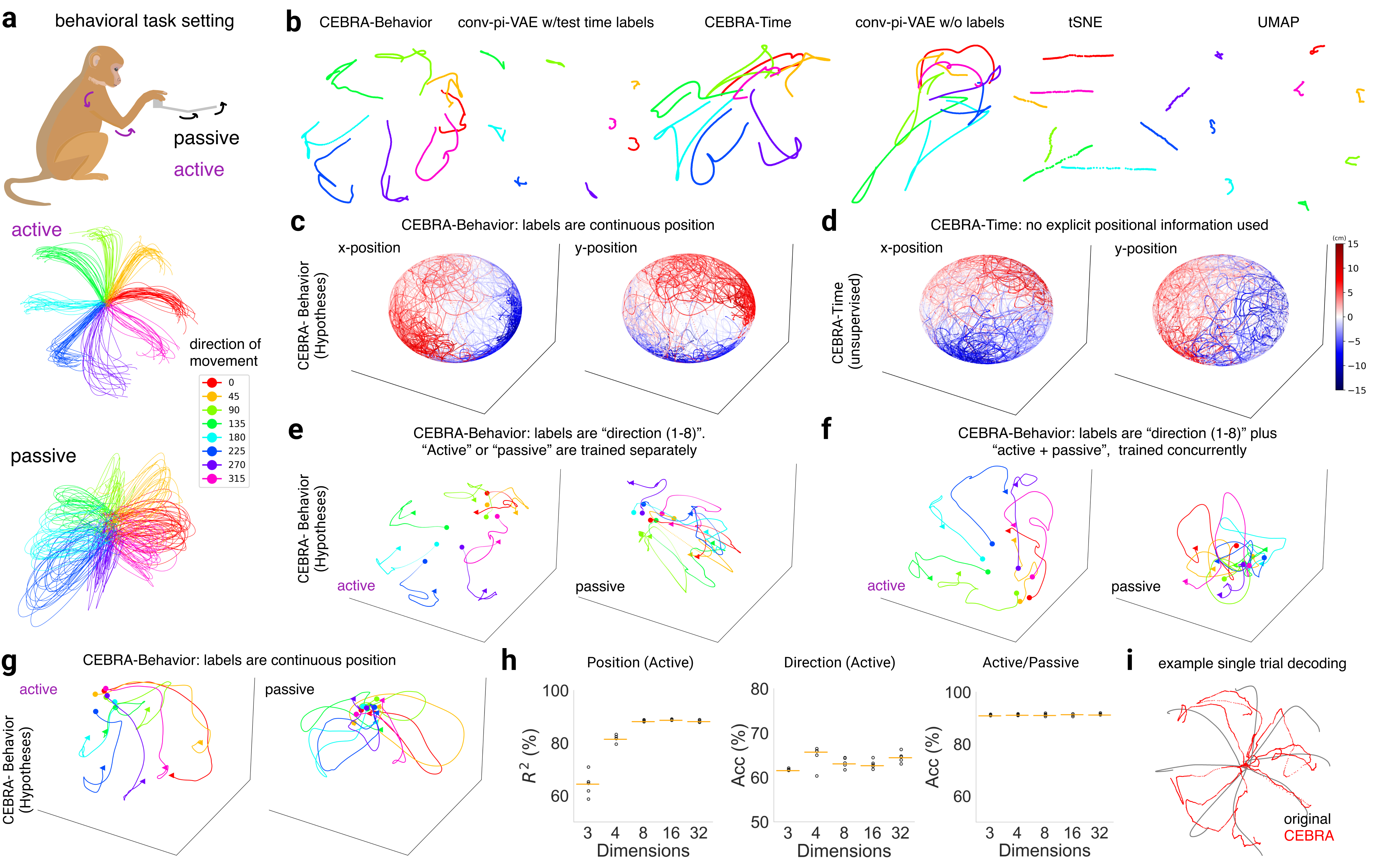}
\end{center}
\vspace{-5pt}
\caption{{\bf Forelimb movement behavior in a primate} 
{\bf (a)}: Behavioral setup: monkey makes either active movements in 8 directions with the manipulandum, or the arm is passively moved via the manipulandum (real behavioral trajectories shown, with cartoon depicting the task setup). Behavior and neural recordings are from area 2 of the primary somatosensory cortex from Chowdhury et al.~\cite{chowdhury2020area}. 
{\bf (b)}: Comparison of embeddings of active trials generated with \cebra-Behavior, \cebra-Time, conv-pi-VAE variants, tSNE, and UMAP. The embeddings of trials (n=364) of each direction are post-hoc averaged.
{\bf (c)}: \cebra-Behavior trained with x,y position of the hand. Left panel is color-coded to x position and right panel is color-coded to y position, as in \textbf{d}.
{\bf (d)}: \cebra-Time without any external behavior variables. As in \textbf{c}, left and right are color-coded to x and y position, respectively.
{\bf (e)}: Left, \cebra-Behavior embedding trained with a 4D latent space, with discrete target direction as behavior labels, trained and plotted separately for active and passive trials. 
{\bf (f)}: Left, \cebra-Behavior embedding trained with a 4D latent space, with discrete target direction and active and passive trials as behavior labels, plotted separately, active vs. passive trials.
{\bf (g)}: \cebra-Behavior embedding trained with a 4D latent space using active and passive trials with continuous (x,y) position as behavior labels, plotted separately, active vs. passive trials. The trajectory of each direction is averaged across trials (n=18--30 each, per directions) over time. Each trajectory represents 600ms from -100ms before the start of the movement.
{\bf (h)}: Left to right: Decoding performance of: position using \cebra-Behavior trained with x,y position (active trials); target direction using \cebra-Behavior trained with target direction (active trials); or active vs. passive accuracy using \cebra-Behavior trained with both active and passive movements. For each case, we trained and evaluated 5 seeds represented by black dot and the orange line represents median.
 {\bf (i)}: Decoded trajectory of hand position using \cebra-Behavior trained on active trial with x,y position of hand. Grey line is true trajectory and red line is decoded trajectory.
 }
\label{fig:Reaching}
\end{figure*}   
    
\subsection*{Co-homology as a metric for robust embeddings}
    
    \justify \cebra \space can be trained across a range of dimensions and models can be selected based on decoding, goodness of fit, and consistency. Yet, we also sought to find a principled approach to verify the robustness of embeddings, which might yield insight into neural computations~\cite{Curto2016WhatCT, Chaudhuri2019TheIA} (Fig.~\ref{fig:Topology}e). We used algebraic topology to measure the persistent co-homology, for comparing if learned latent spaces are equivalent. 
    While it is not required to project embeddings onto a sphere, this has the advantage that there are default Betti numbers (for a $d$-dimensional uniform embedding, $H^0 = 1, H^1 = 0, \cdots, H^{d-1} = 1$, i.e., 1,0,1 for the 2-sphere). 
    We used the distance from the unity line (and thresholded based on a computed null shuffled distribution in Births vs. Deaths to compute Betti numbers; Extended Data Fig.~\ref{fig:SFigure3_persistence}). Using \cebra-Behavior or -Time we find a ring topology (1,1,0; Fig.~\ref{fig:Topology}f), as one would expect from a linear track for place cells. We then computed the Eilenberg-MacLane coordinates for the identified co-cycle (H1) for each model~\cite{Silva2009PersistentCA, Gardner2021.02.25.432776}---this allowed us to map each time-point to topology-preserving coordinates---and indeed we find that the ring topology for the CEBRA models matches space (position) across dimensions (Fig.~\ref{fig:Topology}g, Extended Data Fig.~\ref{fig:SFigure3_persistence}).
    Note, this topology differs from (1,0,1); i.e., Betti numbers for a uniformly covered sphere, which in our setting would indicate a random embedding as found by shuffling (Fig.~\ref{fig:Topology}g).

\begin{figure*}[b]
\begin{center}
\includegraphics[width=\textwidth]{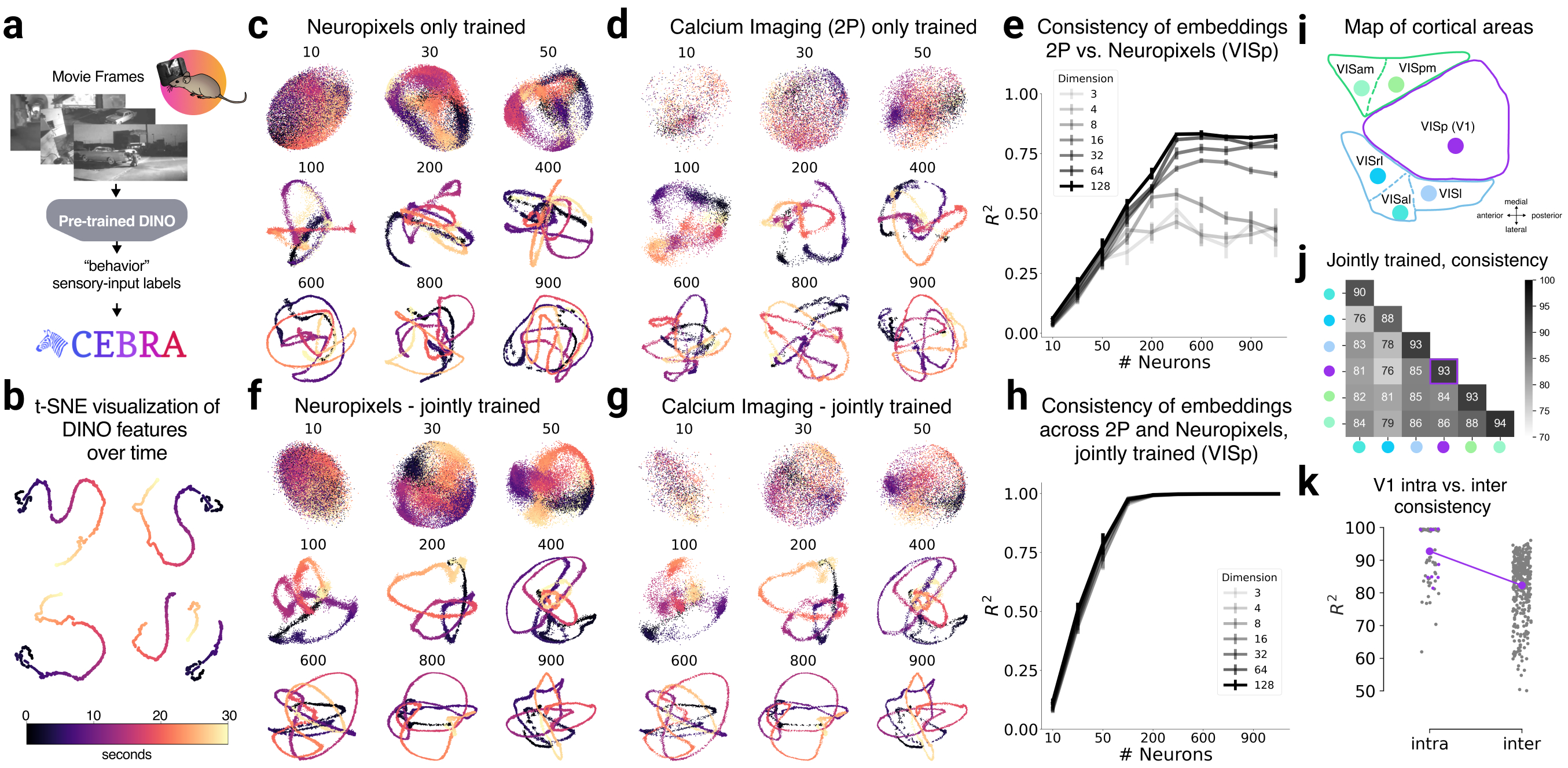}
\end{center}
\vspace{-10pt}
\caption{{\bf Spikes and calcium signaling reveal similar \cebra \space embeddings} {\bf (a)}: \cebra-Behavior can use frame-by-frame video feature as a label of sensory input to extract neural latent space of visual cortex of mice watching a movie.
{\bf (b)}: tSNE visualization of the DINO features of the movie frames from four different DINO configurations (latent size, model size) commonly show continuous evolution of the movie frames over time. 
{\bf (c, d)}: Visualization of trained 8D latent \cebra-Behavior embeddings with Neuropixels data or calcium imaging, respectively. The numbers on top of each embedding is the number of neurons subsampled from the multi-session concatenated dataset. Color map is the same as in \textbf{b}.
{\bf (e)}: Linear consistency between embeddings trained with either calcium imaging data or Neuropixels data. 
{\bf (f, g)}: Visualization of \cebra-Behavior embedding (8D) trained with Neuropixels and calcium imaging, jointly. Color map is the same as in \textbf{b}.
{\bf (h)}: Linear consistency between embeddings of calcium imaging and Neuropixels which were trained jointly using a multi-session CEBRA model. 
{\bf (i)}: Diagram of mouse primary visual cortex (V1, VIsp), PPC (VIsrl) and higher visual areas.
{\bf (j)}: CEBRA-Behavior 32D model jointly trained with 2P+NP with 400 neurons then consistency measured within or across areas (2P vs. NP) across 2 unique sets of disjoint neurons for 3 seeds and averaged.
{\bf (k)}: Models trained as in \textbf{h}, with intra-V1 consistency measurement vs. all inter-area vs. V1 comparison. Purple dots indicate mean of V1 intra-V1 consistency (across n=12 runs) and inter-V1 consistency (n=60). Intra-V1 consistency is significantly higher than inter-area consistency (Welch's t-test, T(19,53)=4.55, p=0.00019).
}
\label{fig:AllenDataFigure}
\end{figure*}  

\subsection*{Multi-session, multi-animal CEBRA}

    \justify \cebra \space can also be used to jointly train across sessions and different animals, which can be highly advantageous when there is limited access to simultaneously recorded neurons, or when looking for animal-invariant features in the neural data.
    We trained  \cebra \space across animals within each multi-animal dataset and find this joint embedding allows for even more consistent embeddings across subjects (Extended Data Fig.~\ref{fig:multiSession}a-c; one-sided, paired T-tests, Allen data: (-5.80), p=\num{5.99e-5}; Hippocampus: (-2.22), p=\num{0.024}). 
    \medskip

    While consistency increased, it is not \textit{a priori} clear that decoding from ``pseudo-subjects'' would be equally good, as there could be session or animal specific information that is lost in pseudo-decoding (as decoding is usually performed within session). Alternatively, if this joint latent space was as high-performance as the single subject, this would suggest that \cebra \space is able to produce robust latent spaces across subjects. Indeed, we find no loss in decoding performance (Extended Data Fig.~\ref{fig:multiSession}c).
    \medskip
    
    It is also possible to rapidly decode from a new session that is \textit{unseen} during training, which is an attractive setting for brain machine interface deployment. We show that by pretraining on a subset of the subjects, we can apply and rapidly adapt \cebra-Behavior on unseen data (i.e., it runs at 50--100 steps/second, and positional decoding error already decreased by 10 cm after adapting the pretrained network for one step). Lastly, we can achieve a lower error more rapidly compared to training fully on the unseen individual (Extended Data Fig.~\ref{fig:multiSession}d). Collectively, this shows that \cebra \space can rapidly produce high-performance, consistent and robust latent spaces.

\subsection*{Discovering latent dynamics during a motor task}

    \justify  We next consider an eight direction ``center-out'' reaching task paired with electrophysiology recordings in somatosensory cortex (S1) of a primate~\cite{chowdhury2020area} (Fig.~\ref{fig:Reaching}a). The monkey performed active movements and in a subset of trials experienced randomized bumps that caused a passive limb movement.
    \cebra \space produced highly informative visualisations of the data compared to other methods (Fig.~\ref{fig:Reaching}b), and \cebra-Behavior can be used in order to test encoding properties of S1. Using position or time information showed embeddings with clear positional encoding (Fig.~\ref{fig:Reaching}c, d, and Extended Data Fig.~\ref{SupplReaching}a-c).
    \medskip
    
    To then test how directional information and active vs. passive movements influence population dynamics in S1~\cite{Prudhomme1994ProprioceptiveAI,London2013ResponsesOS, chowdhury2020area}, we trained embedding spaces with directional information and then either separated the trials into active and passive for training (Fig.~\ref{fig:Reaching}e), or trained jointly and post-hoc plotted separately (Fig.~\ref{fig:Reaching}f).
    We find striking similarities that suggest active vs. passive strongly influences the neural latent space: the embeddings for active trials show a clear start and stop, while for passive trials it shows a continuous trajectory through the embedding, independently of how they are trained. This finding is confirmed in embeddings that used only the \textit{continuous} position of the end-effector as the behavioral label (Fig.~\ref{fig:Reaching}g). Notably, direction is a less prominent feature (Fig.~\ref{fig:Reaching}g), although they are entangled parameters in this task. %
     \medskip
   
    Next, since position and active or passive trial type appear robust in the embeddings, we further explored the decodability of the embeddings. Both position and trial type were readily decodable from 8D+ embeddings with a kNN decoder trained on position-only, but directional information was not as decodable (Fig.~\ref{fig:Reaching}h). Here too the loss function is informative for hypothesis testing (Extended Data Fig.~\ref{SupplReaching}d-f). Notably, we could recover the hand trajectory with an $R^2$ of $88\%$ (concatenated across 26 held-out test trials, Fig.~\ref{fig:Reaching}i) using a 16D CEBRA-Behavior model trained on position (Fig.~\ref{fig:Reaching}i). For comparison, a L1 regression using all neurons achieved $R^2$ $74\%$, and 16D conv-pi-VAE achieved $R^2$ $82\%$, making our approach state-of-the-art on this data.

\subsection*{Consistent embeddings across modalities}

     \justify \cebra \space is agnostic to the recording modality of neural data. But do different modalities produce similar latent embeddings?
     Understanding the relationship of calcium signaling and electrophysiology is a debated topic, yet an underlying assumption is that they inherently encode related, yet not identical, information. Although there are a wealth of excellent tools aimed at inferring spike trains from calcium data,  currently the pseudo-R$^2$ of algorithms on paired spiking and calcium data tops out at around 0.6~\cite{Berens2018CommunitybasedBI}. Nonetheless, it is clear that recording with either modality has lead to similar global conclusions---for example, grid cells can be uncovered in spiking or calcium signals~\cite{Hafting2005MicrostructureOA, Gardner2021.02.25.432776}, reward prediction errors can be found in dopamine neurons across species and recording modalities~\cite{Schultz1997ANS, Cohen2012NeurontypeSS, Menegas2015DopamineNP}, and visual cortex shows orientation tuning across species and modalities~\cite{Hubel1977FerrierL,Niell2008HighlySR,Ringach2016SpatialCO}. 
     \medskip
     
     We aimed to formally study whether \cebra \space could capture the same neural population dynamics either from spikes or calcium imaging. We utilized a dataset from the Allen Brain Observatory where mice passively watched three movies repeatedly. We focused on paired data from 10 repeats of ``Natural Movie 1'' where neural data were recorded with either Neuropixels probes or via calcium imaging with a 2-photon (2P) microscope (from separate mice)~\cite{de2020large, Siegle2021SurveyOS}.  
     Note, the data we considered thus far have goal-driven actions of the animals (such as running down a linear track or reaching to targets), yet this visual cortex dataset is collected during passive viewing (Fig.~\ref{fig:AllenDataFigure}a).
     \medskip

     We used the movie features as ``behavior'' labels by extracting the high-level visual features from the movie on a frame-by-frame basis using DINO, a powerful vision transformer~\cite{Caron2021EmergingPI}. Those were then used to sample the neural data with feature-labels (Fig.~\ref{fig:AllenDataFigure}b). Next, we used Neuropixels data or calcium (2P) data (each with multi-session training) in order to generate (from 8D to 128D) latent spaces from varying number of neurons recorded from V1 (Fig.~\ref{fig:AllenDataFigure}c, d). The visualization of CEBRA-Behavior showed trajectories that smoothly capture the video with either modality with an increasing number of neurons. This is reflected quantitatively in the consistency metric (Fig.~\ref{fig:AllenDataFigure}e). Strikingly, CEBRA-Time nicely captured the 10 repeats of the movie (Extended Data Fig.~\ref{fig:SUPPLAllenDataFigureMice}).
     This result demonstrates that there is a highly consistent latent space independent of the recording method. 
    \medskip
    
     Next, we stacked the neurons from different mice and modalities and then sampled random subsets of V1 neurons to construct a pseudo-mouse.
     We did not find that joint training lowered consistency within modality (Extended Data Fig.~\ref{fig:SUPPLAllenDataFigure}a, b), and overall we found considerable improvement in consistency with joint training (Fig.~\ref{fig:AllenDataFigure}f-h). 
     \medskip
     
     Using CEBRA-Behavior or -Time we trained models on four higher visual areas (HVAs) and one posterior parietal cortex (PPC) area and measured the consistency with and without joint training, and within or across areas. Our results show that with joint training intra-area consistency is higher vs. other areas (Fig.~\ref{fig:AllenDataFigure}i-k), suggesting that with \cebra \space we are not removing biological differences across areas (that have known differences in decodability and feature representations~\cite{Esfahany2018OrganizationON,Jin2020MouseHV}). Moreover, we test within modality and find a similar effect with CEBRA-Behavior or -Time within recording modality (Extended Data Fig.~\ref{fig:SUPPLAllenDataFigure}c-f).
     \medskip

\begin{figure*}[t]
\begin{center}
\includegraphics[width=\textwidth]{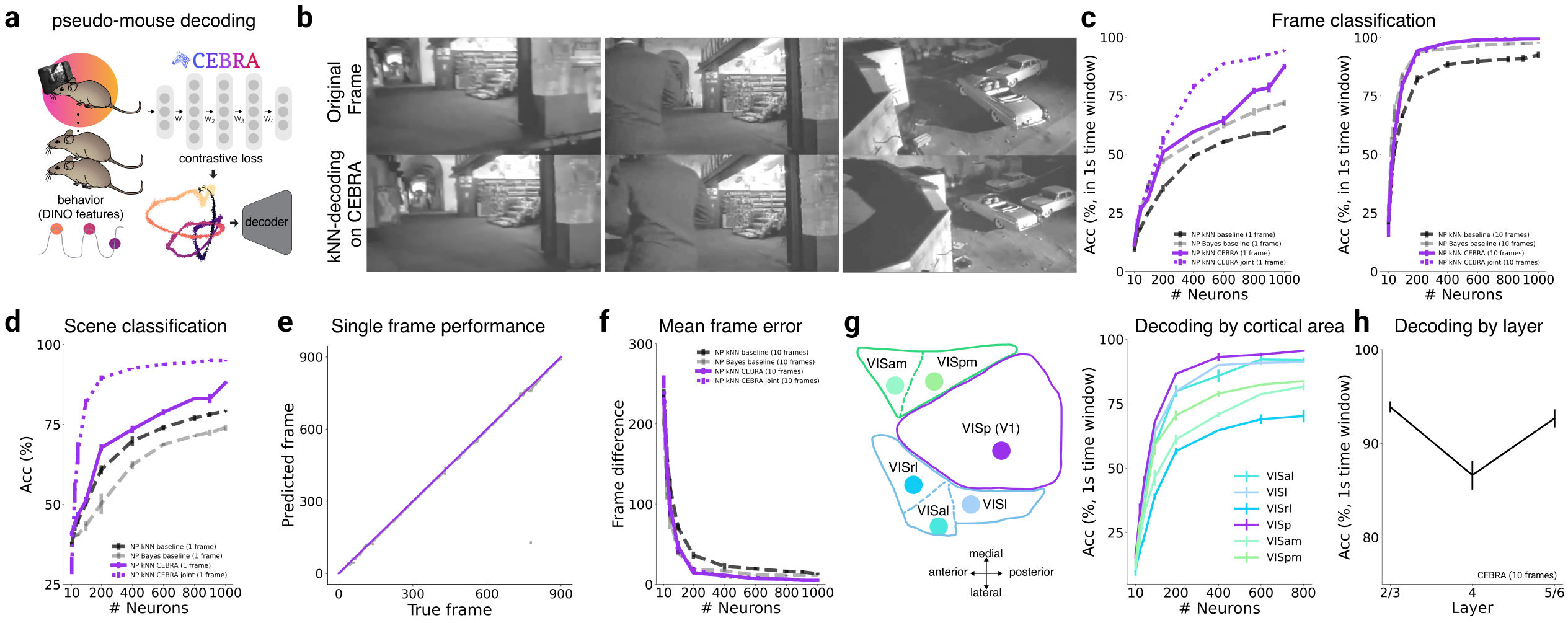}
\end{center}
\vspace{-10pt}
\caption{{\bf Decoding of natural movie features from mouse visual cortical areas.}
{\bf (a)}: Schematic of the CEBRA encoder and kNN (or naive Bayes) decoder.
{\bf (b)}: Examples of original frames (top row) and frames decoded from CEBRA embedding of V1 calcium recording using kNN decoding (bottom row). The last repeat among 10 repeats was used as the held-out test. 
{\bf (c)}: Decoding accuracy measured by considering a predicted frame being within 1 sec to the true frame as a correct prediction using CEBRA (NP only), jointly trained (2P+NP), or a baseline population-vector plus kNN or naive Bayes decoder using either a 1 frame (33 ms) receptive field or 10 frames (330 ms); results shown for Neuropixels dataset (V1 data).
{\bf (d)}: Decoding accuracy measured by the correct scene prediction using either CEBRA (NP only), jointly trained (2P+NP), or baseline population-vector plus kNN or Bayes decoder using a 1 frame (33 ms) receptive field (V1 data).
{\bf (e)}: Single frame ground truth frame ID vs predicted frame ID for Neuropixels using a \cebra-Behavior model trained with a 330 ms receptive field (1,000 V1 neurons across mice used).
{\bf (f)}: The mean absolute error of the correct frame index. Shown for baseline and CEBRA models as computed in c, d, e.
{\bf (g)}: Diagram of the cortical areas considered, and decoding performance from CEBRA (NP only), 10 frame receptive field.
{\bf (h)}: V1 decoding performance vs. layer category using 900 neurons with a 330 ms receptive field \cebra-Behavior model.
}
\label{fig:AllenDataDecoding}
\end{figure*}  

\subsection*{Decoding of natural movies from visual cortex}
     
    \justify We performed V1 decoding analysis using CEBRA models that are either joint-modality trained, single-modality trained, or with a baseline population vector then paired with a simple kNN or naive Bayes decoder. We aimed to see if we could decode on a frame-by-frame basis the natural movie the mice watched. We used the last movie repeat as a held-out test set and nine repeats as the training set. We could achieve greater than 95\% decoding accuracy, which is significantly better than the baseline decoding methods (naive Bayes or kNN) for Neuropixels recordings, and joint training CEBRA outperformed Neuropixels-only CEBRA based training (single frame: one-way ANOVA, F(3,197)=5.88, p=\num{0.0007}, Tables~\ref{table:dataSTATS_10frame}, ~\ref{table:dataSTATS_1frame}, ~\ref{table:dataSTATS_SCENE}, Fig.~\ref{fig:AllenDataDecoding}a-d, Extended Data Fig.~\ref{fig:SUPPLAllenDataFigure}g, h). Accuracy was defined as the fraction of correct frames within a 1-second window or by the correct scene being identified. Frame-by-frame results also showed reduced frame ID errors (one-way ANOVA, F(3,16)=20.22, p=\num{1.09e-5}, n=1000 neurons, Table~\ref{table:dataSTATS_FRAMEDIFF}) which can be appreciated in Fig.~\ref{fig:AllenDataDecoding}e, f, Extended Data Fig.~\ref{fig:SUPPLAllenDataFigure}i, and Suppl. Video 2. The DINO features themselves did not drive performance, as shuffling the features showed poor decoding (Extended Data Fig.~\ref{fig:SUPPLAllenDataFigure}j).
     \medskip

     Lastly, we tested decoding from other HVAs and PPC using DINO features. Overall, decoding from V1 had the highest performance, and PPC (VISrl) the lowest (Fig.~\ref{fig:AllenDataDecoding}g, Extended Data Fig.~\ref{fig:SUPPLAllenDataFigure}k). Given the high decoding performance of \cebra, we tested if there was a particular V1 layer that was most informative. We leveraged \cebra-Behavior by training models on each category and find that layer 2/3 and layer 5/6 show significantly higher decoding performance compared to layer 4 (one-way ANOVA, F(2,12)=9.88, p=0.003; Fig.~\ref{fig:AllenDataDecoding}h). 
     Given the known cortical connectivity, this suggests that the non-thalamic input layers make frame information more explicit, perhaps via feedback or predictive processing.

\section*{Discussion}

    Mapping neural activity to behavioral outputs is one of the fundamental quests of neuroscience. Here, we present \cebra, a new dimensionality reduction method to explicitly leverage behavior or time in order to discover latent neural embeddings. We find these embeddings provide high decoding performance across a broad spectrum of behaviors---from positional decoding in hippocampus to reconstruction of natural movies from visual cortex in the mouse. \cebra \space produces both consistent embeddings across subjects (thus revealing common structure) and can find the dimensionality of neural spaces that are topologically robust. While there remains a gap in understanding how these latent spaces map to neural-level computations, we believe this tool provides an advance in our ability to map behavior to neural populations.
\medskip 

    Contrastive learning is highly attractive to use in this problem setting of using so-called auxiliary variables~\cite{Hyvrinen2019NonlinearIU,khosla2020supervised}. Recent work to develop more robust non-linear ICA has also shown that the InfoNCE loss encodes useful inductive biases~\citep{Zimmermann2021ContrastiveLI}. The unique property of \cebra \space is the extension of the standard InfoNCE objective by introducing a variety of different \emph{sampling strategies} tuned for usage of the algorithm in the experimental sciences, and for analysis of time series datasets. In contrast to other usages of contrastive learning~\cite{Chen2020ASF}, \cebra \space does not rely on data augmentation techniques (that need to be designed specifically for a particular dataset, potentially using domain knowledge), and is still flexible and easy to adapt to different data processing needs.
\medskip

    Dimensionality reduction is often tightly linked to data visualization, and here we make an empirical argument that ultimately this is only useful when you are getting consistent results, and discovering robust features. Unsupervised tSNE and UMAP are examples of algorithms widely used in life sciences for discovery-based analysis. However, they do not leverage time, and for neural recordings, this is always available and can be used. Even more critical is that concatenating data from different animals can lead to shifted clusters with tSNE or UMAP due to inherent small changes across animals or in how the data was collected. CEBRA allows the user to remove this unwanted variance and discover robust latents that are invariant to animal ID, sessions, or any-other-user-defined nuisance variable. Collectively, we believe CEBRA will become a complement to (or replacement for) these methods such that, at minimum, the structure of time in the neural code is leveraged, and robustness is prioritized.
\medskip

    CEBRA is highly versatile: it can be used for supervised and self-supervised analysis and thereby directly allows for hypothesis- and discovery-driven science (Fig.~\ref{fig:Topology}). For example, our multi-session and multi-animal training allows for domain generalization and mitigation of batch effects common in biological data (i.e., constant distribution shifts that appear between recording days or sessions due to static changes in the experimental setup, acquisition method, etc.). It also allows for exploratory data analysis of time series data only, and/or data paired with a rich variety of context variables that can be used to do hypothesis-driven decoding (e.g., recordings of other---potentially confounding---signals, such as pose estimation, EMG signals, etc.), where, for example, testing dependencies between variables, or difference of experimental conditions or recording setups is of interest. We demonstrate this feature by using both continuous labels (such as position from the hippocampus task setting, Fig.~\ref{fig:CEBRAvsAll}), or discrete labels, such as ``active'' and ``passive'' in the monkey reaching dataset (Fig.~\ref{fig:Reaching}). We also show that it does not require kinematic data, as any ``behavior'' labels are usable, such as DINO-based features from a natural movie which we show can be powerfully used to decode on a frame-by-frame basis images from the visual cortex of mice (Figs.~\ref{fig:AllenDataFigure}, ~\ref{fig:AllenDataDecoding}).
\medskip

\medskip

    We demonstrate the scientific utility of CEBRA by exploring datasets collected from visual areas of mice while they passively observe a natural movie. We find that we can decode frames with greater than 95\% accuracy from a  ``pseudo-mouse'' model trained from both Neuropixels and 2P data across animals. To achieve this result, we first showed that CEBRA outperforms classical algorithms such as UMAP and tSNE and state-of-the-art neural denoising/decoding algorithm, pi-VAE. In the course of us attempting to fairly benchmark our method we incidentally improved pi-VAE (Extended Data~\ref{fig:CEBRAintroData}). Nonetheless, we show CEBRA can significantly outperform our modified conv-pi-VAE in consistency, and decodability. Secondly, we showed that we could jointly train across animals, a feature that is not present in other methods benchmarked here (but see~\cite{Pandarinath2018InferringSN}), to generate more robust (consistent) latent spaces. 
\medskip

    Pretrained CEBRA models can be used for decoding in new animals within tens of steps (milliseconds); we can thereby get equal or better performance compared to training on the unseen animal alone. Considering the fact that time efficiency is a highly relevant factor especially in brain machine interface applications, it is worthwhile to note that \cebra \space provides much faster training compared to pi-VAE where the sampling method to approximate the test label is very time-costly. We believe our approach will be crucial for real-time, adaptive decoding. 
\medskip

\subsection*{Data Availability}

\justify Hippocampus dataset: \url{https://crcns.org/data-sets/hc/hc-11/about-hc-11} and we used the preprocessing script from \url{https://github.com/zhd96/pi-vae/blob/main/code/rat_preprocess_data.py}.
Primate dataset: \url{https://gui.dandiarchive.org/#/dandiset/000127}.
Allen Institute dataset: Neuropixels data are at \url{https://allensdk.readthedocs.io/en/latest/visual_coding_neuropixels.html}. The pre-processed 2P recordings are available at \url{https://github.com/zivlab/visual_drift/tree/main/data}.
As examples with CEBRA, packaged datasets are available at \url{https://github.com/AdaptiveMotorControlLab/CEBRA}.

\subsection*{Code Availability}

\justify Code: \url{https://github.com/AdaptiveMotorControlLab/CEBRA}. Documentation: \url{https://cebra.ai/docs/}. Code (and data) to reproduce the figures: \url{https://github.com/AdaptiveMotorControlLab/CEBRA-figures}. All other requests should be made to the corresponding author. 

\section*{References}
\bibliography{references}

\begin{thebibliography}{62}
\providecommand{\natexlab}[1]{#1}
\providecommand{\url}[1]{\texttt{#1}}
\expandafter\ifx\csname urlstyle\endcsname\relax
  \providecommand{\doi}[1]{doi: #1}\else
  \providecommand{\doi}{doi: \begingroup \urlstyle{rm}\Url}\fi

\bibitem[Urai et~al.(2022)Urai, Doiron, Leifer, and
  Churchland]{Urai2022LargescaleNR}
Anne~E. Urai, Brent Doiron, Andrew~Michael Leifer, and Anne~K. Churchland.
\newblock Large-scale neural recordings call for new insights to link brain and
  behavior.
\newblock \emph{Nature Neuroscience}, 25:\penalty0 11--19, 2022.

\bibitem[Krakauer et~al.(2017)Krakauer, Ghazanfar, Gomez-Marin, MacIver, and
  Poeppel]{Krakauer2017NeuroscienceNB}
John~W. Krakauer, Asif~A. Ghazanfar, Alex Gomez-Marin, Malcolm~A. MacIver, and
  David Poeppel.
\newblock Neuroscience needs behavior: Correcting a reductionist bias.
\newblock \emph{Neuron}, 93:\penalty0 480--490, 2017.

\bibitem[Jazayeri and Ostojic(2021)]{Jazayeri2021InterpretingNC}
Mehrdad Jazayeri and Srdjan Ostojic.
\newblock Interpreting neural computations by examining intrinsic and embedding
  dimensionality of neural activity.
\newblock \emph{Current Opinion in Neurobiology}, 70:\penalty0 113--120, 2021.

\bibitem[Humphries(2021)]{humphries2021strong}
Mark~D Humphries.
\newblock Strong and weak principles of neural dimension reduction, 2021.

\bibitem[Zhou and Wei(2020)]{zhou2020learning}
Ding Zhou and Xue{-}Xin Wei.
\newblock Learning identifiable and interpretable latent models of
  high-dimensional neural activity using pi-vae.
\newblock In \emph{Advances in Neural Information Processing Systems 33}, 2020.

\bibitem[Vargas-Irwin et~al.(2010)Vargas-Irwin, Shakhnarovich, Yadollahpour,
  Mislow, Black, and Donoghue]{vargas2010decoding}
Carlos~E Vargas-Irwin, Gregory Shakhnarovich, Payman Yadollahpour, John~MK
  Mislow, Michael~J Black, and John~P Donoghue.
\newblock Decoding complete reach and grasp actions from local primary motor
  cortex populations.
\newblock \emph{Journal of neuroscience}, 30\penalty0 (29):\penalty0
  9659--9669, 2010.

\bibitem[Okorokova et~al.(2020)Okorokova, Goodman, Hatsopoulos, and
  Bensmaia]{Okorokova2020DecodingHK}
Elizaveta~V Okorokova, James~M. Goodman, Nicholas~G. Hatsopoulos, and Sliman~J.
  Bensmaia.
\newblock Decoding hand kinematics from population responses in sensorimotor
  cortex during grasping.
\newblock \emph{Journal of neural engineering}, 2020.

\bibitem[Yu et~al.(2008)Yu, Cunningham, Santhanam, Ryu, Shenoy, and
  Sahani]{Yu2008GaussianprocessFA}
Byron~M. Yu, John~P. Cunningham, Gopal Santhanam, Stephen~I. Ryu, Krishna~V.
  Shenoy, and Maneesh Sahani.
\newblock Gaussian-process factor analysis for low-dimensional single-trial
  analysis of neural population activity.
\newblock \emph{Journal of neurophysiology}, 102 1:\penalty0 614--35, 2008.

\bibitem[Churchland et~al.(2012)Churchland, Cunningham, Kaufman, Foster,
  Nuyujukian, Ryu, and Shenoy]{Churchland2012NeuralPD}
MM~Churchland, JP~Cunningham, M.~Kaufman, J.~Foster, Paul Nuyujukian, Si~Ryu,
  and K.~V. Shenoy.
\newblock Neural population dynamics during reaching.
\newblock \emph{Nature}, 487:\penalty0 51 -- 56, 2012.

\bibitem[Gallego et~al.(2018)Gallego, Perich, Naufel, Ethier, Solla, and
  Miller]{Gallego2018CorticalPA}
Juan~Alvaro Gallego, Matthew~G. Perich, Stephanie Naufel, Christian Ethier,
  Sara~A. Solla, and Lee~E. Miller.
\newblock Cortical population activity within a preserved neural manifold
  underlies multiple motor behaviors.
\newblock \emph{Nature Communications}, 9, 2018.

\bibitem[McInnes et~al.(2018)McInnes, Healy, and Melville]{mcinnes2018umap}
Leland McInnes, John Healy, and James Melville.
\newblock Umap: Uniform manifold approximation and projection for dimension
  reduction.
\newblock \emph{arXiv preprint arXiv:1802.03426}, 2018.

\bibitem[Van Der~Maaten et~al.(2009)Van Der~Maaten, Postma, Van~den Herik,
  et~al.]{van2009dimensionality}
Laurens Van Der~Maaten, Eric Postma, Jaap Van~den Herik, et~al.
\newblock Dimensionality reduction: a comparative.
\newblock \emph{J Mach Learn Res}, 10\penalty0 (66-71):\penalty0 13, 2009.

\bibitem[Roeder et~al.(2020)Roeder, Metz, and Kingma]{Roeder2020}
Geoffrey Roeder, Luke Metz, and Diederik~P. Kingma.
\newblock On linear identifiability of learned representations.
\newblock \emph{arXiv}, 2020.
\newblock \doi{10.48550/ARXIV.2007.00810}.

\bibitem[Hyv{\"{a}}rinen et~al.(2019)Hyv{\"{a}}rinen, Sasaki, and
  Turner]{Hyvrinen2019NonlinearIU}
Aapo Hyv{\"{a}}rinen, Hiroaki Sasaki, and Richard~E. Turner.
\newblock Nonlinear {ICA} using auxiliary variables and generalized contrastive
  learning.
\newblock In \emph{The 22nd International Conference on Artificial Intelligence
  and Statistics}, volume~89 of \emph{Proceedings of Machine Learning
  Research}, pages 859--868. {PMLR}, 2019.

\bibitem[Sani et~al.(2020)Sani, Abbaspourazad, Wong, Pesaran, and
  Shanechi]{Sani2020ModelingBR}
Omid~G. Sani, Hamidreza Abbaspourazad, Y.~Wong, Bijan Pesaran, and M.~Shanechi.
\newblock Modeling behaviorally relevant neural dynamics enabled by
  preferential subspace identification.
\newblock \emph{Nature Neuroscience}, 24:\penalty0 140--149, 2020.

\bibitem[Klindt et~al.(2021)Klindt, Schott, Sharma, Ustyuzhaninov, Brendel,
  Bethge, and Paiton]{klindt2021towards}
David~A. Klindt, Lukas Schott, Yash Sharma, Ivan Ustyuzhaninov, Wieland
  Brendel, Matthias Bethge, and Dylan Paiton.
\newblock Towards nonlinear disentanglement in natural data with temporal
  sparse coding.
\newblock In \emph{International Conference on Learning Representations}, 2021.

\bibitem[Pandarinath et~al.(2018)Pandarinath, O’Shea, Collins,
  J{\'o}zefowicz, Stavisky, Kao, Trautmann, Kaufman, Ryu, Hochberg, Henderson,
  Shenoy, Abbott, and Sussillo]{Pandarinath2018InferringSN}
Chethan Pandarinath, Daniel~J. O’Shea, Jasmine Collins, Rafal J{\'o}zefowicz,
  Sergey~D. Stavisky, Jonathan~C. Kao, Eric~M. Trautmann, Matthew~T. Kaufman,
  Stephen~I. Ryu, Leigh~R. Hochberg, Jaimie~M. Henderson, Krishna~V. Shenoy,
  L.~F. Abbott, and David Sussillo.
\newblock Inferring single-trial neural population dynamics using sequential
  auto-encoders.
\newblock \emph{Nature methods}, 15:\penalty0 805 -- 815, 2018.

\bibitem[Prince et~al.(2021)Prince, Bakhtiari, Gillon, and
  Richards]{Prince2021ParallelIO}
Luke~Y. Prince, Shahab Bakhtiari, Colleen~J. Gillon, and Blake~A. Richards.
\newblock Parallel inference of hierarchical latent dynamics in two-photon
  calcium imaging of neuronal populations.
\newblock \emph{bioRxiv}, 2021.

\bibitem[Gutmann and Hyv\"{a}rinen(2012)]{Gutmann12JMLR}
Michael~U. Gutmann and Aapo Hyv\"{a}rinen.
\newblock Noise-contrastive estimation of unnormalized statistical models, with
  applications to natural image statistics.
\newblock \emph{The Journal of Machine Learning Research}, 13:\penalty0
  307--361, 2012.

\bibitem[Oord et~al.(2018)Oord, Li, and Vinyals]{oord2018representation}
Aaron van~den Oord, Yazhe Li, and Oriol Vinyals.
\newblock Representation learning with contrastive predictive coding.
\newblock \emph{arXiv preprint arXiv:1807.03748}, 2018.

\bibitem[Khosla et~al.(2020)Khosla, Teterwak, Wang, Sarna, Tian, Isola,
  Maschinot, Liu, and Krishnan]{khosla2020supervised}
Prannay Khosla, Piotr Teterwak, Chen Wang, Aaron Sarna, Yonglong Tian, Phillip
  Isola, Aaron Maschinot, Ce~Liu, and Dilip Krishnan.
\newblock Supervised contrastive learning.
\newblock \emph{arXiv preprint arXiv:2004.11362}, 2020.

\bibitem[Chen et~al.(2020)Chen, Kornblith, Norouzi, and Hinton]{Chen2020ASF}
Ting Chen, Simon Kornblith, Mohammad Norouzi, and Geoffrey~E. Hinton.
\newblock A simple framework for contrastive learning of visual
  representations.
\newblock \emph{ArXiv}, abs/2002.05709, 2020.

\bibitem[Grosmark and Buzs{\'a}ki(2016)]{grosmark2016diversity}
Andres~D Grosmark and Gy{\"o}rgy Buzs{\'a}ki.
\newblock Diversity in neural firing dynamics supports both rigid and learned
  hippocampal sequences.
\newblock \emph{Science}, 351\penalty0 (6280):\penalty0 1440--1443, 2016.

\bibitem[H{\"a}lv{\"a} et~al.(2021)H{\"a}lv{\"a}, Corff, Leh'ericy, So, Zhu,
  Gassiat, and Hyv{\"a}rinen]{Hlv2021DisentanglingIF}
Hermanni H{\"a}lv{\"a}, Sylvain~Le Corff, Luc Leh'ericy, Jonathan So, Yongjie
  Zhu, Elisabeth Gassiat, and Aapo Hyv{\"a}rinen.
\newblock Disentangling identifiable features from noisy data with structured
  nonlinear ica.
\newblock \emph{ArXiv}, abs/2106.09620, 2021.

\bibitem[Zimmermann et~al.(2021)Zimmermann, Sharma, Schneider, Bethge, and
  Brendel]{Zimmermann2021ContrastiveLI}
Roland~S. Zimmermann, Yash Sharma, Steffen Schneider, Matthias Bethge, and
  Wieland Brendel.
\newblock Contrastive learning inverts the data generating process.
\newblock In \emph{Proceedings of the 38th International Conference on Machine
  Learning}, volume 139 of \emph{Proceedings of Machine Learning Research},
  pages 12979--12990. {PMLR}, 2021.

\bibitem[Huxter et~al.(2003)Huxter, Burgess, and
  O’Keefe]{Huxter2003IndependentRA}
John~R. Huxter, Neil Burgess, and John O’Keefe.
\newblock Independent rate and temporal coding in hippocampal pyramidal cells.
\newblock \emph{Nature}, 425:\penalty0 828--832, 2003.

\bibitem[Moser et~al.(2008)Moser, Kropff, and Moser]{Moser2008PlaceCG}
Edvard~I. Moser, Emilio Kropff, and May-Britt Moser.
\newblock Place cells, grid cells, and the brain's spatial representation
  system.
\newblock \emph{Annual review of neuroscience}, 31:\penalty0 69--89, 2008.

\bibitem[Dombeck et~al.(2010)Dombeck, Harvey, Tian, Looger, and
  Tank]{Dombeck2010FunctionalIO}
Daniel~A. Dombeck, Christopher~D. Harvey, Lin Tian, Loren~L. Looger, and
  David~W. Tank.
\newblock Functional imaging of hippocampal place cells at cellular resolution
  during virtual navigation.
\newblock \emph{Nature neuroscience}, 13:\penalty0 1433 -- 1440, 2010.

\bibitem[Chowdhury et~al.(2020)Chowdhury, Glaser, and
  Miller]{chowdhury2020area}
Raeed~H Chowdhury, Joshua~I Glaser, and Lee~E Miller.
\newblock Area 2 of primary somatosensory cortex encodes kinematics of the
  whole arm.
\newblock \emph{ELife}, 9:\penalty0 e48198, 2020.

\bibitem[Curto(2016)]{Curto2016WhatCT}
Carina Curto.
\newblock What can topology tell us about the neural code.
\newblock \emph{arXiv: Neurons and Cognition}, 2016.

\bibitem[Chaudhuri et~al.(2019)Chaudhuri, Gerçek, Pandey, Peyrache, and
  Fiete]{Chaudhuri2019TheIA}
Rishidev Chaudhuri, Berk Gerçek, Biraj Pandey, Adrien Peyrache, and Ila~R.
  Fiete.
\newblock The intrinsic attractor manifold and population dynamics of a
  canonical cognitive circuit across waking and sleep.
\newblock \emph{Nature Neuroscience}, 22:\penalty0 1512--1520, 2019.

\bibitem[de~Silva et~al.(2009)de~Silva, Morozov, and
  Vejdemo-Johansson]{Silva2009PersistentCA}
Vin de~Silva, Dmitriy Morozov, and Mikael Vejdemo-Johansson.
\newblock Persistent cohomology and circular coordinates.
\newblock \emph{Discrete \& Computational Geometry}, 45:\penalty0 737--759,
  2009.

\bibitem[Gardner et~al.(2022)Gardner, Hermansen, Pachitariu, Burak, Baas, Dunn,
  Moser, and Moser]{Gardner2021.02.25.432776}
Richard~J. Gardner, Erik Hermansen, Marius Pachitariu, Yoram Burak, Nils~A.
  Baas, Benjamin~A. Dunn, May-Britt Moser, and Edvard~I. Moser.
\newblock Toroidal topology of population activity in grid cells.
\newblock \emph{Nature}, 602\penalty0 (7895):\penalty0 123--128, Feb 2022.
\newblock ISSN 1476-4687.
\newblock \doi{10.1038/s41586-021-04268-7}.

\bibitem[Prud'homme and Kalaska(1994)]{Prudhomme1994ProprioceptiveAI}
M.~J. Prud'homme and John~F. Kalaska.
\newblock Proprioceptive activity in primate primary somatosensory cortex
  during active arm reaching movements.
\newblock \emph{Journal of neurophysiology}, 72 5:\penalty0 2280--301, 1994.

\bibitem[London and Miller(2013)]{London2013ResponsesOS}
Brian~M. London and Lee~E. Miller.
\newblock Responses of somatosensory area 2 neurons to actively and passively
  generated limb movements.
\newblock \emph{Journal of neurophysiology}, 109 6:\penalty0 1505--13, 2013.

\bibitem[Berens et~al.(2018)Berens, Freeman, Deneux, Chenkov, McColgan,
  Speiser, Macke, Turaga, Mineault, Rupprecht, Gerhard, Friedrich, Friedrich,
  Paninski, Pachitariu, Harris, Bolte, Machado, Ringach, Stone, Rogerson,
  Sofroniew, Reimer, Froudarakis, Euler, Ros{\'o}n, Theis, Tolias, and
  Bethge]{Berens2018CommunitybasedBI}
Philipp Berens, Jeremy Freeman, Thomas Deneux, Nicolay Chenkov, Thomas
  McColgan, Artur Speiser, Jakob~H. Macke, Srinivas~C. Turaga, Patrick~J.
  Mineault, Peter Rupprecht, Stephan Gerhard, Rainer~W. Friedrich, Johannes
  Friedrich, Liam Paninski, Marius Pachitariu, Kenneth~D. Harris, Ben Bolte,
  Timothy~A. Machado, Dario~L. Ringach, Jasmine Stone, Luke~E. Rogerson,
  Nicolas~J. Sofroniew, Jacob Reimer, Emmanouil Froudarakis, Thomas Euler,
  Miroslav~Rom{\'a}n Ros{\'o}n, Lucas Theis, Andreas~Savas Tolias, and Matthias
  Bethge.
\newblock Community-based benchmarking improves spike rate inference from
  two-photon calcium imaging data.
\newblock \emph{PLoS Computational Biology}, 14, 2018.

\bibitem[Hafting et~al.(2005)Hafting, Fyhn, Molden, Moser, and
  Moser]{Hafting2005MicrostructureOA}
Torkel Hafting, Marianne Fyhn, Sturla Molden, May-Britt Moser, and Edvard~I.
  Moser.
\newblock Microstructure of a spatial map in the entorhinal cortex.
\newblock \emph{Nature}, 436:\penalty0 801--806, 2005.

\bibitem[Schultz et~al.(1997)Schultz, Dayan, and Montague]{Schultz1997ANS}
Wolfram Schultz, Peter Dayan, and P.~Read Montague.
\newblock A neural substrate of prediction and reward.
\newblock \emph{Science}, 275:\penalty0 1593 -- 1599, 1997.

\bibitem[Cohen et~al.(2012)Cohen, Haesler, Vong, Lowell, and
  Uchida]{Cohen2012NeurontypeSS}
Jeremiah~Y. Cohen, Sebastian Haesler, Linh Vong, Bradford~B. Lowell, and
  Naoshige Uchida.
\newblock Neuron-type specific signals for reward and punishment in the ventral
  tegmental area.
\newblock \emph{Nature}, 482:\penalty0 85 -- 88, 2012.

\bibitem[Menegas et~al.(2015)Menegas, Bergan, Ogawa, Isogai, Venkataraju,
  Osten, Uchida, and Watabe-Uchida]{Menegas2015DopamineNP}
William Menegas, Joseph~F Bergan, Sachie~K. Ogawa, Yoh Isogai, Kannan~Umadevi
  Venkataraju, Pavel Osten, Naoshige Uchida, and Mitsuko Watabe-Uchida.
\newblock Dopamine neurons projecting to the posterior striatum form an
  anatomically distinct subclass.
\newblock \emph{eLife}, 4, 2015.

\bibitem[Hubel and Wiesel(1977)]{Hubel1977FerrierL}
David~H. Hubel and Torsten~N. Wiesel.
\newblock Ferrier lecture - functional architecture of macaque monkey visual
  cortex.
\newblock \emph{Proceedings of the Royal Society of London. Series B.
  Biological Sciences}, 198:\penalty0 1 -- 59, 1977.

\bibitem[Niell et~al.(2008)Niell, Stryker, and Keck]{Niell2008HighlySR}
Cristopher~M. Niell, Michael~P. Stryker, and Wendell~M. Keck.
\newblock Highly selective receptive fields in mouse visual cortex.
\newblock \emph{The Journal of Neuroscience}, 28:\penalty0 7520 -- 7536, 2008.

\bibitem[Ringach et~al.(2016)Ringach, Mineault, Tring, Olivas,
  Garc{\'i}a-Junco-Clemente, and Trachtenberg]{Ringach2016SpatialCO}
Dario~L. Ringach, Patrick~J. Mineault, Elaine Tring, Nicholas~D. Olivas, Pablo
  Garc{\'i}a-Junco-Clemente, and Joshua~T. Trachtenberg.
\newblock Spatial clustering of tuning in mouse primary visual cortex.
\newblock \emph{Nature Communications}, 7, 2016.

\bibitem[de~Vries et~al.(2020)de~Vries, Lecoq, Buice, Groblewski, Ocker,
  Oliver, Feng, Cain, Ledochowitsch, Millman, et~al.]{de2020large}
Saskia~EJ de~Vries, Jerome~A Lecoq, Michael~A Buice, Peter~A Groblewski,
  Gabriel~K Ocker, Michael Oliver, David Feng, Nicholas Cain, Peter
  Ledochowitsch, Daniel Millman, et~al.
\newblock A large-scale standardized physiological survey reveals functional
  organization of the mouse visual cortex.
\newblock \emph{Nature Neuroscience}, 23\penalty0 (1):\penalty0 138--151, 2020.

\bibitem[Siegle et~al.(2021)Siegle, Jia, Durand, Gale, Bennett, Graddis,
  Heller, Ramirez, Choi, Luviano, Groblewski, Ahmed, Arkhipov, Bernard, Billeh,
  Brown, Buice, Cain, Caldejon, Casal, Cho, Chvilicek, Cox, Dai, Denman,
  de~Vries, Dietzman, Esposito, Farrell, Feng, Galbraith, Garrett, Gelfand,
  Hancock, Harris, Howard, Hu, Hytnen, Iyer, Jessett, Johnson, Kato, Kiggins,
  Lambert, Lecoq, Ledochowitsch, Lee, Leon, Li, Liang, Long, Mace, Melchior,
  Millman, Mollenkopf, Nayan, Ng, Ngo, Nguyen, Nicovich, North, Ocker,
  Ollerenshaw, Oliver, Pachitariu, Perkins, Reding, Reid, Robertson,
  Ronellenfitch, Seid, Slaughterbeck, Stoecklin, Sullivan, Sutton, Swapp,
  Thompson, Turner, Wakeman, Whitesell, Williams, Williford, Young, Zeng,
  Naylor, Phillips, Reid, Mihalas, Olsen, and Koch]{Siegle2021SurveyOS}
Joshua~H. Siegle, Xiaoxuan Jia, S{\'e}verine Durand, Samuel~D. Gale, Corbett
  Bennett, Nile Graddis, Greggory Heller, Tamina Ramirez, Hannah Choi,
  Jennifer~A. Luviano, Peter~A. Groblewski, Ruweida Ahmed, Anton Arkhipov, Amy
  Bernard, Yazan~N. Billeh, Dillan Brown, Michael~A. Buice, Nicolas Cain,
  Shiella Caldejon, Linzy Casal, Andrew Cho, Maggie Chvilicek, Timothy~C Cox,
  Kael Dai, Daniel~J Denman, Saskia E.~J. de~Vries, Roald Dietzman, Luke
  Esposito, Colin Farrell, David Feng, J.~Galbraith, Marina Garrett, Emily~C.
  Gelfand, Nicole Hancock, Julie~A. Harris, Robert~E. Howard, Brian Hu, Ross
  Hytnen, Ramakrishnan Iyer, Erika Jessett, Katelyn Johnson, India Kato, Justin
  Kiggins, Sophie Lambert, J{\'e}r{\^o}me~A. Lecoq, Peter Ledochowitsch,
  Jung~Hoon Lee, Arielle Leon, Yang Li, Elizabeth Liang, Fuhui Long, Kyla Mace,
  Josef Melchior, Daniel~J. Millman, Tyler Mollenkopf, Chelsea Nayan, Lydia Ng,
  Kiet Ngo, Thuyahn Nguyen, Philip~R. Nicovich, Kat North, Gabriel~Koch Ocker,
  Douglas~R. Ollerenshaw, Michael Oliver, Marius Pachitariu, Jed Perkins,
  Melissa Reding, David Reid, Miranda Robertson, Kara Ronellenfitch, Sam Seid,
  Cliff Slaughterbeck, Michelle Stoecklin, David Sullivan, Ben~B. Sutton,
  Jackie Swapp, Carol~L. Thompson, Kristen Turner, Wayne Wakeman, Jennifer~D.
  Whitesell, Derric Williams, Ali Williford, R.~D. Young, Hongkui Zeng,
  Sarah~R. Naylor, John~W. Phillips, R.~Clay Reid, Stefan Mihalas, Shawn~R.
  Olsen, and Christof Koch.
\newblock Survey of spiking in the mouse visual system reveals functional
  hierarchy.
\newblock \emph{Nature}, 2021.

\bibitem[Caron et~al.(2021)Caron, Touvron, Misra, J{\'e}gou, Mairal,
  Bojanowski, and Joulin]{Caron2021EmergingPI}
Mathilde Caron, Hugo Touvron, Ishan Misra, Herv{\'e} J{\'e}gou, Julien Mairal,
  Piotr Bojanowski, and Armand Joulin.
\newblock Emerging properties in self-supervised vision transformers.
\newblock In \emph{Proceedings of the IEEE/CVF International Conference on
  Computer Vision}, pages 9650--9660, 2021.

\bibitem[Esfahany et~al.(2018)Esfahany, Siergiej, Zhao, and
  Park]{Esfahany2018OrganizationON}
Kathleen Esfahany, Isabel Siergiej, Yuan Zhao, and Il~Memming Park.
\newblock Organization of neural population code in mouse visual system.
\newblock \emph{eNeuro}, 5, 2018.

\bibitem[Jin and Glickfeld(2020)]{Jin2020MouseHV}
Miaomiao Jin and Lindsey~L. Glickfeld.
\newblock Mouse higher visual areas provide both distributed and specialized
  contributions to visually guided behaviors.
\newblock \emph{Current Biology}, 30:\penalty0 4682--4692.e7, 2020.

\bibitem[Dinh et~al.(2016)Dinh, Sohl-Dickstein, and Bengio]{dinh2016density}
Laurent Dinh, Jascha Sohl-Dickstein, and Samy Bengio.
\newblock Density estimation using real nvp.
\newblock \emph{arXiv preprint arXiv:1605.08803}, 2016.

\bibitem[Pei et~al.(2021)Pei, Ye, Zoltowski, Wu, Chowdhury, Sohn, O'Doherty,
  Shenoy, Kaufman, Churchland, et~al.]{pei2021neural}
Felix Pei, Joel Ye, David Zoltowski, Anqi Wu, Raeed~H Chowdhury, Hansem Sohn,
  Joseph~E O'Doherty, Krishna~V Shenoy, Matthew~T Kaufman, Mark Churchland,
  et~al.
\newblock Neural latents benchmark'21: Evaluating latent variable models of
  neural population activity.
\newblock \emph{arXiv preprint arXiv:2109.04463}, 2021.

\bibitem[Deitch et~al.(2021)Deitch, Rubin, and Ziv]{deitch2021representational}
Daniel Deitch, Alon Rubin, and Yaniv Ziv.
\newblock Representational drift in the mouse visual cortex.
\newblock \emph{Current Biology}, 31\penalty0 (19):\penalty0 4327--4339, 2021.

\bibitem[Wang and Isola(2020)]{wang2020understanding}
Tongzhou Wang and Phillip Isola.
\newblock Understanding contrastive representation learning through alignment
  and uniformity on the hypersphere.
\newblock In \emph{International Conference on Machine Learning}, pages
  9929--9939. PMLR, 2020.

\bibitem[Hendrycks and Gimpel(2016)]{hendrycks2016gelu}
Dan Hendrycks and Kevin Gimpel.
\newblock Gaussian error linear units (gelus).
\newblock \emph{arXiv preprint arXiv:1606.08415}, 2016.

\bibitem[Paszke et~al.(2019)Paszke, Gross, Massa, Lerer, Bradbury, Chanan,
  Killeen, Lin, Gimelshein, Antiga, Desmaison, Kopf, Yang, DeVito, Raison,
  Tejani, Chilamkurthy, Steiner, Fang, Bai, and Chintala]{NEURIPS2019_9015}
Adam Paszke, Sam Gross, Francisco Massa, Adam Lerer, James Bradbury, Gregory
  Chanan, Trevor Killeen, Zeming Lin, Natalia Gimelshein, Luca Antiga, Alban
  Desmaison, Andreas Kopf, Edward Yang, Zachary DeVito, Martin Raison, Alykhan
  Tejani, Sasank Chilamkurthy, Benoit Steiner, Lu~Fang, Junjie Bai, and Soumith
  Chintala.
\newblock Pytorch: An imperative style, high-performance deep learning library.
\newblock In H.~Wallach, H.~Larochelle, A.~Beygelzimer, F.~d\textquotesingle
  Alch\'{e}-Buc, E.~Fox, and R.~Garnett, editors, \emph{Advances in Neural
  Information Processing Systems 32}, pages 8024--8035. Curran Associates,
  Inc., 2019.

\bibitem[Walt et~al.(2011)Walt, Colbert, and Varoquaux]{walt2011numpy}
St{\'e}fan van~der Walt, S~Chris Colbert, and Gael Varoquaux.
\newblock The numpy array: a structure for efficient numerical computation.
\newblock \emph{Computing in Science \& Engineering}, 13\penalty0 (2):\penalty0
  22--30, 2011.

\bibitem[Pedregosa et~al.(2011)Pedregosa, Varoquaux, Gramfort, Michel, Thirion,
  Grisel, Blondel, Prettenhofer, Weiss, Dubourg, Vanderplas, Passos,
  Cournapeau, Brucher, Perrot, and Duchesnay]{scikit-learn}
F.~Pedregosa, G.~Varoquaux, A.~Gramfort, V.~Michel, B.~Thirion, O.~Grisel,
  M.~Blondel, P.~Prettenhofer, R.~Weiss, V.~Dubourg, J.~Vanderplas, A.~Passos,
  D.~Cournapeau, M.~Brucher, M.~Perrot, and E.~Duchesnay.
\newblock Scikit-learn: Machine learning in {P}ython.
\newblock \emph{Journal of Machine Learning Research}, 12:\penalty0 2825--2830,
  2011.

\bibitem[Poli{\v c}ar et~al.(2019)Poli{\v c}ar, Stra{\v z}ar, and
  Zupan]{PolicarOPENTSNE}
Pavlin~G. Poli{\v c}ar, Martin Stra{\v z}ar, and Bla{\v z} Zupan.
\newblock opentsne: a modular python library for t-sne dimensionality reduction
  and embedding.
\newblock \emph{bioRxiv}, 2019.
\newblock \doi{10.1101/731877}.

\bibitem[Kobak and Linderman(2021)]{Kobak2021tsnepca}
Dmitry Kobak and George~C. Linderman.
\newblock Initialization is critical for preserving global data structure in
  both t-sne and umap.
\newblock \emph{Nature Biotechnology}, 39\penalty0 (2):\penalty0 156--157, Feb
  2021.
\newblock ISSN 1546-1696.
\newblock \doi{10.1038/s41587-020-00809-z}.

\bibitem[Tralie et~al.(2018{\natexlab{a}})Tralie, Saul, and
  Bar-On]{ctralie2018ripser}
Christopher Tralie, Nathaniel Saul, and Rann Bar-On.
\newblock {Ripser.py}: A lean persistent homology library for python.
\newblock \emph{The Journal of Open Source Software}, 3\penalty0 (29):\penalty0
  925, Sep 2018{\natexlab{a}}.
\newblock \doi{10.21105/joss.00925}.

\bibitem[Tralie et~al.(2018{\natexlab{b}})Tralie, Mease, and J.Perea]{dreimac}
C.~Tralie, T.~Mease, and J.Perea.
\newblock Dreimac: Dimension reduction with eilenberg-maclane coordinates.
\newblock \emph{GitHub}, 2018{\natexlab{b}}.

\bibitem[Kobak et~al.(2016)Kobak, Brendel, Constantinidis, Feierstein, Kepecs,
  Mainen, Qi, Romo, Uchida, and Machens]{Kobak2016DemixedPC}
Dmitry Kobak, Wieland Brendel, Christos Constantinidis, Claudia~E Feierstein,
  Adam Kepecs, Zachary~F. Mainen, Xue-Lian Qi, Ranulfo Romo, Naoshige Uchida,
  and Christian~K. Machens.
\newblock Demixed principal component analysis of neural population data.
\newblock \emph{eLife}, 5, 2016.

\bibitem[Gao et~al.(2016)Gao, Archer, Paninski, and
  Cunningham]{Gao2016LinearDN}
Yuanjun Gao, Evan Archer, Liam Paninski, and John~P. Cunningham.
\newblock Linear dynamical neural population models through nonlinear
  embeddings.
\newblock In \emph{NIPS}, 2016.

\end{thebibliography}

\balance

\vspace{15pt} 
\textbf{Acknowledgments:} The authors thank Matthias Bethge, Roland S. Zimmermann, Luisa Eck, Alexander Mathis, Dylan Paiton, Jakob Macke, Dmitry Kobak, Jessy Lauer, Rodrigo González, and Gary Kane for discussions and feedback on earlier versions of the manuscript or code, and the T\"ubingen AI Center for computing resources.
Funding was provided by SNSF grant no. 310030\_201057, a Novartis Foundation for Medical-Biological Research Young Investigator Grant to MWM; Google PhD Fellowship to StS; the German Academic Exchange Service (DAAD) to JHL. StS acknowledges the IMPRS-IS T\"ubingen and ELLIS PhD program, and JHL thanks the TUM Program in Neuroengineering. MWM is the Bertarelli Foundation Chair of Integrative Neuroscience.
\medskip

\textbf{Author contributions:} 
Conceptualization: MWM, StS;
Methodology: StS, JHL, MWM; 
Software: StS, JHL; 
Theory: StS;
Formal analysis: StS, JHL; 
Investigation:  StS, JHL;
Data Curation: JHL, StS;
Writing-Original Draft: MWM; 
Writing-Editing: MWM, StS, JHL.
\medskip

\textbf{Conflicts:} StS and MWM have filed a patent pertaining to this work. The authors declare no additional conflicts of interest. The funders had no role in the conceptualization, design, data collection, analysis, decision to publish, or preparation of the manuscript.

\section*{Methods}
\small

\subsection*{Datasets}

\subsubsection*{Artificial Spiking Dataset} 

    Synthetic spiking data for benchmarking in Fig.~\ref{fig:CEBRAvsAll} was adopted from~\cite{zhou2020learning}. The continuous 1D behavior variable $c \in [0, 2\pi)$ was sampled uniformly in the interval $[0, 2\pi)$. The true 2D latent variable $\zz \in \R^2$ was then sampled from a Gaussian distribution $\mathcal{N} ( \mu(c), \Sigma(c)) $ with mean $\mu(c) = (c, 2\sin c)^\top$ and covariance $\Sigma(c) = \mathrm{diag} (0.6-0.3|\sin{c}|, 0.3|\sin{c}|)$. 
    After sampling, the 2D latent variable $\zz$ was mapped to spiking rates of 100 neurons by applying four randomly initialized RealNVP \cite{dinh2016density} blocks. Poisson noise was then applied \cite{zhou2020learning} to map firing rates onto spike counts.
    The final dataset consisted of $1.5\times10^{4}$ data points, and was split into train (80\%) and validation (20\%) sets. We quantified consistency across the entire dataset.
    The additional synthetic data, used in Extended Data Fig. 1, was generated by varying the noise distribution in the above generative process. Beside Poisson noise, we used additive truncated ($[0, 1000]$) Gaussian noise with standard deviation 1 and additive uniform noise defined in $[0,2)$ which was applied to the spiking rate.
    We also adapted the Poisson spiking by simulating neurons with a refractory period. For this, we scaled the spiking rates to an average rate of 110Hz. We sample inter-spike intervals from an exponential distribution with the given rate and add a refractory period of 10ms. 

\subsubsection*{Rat Hippocampus Dataset}

    We used the dataset presented in \citet{grosmark2016diversity}. In brief, bilaterally implanted silicon-probes recorded multi-cellular electrophysiological data from the CA1 hippocampus areas from each of four male Long-Evans rats. During a given session, each rat independently ran on a 1.6 meter long linear track, where they were rewarded with water at each end of the track. The numbers of recorded putative pyramidal neurons for each rat ranged between 48 to 120. Here, we processed the data as in \cite{zhou2020learning}. Specifically, the spikes were binned into 25ms time windows. %
    The position and running direction (left or right) of the rat was encoded into a 3D vector, which consisted of the continuous position value and two binary values indicating right or left direction. Recordings from each rat was parsed into trials (a round trip from one end of the track as a trial) and then split into a train, validation, and test set with a k=3 nested cross-validation scheme for the decoding task.

\subsubsection*{Macaque Dataset}

    We used the dataset presented in \citet{chowdhury2020area}. In brief, electrophysiological recordings were performed in Area 2 of somatosensory cortex (S1) in a rhesus macaque (monkey) during a center-out reaching task with a manipulandum. Specifically, the monkey performed an eight direction reaching task where on 50\% of trials they actively made center-out movements to a presented target. The remaining trials were ``passive'' trials, where an unexpected 2N force bump was given to the manipulandum towards one of the eight target directions during a holding period. The trials were aligned as in~\cite{pei2021neural}, and we used the data from -100ms and 500ms from the movement onset. We used 1ms time bins and convolved the data with a Gaussian kernel with standard deviation of 40ms.

\subsubsection*{Mouse Visual Cortex Datasets}
    
    We utilized the Allen Institute 2-photon calcium imaging and Neuropixels data recorded from five mouse visual cortical areas (VISp, VISl, VISal, VISam, VISpm) and one posterior parietal cortex (PPC)-like area (VISrl) during presentation of a black-and-white movie with 30 Hz frame rate, as presented previously~\cite{deitch2021representational, de2020large, Siegle2021SurveyOS}.
    For calcium imaging (2P), we used the processed dataset by \citet{de2020large} with a sampling rate of 30 Hz, aligned to the video frames. We considered the recordings from excitatory neurons (Emx1-IRES-Cre, Slc17a7-IRES2-Cre, Cux2-CreERT2, Rorb-IRES2-Cre, Scnn1a-Tg3-Cre, Nr5a1-Cre, Rbp4-Cre\_KL100, Fezf2-CreER, Tlx3-Cre\_PL56) in the ``Visual Coding-2P'' dataset.
    Ten repeats of the first movie (Movie 1) were shown in all session types (A,B,C) for each mouse and we used the neurons that were recorded in all three session types, found by using the cell registration~\cite{de2020large}.
    The Neuropixels recordings were obtained from the ``Brain Observatory 1.1'' dataset~\cite{Siegle2021SurveyOS}.
    We used the pre-processed spike-timings and binned them to a sampling frequency of 120 Hz, aligned with the movie timestamps (i.e., exactly 4 bins are aligned with each frame).
    The dataset contains recordings for 10 repeats, and we used the same Movie 1 that was used for the 2P recordings.
    For the analysis of consistency across the visual and PPC cortical areas, we used a disjoint set of neurons for each seed to avoid higher intra-consistency due to overlapping neuron identities. We made 3 disjoint set of neurons by only considering neurons from session A (for 2P data) and non-overlapping random sampling for each seed.

\subsection*{\cebra \space Model Framework}

\subsubsection*{Notation}

    We will use $\xx, \yy$ as general placeholder variables, and denote the multidimensional, time-varying signal as $\ss_t$, parameterized by the time $t$. The multidimensional, continuous context variable $\cc_t$ contains additional information about the experimental condition and additional recordings, similar to the discrete categorical variable $k_t$.
    \medskip
    
    The exact composition of $\ss$, $\cc$ and $k$ depends on the experimental context. CEBRA is agnostic to the exact signal types; with the default parameterizations, $\ss_t$ and $\cc_t$ can have up to an order of hundred or thousand dimensions. For even higher dimensional datasets (e.g. raw video, audio, ...) other optimized deep learning tools can be used for feature extraction prior to the application of CEBRA.

\subsubsection*{Applicable problem setup}

    We refer to $\xx \in \cX$ as the \emph{reference} sample, and to $\yy \in \cY$ as a corresponding \emph{positive} or \emph{negative} sample.
    Together, $(\xx, \yy)$ form a positive or negative pair, based on the distribution $\yy$ is sampled from.
    We denote the distribution and density function of $\xx$ as $p(\xx)$, the conditional distribution and density of the positive sample $\yy$ given $\xx$ as $p(\xx | \yy)$ and the conditional distribution and density of the negative sample $\yy$ given $\xx$ as $q(\yy | \xx)$.
    \medskip
    
    After sampling---and no matter whether we are considering a positive or negative pair---both samples $\xx \in \Rx$ and $\yy \in \Ry$ are encoded by feature extractors $\ffx: \cX \mapsto \cZ$ and $\ffy: \cY \mapsto \cZ$. The feature extractors map both samples from signal space $\cX \subseteq \Rx, \cY \subseteq \Ry$ into a common embedding space $\cZ \subseteq \RE$. The design and parameterization of the feature extractor is chosen by the user of the algorithm.
    Note that the spaces $\cX$ and $\cY$ and their corresponding feature extractors can be the same (which is the case for single-session experiments in this work), but that this is not a strict requirement within the CEBRA framework (e.g., in multi-session training across animals or modalities, $\cX$ and $\cY$ are selected to be signals from different mice or modalities, respectively). It is also possible to include the context variable (e.g., behavior) into $\cX$, or it is possible to set $\xx$ to the context variable, and $\yy$ to the signal variable.
    \medskip

    Given two encoded samples, a similarity measure $\phi: \cZ \times \cZ \mapsto \R$ assigns a score to a pair of embeddings. The similarity measure needs to assign a higher score to more similar pairs of points, and have an upper bound. For this work, we consider the dot product between normalized feature vectors, $\phi(\zz, \zz') = \zz^\top \zz' / \tau$, in most analyses (latents on a hypersphere), or the negative mean squared error, $\phi(\zz, \zz') = - \|\zz - \zz'\|^2 / \tau$ (latents in Euclidean space). Both metrics can be scaled by a temperature parameter $\tau$ which is either fixed, or jointly learned with the network. Other $L_p$ norms and other similarity metrics, or even a trainable neural network (a so-called projection head commonly used in contrastive learning algorithms, cf. \citet{Hyvrinen2019NonlinearIU,Chen2020ASF}), are possible choices within the CEBRA software package. The exact choice of $\phi$ shapes the properties of the embedding space, and encodes assumptions about the distributions $p$ and $q$.
    \medskip
    
    The technique requires paired data recordings, e.g. as common in aligned time-series. The signal $\ss_t$, continuous context $\cc_t$ and discrete context $k_t$ are synced in their time-point $t$.
    How the reference, positive and negative samples are constructed from these available signals is a configuration choice made by the algorithm user, and depends on the scientific question to investigate.

\subsubsection*{Optimization}\label{sec:optimization}

    Given the feature encoders $\ffx$ and $\ffy$ for the different sample types, as well as the similarity measure $\phi$, we introduce the shorthand $\psi(\xx, \yy) = \phi(\ffx(\xx), \ffy(\yy))$.
    The objective function can then be compactly written as:
    \begin{equation}
        \label{eq:objective-asympt}
        \cint{\xx \in \cX} \mathrm{d}\xx p(\xx)\!\!
            \left[ 
                \text{-}\cint{\yy\in \cY}\mathrm{d}\yy p(\yy | \xx) \psi(\xx, \yy)
                + \log \cint{\yy\in \cY} \mathrm{d}\yy q(\yy | \xx) e^{\psi(\xx, \yy)}
            \right].
    \end{equation}
    
    We approximate this objective \citep{Zimmermann2021ContrastiveLI,wang2020understanding} by drawing a single positive example $\yy_+$, and multiple \emph{negative examples} $\yy_i$ from the distributions outlined above, and minimize the loss function
    \begin{equation}
        \label{eq:objective-samples}
        \mathop{\mathbb{E}}\limits_{{\substack{
            \xx \sim p(\xx),\ 
            \yy_+ \sim p(\yy|\xx)\\
            \yy_1,\dots,\yy_n \sim q(\yy|\xx)\\
        }}}
        \left[ - \psi(\xx, \yy_+) + \log \sum_{i=1}^{n} e^{\psi(\xx, \yy_i)}
        \right],
    \end{equation}
    with a gradient-based optimization algorithm. The number of negative samples is a hyperparameter of the algorithm and larger batch sizes are generally preferable.
    \medskip

    For sufficiently small datasets as used in this paper, both positive and negative samples are drawn from all available samples in the dataset.
    This is in contrast to the common practice in many contrastive learning frameworks, where a mini-batch of samples is drawn first, which are then grouped into positive and negative pairs.
    Allowing to access the whole dataset to form the pairs gives a better approximation of the respective distributions $p(\yy | \xx)$ and $q(\yy | \xx)$, and considerably improves the quality of the obtained embeddings.
    If the dataset is small enough to fit into the GPU memory, CEBRA can be optimized with batch gradient descent, i.e., use the whole dataset at each optimizer step.

\subsubsection*{Goodness of fit}

    Comparing the loss value---at both the absolute value and relative value across models at the same point in training time---can be used to determine the goodness of fit. Practically, this means one can find which hypothesis best fits one's data, in the case of using CEBRA-Behavior. Specifically, let us denote the objective in Eq.~\ref{eq:objective-asympt} as $L_\text{asympt}$ and its approximation in Eq.~\ref{eq:objective-samples} with a batch size of $n$ as $L_{n}$.
    In the limit of many samples, the objective converge up to a constant, $L_\text{asympt} = \lim_{n \to \infty} [L_{n} - \log n]$ (cf. Suppl. Note 2, and \citet{wang2020understanding}).
    \medskip
    
    The objective has also two trivial solutions: The first one is obtained for a constant $\psi(\xx, \yy) = \psi$, which yields a value of $L_n = \log n$. Such a solution is typically not obtained during training, as the network is initialized randomly, causing the initial embedding points to be randomly distributed in space.
    \medskip
    
    If the embedding points are distributed uniformly in space, and $\phi$ is selected such that $\mathbb{E}[\phi(\xx, \yy)] = 0$, we will also get a value that is \emph{approximately} $L_n = \log n$.
    The value can be estimated easily by computing $\phi(\uu, \vv)$ for randomly distributed points.
    \medskip
    
    The minimizer of Eq.~\ref{eq:objective-asympt} is also clearly defined as $D_\text{KL}(p \| q)$ and depends on the positive and negative distribution. For discovery-driven (time contrastive) learning, this value is impossible to estimate as it would require access to the underlying conditional distribution of the latents. However, for hybrid training with pre-defined positive and negative distributions, this quantity can be again numerically estimated.
    \medskip
    
    Interesting values of the loss function when fitting a CEBRA model are therefore
    \begin{equation}
        - D_\text{KL}(p \| q) \le L_n - \log n \le 0
    \end{equation}
    where $L_n - \log n$ is the goodness of fit (lower is better) of the CEBRA model.
    Note that the metric is independent of the batch size used for training.

\subsubsection*{Sampling}\label{sec:sampling}

    Selection of the sampling scheme is CEBRA's key feature to adapt embedding spaces to different datasets and recording setups.
    The conditional distributions $p(\yy|\xx)$ for positive samples and $q(\yy|\xx)$ for negative samples as well as the marginal distribution $p(\xx)$ for reference samples are specified by the user.
    CEBRA offers a set of pre-defined sampling techniques, but customized variants can be specified to implement additional, domain specific distributions.
    This form of training allows to use the context variables to shape the properties of the embedding space, as outlined in the following graphical model:
    \medskip
    
    \begin{center}
    \includegraphics[width=0.75\linewidth]{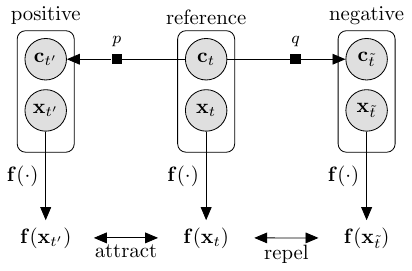}
    \end{center}

    Through the choice of sampling technique, various use cases can be built into the algorithm: For instance, by forcing the positive and negative distributions to sample uniform across a factor, the model will become invariant to this factor, as including it would yield in a sub-optimal value of the objective function.
    \medskip

    When considering different sampling mechanisms, we distinguish between \emph{single-session} and \emph{multi-session} datasets:
    A single-session dataset consists of samples $\ss_t$, which are associated to one or more context variables $\cc_t$ and/or $k_t$. These context variables allow to impose structure on the marginal and conditional distribution used for obtaining the embedding.
    Multi-session datasets consist of multiple single-session datasets. The dimension of context variables $\cc_t$ and/or $k_t$ must be shared across all sessions, while the dimension of the signal $\ss_t$ can vary.
    In such a setting, CEBRA allows to learn a shared embedding space for signals from all sessions.
    \medskip
    
    For single-session datasets, sampling is done in two steps:
    First, based on a specified ``index'' (the user-defined context variable $\cc_t$ and/or $k_t$), locations $t$ are sampled for reference, positive and negative samples.
    The algorithm differentiates between categorical ($k$) and continuous ($\cc$) variables for this purpose.
    \medskip
    
    In the simplest case, negative sampling ($q$) returns a random sample from the empirical distribution, by returning a randomly chosen index from the dataset. Optionally, with a categorical context variable $k_t \in [K]$, negative sampling can be performed to approximate a uniform distribution of samples over this context variable. If this is performed for both the negative and positive samples, the resulting embedding will become invariant with respect to the variable $k_t$. Sampling is performed in this case by computing the cumulative histogram of $k_t$, and sampling uniformly over $k$ using the transformation theory for probability densities.
    \medskip
    
    For positive pairs, different options exist based on the availability of continuous and discrete context variables.
    For a discrete context variable $k_t \in [K]$ with $K$ possible values, sampling from the conditional distribution is done by filtering the whole dataset for the value $k_t$ of the reference sample, and uniformly selecting a positive sample with the same value.
    For a continuous context variable $\cc_t$, we use a set of time offsets $\Delta$ to specify the distribution.
    Given the time offsets, the empirical distribution $P(\cc_{t+\tau} | \cc_t)$ for a particular choice of $\tau \in \Delta$ can be computed from the dataset: We build up a set $D = \{ t\in[T], \tau\in\Delta: \cc_t - \cc_{t+\tau}\}$, sample a $\dd$ uniformly from $D$, and obtain the sample that is closest to the reference sample modified by this distance $\dd$ from the dataset ($\xx + \dd$).
    It is possible to combine a continuous variable $\cc_t$ with a categorical variable $k_t$ for mixed sampling. On top of the continual sampling step above, it is ensured that both samples in the positive pair share the same value $k_t$. 
    \medskip
    
    It is crucial that the context samples $\cc$ and the norm used in the algorithm match in some way; for simple context variables with predictable conditional distributions (e.g., a one or two-dimensional position of a moving animal, which can be most likely well described by a Gaussian conditional distribution based on the previous sample).
    An additional alternative is to use CEBRA also to \emph{pre-process} the original context samples $\cc$ and use the embedded context samples with the metric used for CEBRA training.
    This scheme is especially useful for higher dimensional behavioral data, or even complex inputs like video.
    \medskip

    We next consider the multi-session case, where signals $\ss^{(i)}_t \in \R^{n_i}$ come from $N$ different sessions $i \in [N]$ with session-dependent dimensionality $n_i$.
    Importantly, the corresponding continuous context variables $\cc^{(i)}_t \in \R^{m}$ share the same dimensionality $m$, which makes it possible to relate samples across sessions.
    The multi-session setup is similar to mixed session sampling (if we treat the session ID as a categorical variable $k^{(i)}_t := i$ for all time steps $t$ in session $i$). The conditional distribution for both negative and positive pairs is uniformly sampled across sessions, irrespective of session length.
    Multi session mixed sampling or multi session discrete sampling can be implemented analogously.
    \medskip

    Besides the outlined sampling scheme, CEBRA is flexible to incorporate more specialized sampling schemes.
    For instance, mixed single session sampling could be extended to additionally incorporate a dimension the algorithm should become invariant to. This would add an additional step of uniform sampling with regard to to this desired discrete variable (e.g., via ancestral sampling).
    \medskip

\subsubsection*{Choice of reference and positive and negative samples}

    Depending on the exact application, the contrastive learning step can be performed by explicitly including or excluding the context variable:
    The reference sample $\xx$ can contain information from the signal $\ss_t$, but also from the experimental conditions, behavioral recordings, or other context variables. The positive and negative samples $\yy$ are set to the signal variable $\ss_t$.

\subsubsection*{Theoretical guarantees for linear identifiability of CEBRA models}

    Identifiability describes the property of an algorithm to give a consistent estimate for the model parameters given that the data distributions match. We here apply the relaxed notion of \emph{linear identifiability} that was previously discussed and used by~\citet{Hyvrinen2019NonlinearIU} and \citet{Roeder2020}: After training two encoder models $\ff$ and $\ff'$, the models are linear identifiable if $\ff(\xx) = \LL \ff(\xx)$ where $\LL$ is a linear projection.
    \medskip

    When applying CEBRA, three cases are of potential interest.
    First, when applying discovery-driven CEBRA, will two models estimated on comparable experimental data agree in their inferred representation?
    Second, under which assumptions about the data will we be able to discover the \emph{true} latent distribution?
    Third, in the hypothesis-driven or hybrid application of CEBRA, is the algorithm guaranteed to give a meaningful (non-standard) latent space when we can find signal within the data?
  \medskip
    
    For the first case, we note that the CEBRA objective with a cosine similarity metric follows the canonical discriminative form for which \citet{Roeder2020} showed linear identifiability: For sufficiently diverse datasets, two CEBRA models trained to convergence on the same dataset will be consistent up to linear transformations. Note that consistency of CEBRA is independent of the exact data distribution: The diversity condition merely requires that for any set of samples $\{\yy_1,\dots,\yy_d\}$ from the negative distribution $q(\cdot | \xx)$, the matrices $[\cdots \ffy(\yy_i) - \ffy(\yy_i) \cdots]_{i=1}^{d}$ and the matrix $[\cdots \ffx(\xx_i) \cdots]_{i=1}^{d+1}$ are invertible (i.e., the embeddings are sufficiently diverse).
    Alternatively, we can derive linear identifiability from assumptions about the data distribution: If the ground truth latents are sufficiently diverse (i.e., vary in all directions under the distributions $p$ and $q$), and the model is sufficiently parameterized to fit the data, we will also obtain consistency up to a linear transformation.
    See Suppl. Note 2 for a full formal discussion and proofs.
    \medskip

    For the second case, additional assumptions are required regarding the exact form of the data generating distribution. Within the scope of this work, we consider ground truth latents distributed on the hypersphere or Euclidean space. 
    The metric then needs to match assumptions about the variation of the ground truth latents over time.
    In discovery-driven CEBRA, using the dot product as the similarity measure then encodes the assumption that latents vary according to a von-Mises-Fisher distribution, while the mean squared error encodes an assumption that latents vary according to a Normal distribution. More broadly, if we assume that the latents have a uniform marginal distribution (which can be ensured by designing un-biased experiments), the similarity measure should be chosen as the log-likelihood of the conditional distribution over time.
    In this case, CEBRA identifies the data generating distribution up to affine transforms (in the most general case).
    \medskip

    This result also explains the empirically high performance of CEBRA for decoding applications: If trained for decoding (using the variable to decode for informing the conditional distribution), it is trivial to select matching conditional distributions, as both quantities are directly selected by the user. CEBRA then ``identifies'' the context variable up to a linear transformation.
\medskip

    For the third case, we are interested in the hypothesis testing capabilities. We can show that if a mapping exists between the context variable and the signal space, CEBRA will recover this relationship and yield a meaningful embedding, which is also decodable. However, if such a mapping does not exist, we can show that CEBRA will instead learn a default embedding which is the uniform distribution of points on the hypersphere.

\subsection*{CEBRA models}\label{sec:model-implementation}

    \justify We ran all experiments using our PyTorch implementation of \cebra. 
    We chose $X = Y$ to be the neural signal with varying amounts of recorded neurons and channels based on the dataset.
    We used three types of encoder models based on the required receptive field; a receptive field of one sample was used on the synthetic dataset experiments (Fig.~\ref{fig:CEBRAvsAll}b), a receptive field of 10 samples in all other experiments (rat, monkey, mouse) except for the Neuropixels dataset, where a receptive field of 40 samples is used due to the 4 times higher sampling rate of the dataset.
    \medskip
    
    All feature encoders are parameterized by the number of neurons (input dimension), a hidden dimension to control the model size and capacity, as well as their output (embedding) dimension.
    For the model with the receptive field of one, a four layer MLP was used. The first and second layers map their respective inputs to the hidden dimension, while the third layer introduces a bottleneck and maps to half the hidden dimension. The final layer maps to the requested output dimension.
    For the model with receptive field of 10, a convolutional network with five time convolutional layers was used. The first layer had kernel size 2, the next three layers had kernel size 3 and used skip connections. The final layer had kernel size 3 and mapped the hidden dimensions to the output dimension.
    For the model with receptive field 40, we first preprocessed the signal by concatenating a $2\times$ downsampled version of the signal with a learnable downsample operation implemented as a convolutional layer with kernel size 4 and stride 2, directly followed (without activation function in between) by another convolutional layer with kernel size 3 and stride 2. After these first layers, the signal is subsampled by a factor of 4.
    Afterwards, similar to the receptive field 10 model, we apply three layers with kernel size 3 and skip connections, and a final layer with kernel size 3.
    In all models, Gaussian error linear unit activation functions (GELU; \citealp{hendrycks2016gelu}) were applied after each layer except the last. The feature vector was normalized after the last layer, unless a mean squared error (MSE) based similarity metric was used (as in Extended Data Fig~\ref{SupplReaching}).
    \medskip
    
    Our implementation of the InfoNCE criterion received a mini-batch (or the full dataset) of size $n \times d$ for each of the reference, positive, and negative samples. $n$ dot-product similarities are computed between reference and positive samples, $n\times n$ dot-product similarities are computed between reference and negative samples. The similarities were scaled with the inverse of the temperature parameter $\tau$.
    \begin{minted}[autogobble,escapeinside=||]{python}
        from torch import einsum, logsumexp, no_grad
        
        def info_nce(ref, pos, neg, |$\tau$| = 1.0):
          pos_dist = einsum("nd,nd->n",  ref, pos) / |$\tau$|
          neg_dist = einsum("nd,md->nm", ref, neg) / |$\tau$|
          with no_grad():
            c, _ = neg_dist.max(dim=1)
          pos_dist = pos_dist - c.detach()
          neg_dist = neg_dist - c.detach()
          pos_loss = -pos_dist.mean()
          neg_loss = logsumexp(neg_dist, dim=1).mean()
          return pos_loss + neg_loss
    \end{minted}
    Alternatively, a learnable temperature can be used. For a numerically stable implementation, we store the log inverse temperature $\alpha = - \log \tau$ as a parameter of the loss function. At each step, we scale the distances in the loss function with $\min(\exp \alpha, 1/\tau_\mathrm{min})$. The additional parameter $\tau_\mathrm{min}$ is a lower bound on the temperature. The inverse temperature used for scaling the distances in the loss will hence lie in $(0, 1/\tau_\mathrm{min}]$.

\subsubsection*{CEBRA Model parameters used}
In the main figures we used the default parameters (see \url{https://cebra.ai/docs/api.html}) for fitting CEBRA unless otherwise stated in the text (such as dimension, which varied and is noted in figures), or below.

Synthetic data: \codeword{model_architecture='offset1-model-mse'},
\codeword{conditional='delta'},
\codeword{delta=0.1},
\codeword{distance='euclidean'},
\codeword{batch_size=512}, 
\codeword{learning_rate=1e-4}.

Rat hippocampus: \codeword{model_architecture='offset10-model'},
\codeword{time_offsets=10},
\codeword{batch_size=512}.

Primate S1: \codeword{model_architecture='offset10-model'}, \codeword{time_offsets=10},
\codeword{batch_size=512}.

Allen datasets (2P): \codeword{model_architecture='offset10-model'},  \codeword{time_offsets=10},
\codeword{batch_size=512}.

Allen datasets (NP): 

\codeword{model_architecture='offset40-model-4x-subsample'},
\codeword{time_offsets=10},
 \codeword{batch_size=512}.

\subsubsection*{CEBRA API and example usage}
The Python implementation of CEBRA is written in PyTorch \cite{NEURIPS2019_9015} and NumPy \cite{walt2011numpy} and provides an API which is fully compatible with scikit-learn \citep{scikit-learn}, a commonly used package for machine learning. This allows to use scikit-learn tools for hyperparameter selection and downstream processing of the embeddings, e.g., decoding. CEBRA can be used as a drop-in replacement in existing data pipelines for algorithms like tSNE, UMAP, PCA or FastICA. Both CPU and GPU implementations are available.
\medskip

Using the previously introduced notations, suppose we have a dataset containing signals $\ss_t$, continuous context variables $\cc_t$ and discrete context variables $k_t$ for all time steps $t$,

\begin{minted}{python}
import numpy as np
N = 500
s = np.zeros((N, 55), dtype=float)
k = np.zeros((N,), dtype=int)    
c = np.zeros((N, 10), dtype=float)
\end{minted}
along with a second session of data,
\begin{minted}{python}
s2 = np.zeros((N, 75), dtype=float) 
c2 = np.zeros((N, 10), dtype=float)
assert c2.shape[1] == c.shape[1]
\end{minted}

and note that the number of samples as well as the dimension in $\ss'$ does not need to match $\ss$. Session alignment leverages the fact that the second dimension of $\cc$ and $\cc'$ match.
With this dataset in place, different variants of CEBRA can be applied as follows:
\begin{minted}{python}
import cebra
model = cebra.CEBRA(
    output_dimension=8,
    num_hidden_units=32,
    batch_size=1024,
    learning_rate=3e-4,
    max_iterations=1000
)
\end{minted}

The training mode to use is determined automatically based on what combination of data is passed to the algorithm:
\begin{minted}{python}
# time contrastive learning
model.fit(s)
# discrete behavior contrastive learning 
model.fit(s, k)
# continuous behavior contrastive learning 
model.fit(s, c)
# mixed behavior contrastive learning 
model.fit(s, c, k)
# multi-session training
model.fit([s, s2], [c, c2])
# adapt to new session
model.fit(s, c)
model.fit(s2, c2, adapt = True)
\end{minted}

Since CEBRA is a parametric method training a neural network internally, it is possible to embed new data points after fitting the model:
\begin{minted}{python}
s_test = np.zeros((N, 55), dtype=float)
# obtain and plot embedding
z = model.transform(s_test)
plt.scatter(z[:, 0], z[:, 1])
plt.show()
\end{minted}

\subsection*{Consistency of embeddings across runs, subjects, sessions, recording modalities, and areas}

\justify To measure the consistency of the embeddings, we used the $R^2$ score of the linear regression (including an intercept) between the embeddings from different subjects (or sessions).
Secondly, pi-VAE, which we benchmarked and improved (Extended Data Fig.~\ref{fig:CEBRAintroData}), demonstrated a theoretical guarantee that it can reconstruct the true latent space up to an affine transformation. To measure across runs, we measured the ${R^2}$ score of the linear regression between embeddings across 10 runs of the algorithms, yielding 90 comparisons. The runs were done with the same hyperparameters, model, and training setup.
\medskip

For the rat hippocampus data, the number of neurons recorded were different across subjects. The behavior setting was the same: the rats moved in a 1.6 meter long track, and for analysis the behavior data was binned into 100 bins with equal size for each direction (leftwards, rightwards). %
We computed averaged feature vectors for each bin by averaging all normalized \cebra\space embeddings for a given bin, and re-normalized the average to lie on the hypersphere. If a bin does not contain any sample, it was filled by samples from the two adjacent bins.
\cebra \space was trained with latent dimension 3 (the minimum) such that it is constrained to lie only on a 2-sphere (making this ``3D'' space equivalent to 2D Euclidean space). All other methods were trained with 2 latent dimensions in Euclidean space. Note that $n+1$ dimensions of \cebra \space is equivalent to $n$ dimensions of other methods that we compared, since the feature space of \cebra \space is normalized (i.e., the feature vectors are normalized to have unit length).

\medskip

For Allen visual data where the number of behavioral data points are the same across different sessions (i.e., fixed length of video stimuli), we directly computed the $R^2$ score of linear regression between embeddings from different sessions and the modalities. We surveyed 3, 4, 8, 32, 64, 128 latent dimensions with \cebra.
\medskip 

To compare the consistency of embeddings between or within the areas we considered, we computed intra-area and inter-area consistency within the same recording modality (2P or NP).
Within the same modality, we sampled 400 neurons from each area.
We trained one CEBRA model per area, and computed the linear consistency between all pairs of embeddings.
For the intra-area comparison, we sampled an additional 400 disjoint neurons.
For each area, we trained two CEBRA models on these two sets of neurons, and computed their linear consistency.
We repeated this process three times.

For comparisons across modalities (2P and NP), we sampled 400 neurons from each modality (which are disjoint, as above, because one set was sampled from 2P recordings and the other set from the NP recordings).
We trained a multi-session CEBRA model with one encoder for 2P, and one encoder for NP in the same embedding space.
For an intra-area comparison, we computed the linear consistency between the the NP and 2P decoder from the same area.
For an inter-area comparison, we computed the linear consistency between the NP encoder from one area and the 2P encoder from another area and again considered all combinations of areas.
We repeated this process three times.

\medskip

For the comparison of single- and multi-session training (Extended Data Fig.~\ref{fig:multiSession}), we computed embeddings using encoder models with 8, 16, $\dots$, 128 hidden units for varying the model size, and benchmark 8, 16, $\dots$, 128 latent dimensions. Hyperparameters, except for number of optimization steps, were selected according to validation set decoding $R^2$ (rat) or accuracy (Allen). Consistency is reported at the point in training where the position decoding error is less than 7 cm for the first rat in the hippocampus dataset, and a decoding accuracy of 60\% on the Allen dataset.  For single-session training, four embeddings were trained independently on each of the individual animals, while for multi-session the embeddings were trained jointly on all sessions. For multi-session training, the same number of samples was drawn from each session to learn an embedding invariant to the session ID.
The consistency vs. decoding error trade-off (Extended Data  Fig.~\ref{fig:multiSession}c) was reported as the average consistency across all 12 comparisons (Extended Data  Fig.~\ref{fig:multiSession}b) vs. the average decoding performance across all rats and data splits.
\medskip

\subsection*{Model Comparisons}

\subsubsection*{pi-VAE parameter selection, and modifications to pi-VAE}

The original implementation of pi-VAE used a single time bin spiking rate as a input. Thus, we modified their code to allow for larger time bin inputs and found that time window input with receptive field of 10 time bins (250 ms) gave a higher consistency across subjects and better preserved the qualitative structure of the embedding (thereby outperforming the results presented in~\cite{zhou2020learning}; see Extended Data Fig.~\ref{fig:CEBRAintroData}). 
To do this, we used the same encoder neural network architecture as we used for \cebra, and modified the decoder to a 2D output (we call our modified version conv-pi-VAE). Note, we used this modified pi-VAE for all the experiments except for the synthetic setting, where there is no time dimension, thus the original implementation is sufficient.
\medskip

The original implementation reported a median absolute error of 12 cm on rat 1 (the animal they considered most in the work), and our implementation of time windowed input with 10 bins resulted in a median absolute error of 11 cm 
(Fig.~\ref{fig:Topology}). For hyperparameters, we tested different epochs between 600 (the published value used) and 1000,  and learning rate between $1.0\times 10^{-6}$ and $5.0\times 10^{-4}$ via a grid search.  We fixed the hyperparameters to be those that gave the highest consistency across subjects, which were training epochs of 1000 and learning rate $2.5\times 10^{-4}$. All other hyperparameters were kept as in the original implementation~\cite{zhou2020learning}. Note, that the original paper demonstrated that pi-VAE is fairly robust across different hyperparameters. For decoding (Fig.~\ref{fig:Topology}) we considered both a simple kNN decoder (that we use for CEBRA) and the computationally more expensive 
Monte Carlo sampling method originally proposed for pi-VAE~\cite{zhou2020learning}.
Our implementation of conv-pi-VAE can be found at: \url{https://github.com/AdaptiveMotorControlLab/CEBRA}.

\subsubsection*{UMAP parameter selection}

For UMAP~\cite{mcinnes2018umap}, following the parameter guide (\url{umap-learn.readthedocs.io/}), we focused on tuning the number of neighbors ($n\_neighbors$) and minimum distance ($min\_dist$).
The $n\_components$ parameter was fixed to 2 and we used a cosine metric to make a fair comparison with \cebra, which also used the cosine distance metric for learning. We performed a grid search with 100 total hyperparameter values in the range of [2, 200] for $n\_neighbors$ and range of [0.0001, 0.99] for $min\_dist$.
The highest consistency across runs in the rat hippocampus dataset was achieved with $min\_dist$ of 0.0001 and $n\_neighbors$ of 24.
For the other datasets in Extended Data Fig.~\ref{fig:CEBRAvsAllSUPPL}, we used the default value of $n\_neighbors$ as 15 and $min\_dist$ as 0.1.

\subsubsection*{tSNE parameter selection}

For tSNE~\cite{van2009dimensionality}, we used the implementation in openTSNE~\cite{PolicarOPENTSNE}. We performed a sweep on $perplexity$ in the range of [5, 50] and $early\_exaggeration$ in the range [12, 32] following the parameter guide, while fixing $n\_components$ as 2 and used a cosine metric, to fairly compare to UMAP and \cebra. We use PCA initialization to improve the run consistency of tSNE \cite{Kobak2021tsnepca}. The highest consistency across runs in the rat hippocampus dataset was achieved with $perplexity$ of 10 and  $early\_exaggeration$ of 16.44. For the other datasets in Extended Data Fig.~\ref{fig:CEBRAvsAllSUPPL}, we used the default value of $perplexity$ of 30 and $early\_exaggeration$ of 12.

\subsection*{Decoding Analysis}

\justify  We primarily used a simple k-Nearest Neighbors (kNN) algorithm, which is a non-parametric supervised learning method, as a decoding method for CEBRA. We used the implementation in scikit-learn~\cite{scikit-learn}. We used a kNN regressor for continuous value regression and a kNN classifier for discrete label classification, using uniform weights on distances of k-nearest neighbors. For the embeddings obtained with cosine metrics, we used cosine distance metrics for kNN and Euclidean distance metrics for the embeddings obtained in Euclidean space. 
\medskip

For the rat hippocampus data, a kNN regressor, as implemented in scikit-learn~\cite{scikit-learn}, was used to decode the position, and a kNN classifier to decode the direction. The number of neighbors was searched over the range $\{1,4,9,16,25\}$ and we used the cosine distance metric.
We used the $R^2$ score of predicted position and direction vector on the validation set as a metric to choose the best $n\_neighbors$ parameter.  
We report the median absolute error (MAE) for the positional decoding on the test set. 
For pi-VAE, we additionally evaluate decoding quality using the originally proposed decoding method based on Monte Carlo sampling, using the settings from the original paper~\cite{zhou2020learning}.
Note, UMAP, tSNE and \cebra-Time were trained using the full dataset without label information when learning the embedding, and we used the above split only for training and cross-validation of the decoder. 
\medskip

For the direction decoding within the monkey dataset, we used a Ridge classifier~\cite{scikit-learn} as a baseline. The regularization hyperparameter was searched over $[10^{-6}, 10^{2}]$. For \cebra, we used a kNN classifier for decoding direction with $k$ searched over the range  [1, 2500]. For conv-pi-VAE, we searched for the best learning rate over $[1.0\times 10^{-5}, 1.0\times 10^{-3}]$.
For position decoding, we used Lasso~\cite{scikit-learn} as a baseline. The regularization hyperparameter was searched over $[10^{-6}, 10^{2}]$. For conv-pi-VAE, we used 600 epochs and searched for the best learning rates over $[5\times10^{-4}, 2.5\times10^{-4}, 0.125\times10^{-4}, 5\times10^{-5}]$, via a grid of (x,y) space in 1 cm bin for each axis as the sampling process for decoding. For \cebra, we used the kNN regression, and the number of neighbors $k$ was again searched over [1, 2500]. 

\medskip

For the Allen Institute datasets, we performed decoding (frame number or scene classification) for each frame from Movie 1. Here, we used a kNN classifier~\cite{scikit-learn} with a population vector kNN as a baseline, similar to the decoding of orientation grating as performed in~\cite{de2020large}. For \cebra, we used the same kNN classifier method on the \cebra \space features. In both cases, the number of neighbors $k$ was searched over a range of [1, 100] in an exponential fashion. We used the neural data recorded during the first 8 repeats as the train set, and the $9^{th}$ repeat for validation to choose the hyperparameter, and the last repeat as the test set to report the decoding accuracy.
We also used a Gaussian Naive Bayes decoder~\cite{scikit-learn} to test linear decoding from the \cebra \space model and neural population vector. Here, we assumed uniform priors over frame number and searched over a range of [$10^{-10}, 10^{3}$] in an exponential manner for $smoothing var$ hyperparameter. 
\medskip

For layer specific decoding we used data from excitatory neurons in area VISp:
layer 2/3 [Emx1-IRES-Cre, Slc17a7-IRES2-Cre]; 
layer 4 [Cux2-CreERT2, Rorb-IRES2-Cre, Scnn1a-Tg3-Cre]; 
layer 5/6 [Nr5a1-Cre, Rbp4-Cre\_KL100, Fezf2-CreER, Tlx3-Cre\_PL56, Ntrsr1-cre].

\subsection*{Topological Analysis}

\justify For the persistent co-homology analysis, we utilized ripser.py~\cite{ctralie2018ripser}. For the hippocampus dataset we used 1,000 randomly sampled points from \cebra-Behavior trained with temperature 1, time offset 10 and mini-batch size 512 for 10k training steps on the full dataset, and then analyzed up to the 2D co-homology. Maximum distance considered for filtration was set to infinity. To decide the number of co-cycles in each co-homology dimension with a significant lifespan, we trained 500 \cebra \space embeddings with shuffled labels, similar to the approach in~\cite{Gardner2021.02.25.432776}. We took the maximum lifespan of each dimension across these 500 runs as a threshold to determine robust Betti numbers. We surveyed the Betti numbers of \cebra \space embeddings across 3, 8, 16, 32, and 64 latent dimensions. 
\medskip

Next, we used DREiMac~\cite{dreimac} to obtain topology-preserving circular coordinates (radial angle) of the first co-cycle ($H_1$) from the persistent co-homology analysis. 
Similar to above, we used 1,000 randomly sampled points from the \cebra-Behavior models of embedding dimensions 3, 8, 16, 32 and 64.

\subsection*{Behavior Embeddings for Video Datasets}

\justify High dimensional inputs, such as videos, need further pre-processing for effective use with \cebra.
Firstly, we used the recently presented DINO model~\cite{Caron2021EmergingPI} to embed video frames into a 768-dimensional feature space.
Specifically, we used the pretrained ViT/8 vision transformer model, which was trained by a self-supervised learning objective on the ImageNet database. This model is particularly well-suited for video analysis, and among the state-of-the-art models for embedding natural images into a space appropriate for k-nearest neighbour search \cite{Caron2021EmergingPI}, a desired property to make the dataset compatible with \cebra.
We obtained a normalized feature vector for each video frame, which was then used as the continuous behavior variable for all further \cebra \space experiments.
\medskip

For scene labels, 3 individuals labeled each video frame using 8 candidate descriptive labels allowing multi-label classes. We took the majority vote of the 3 individuals to decide the label of each frame. In case of multi-labels, we considered this as a new class label. The above procedure resulted in 10 classes of frame annotation.

\beginsupplement
\onecolumn

\captionsetup[figure]{name=Extended Data Fig.}

\begin{figure*}
\begin{center}
\includegraphics[width=\textwidth]{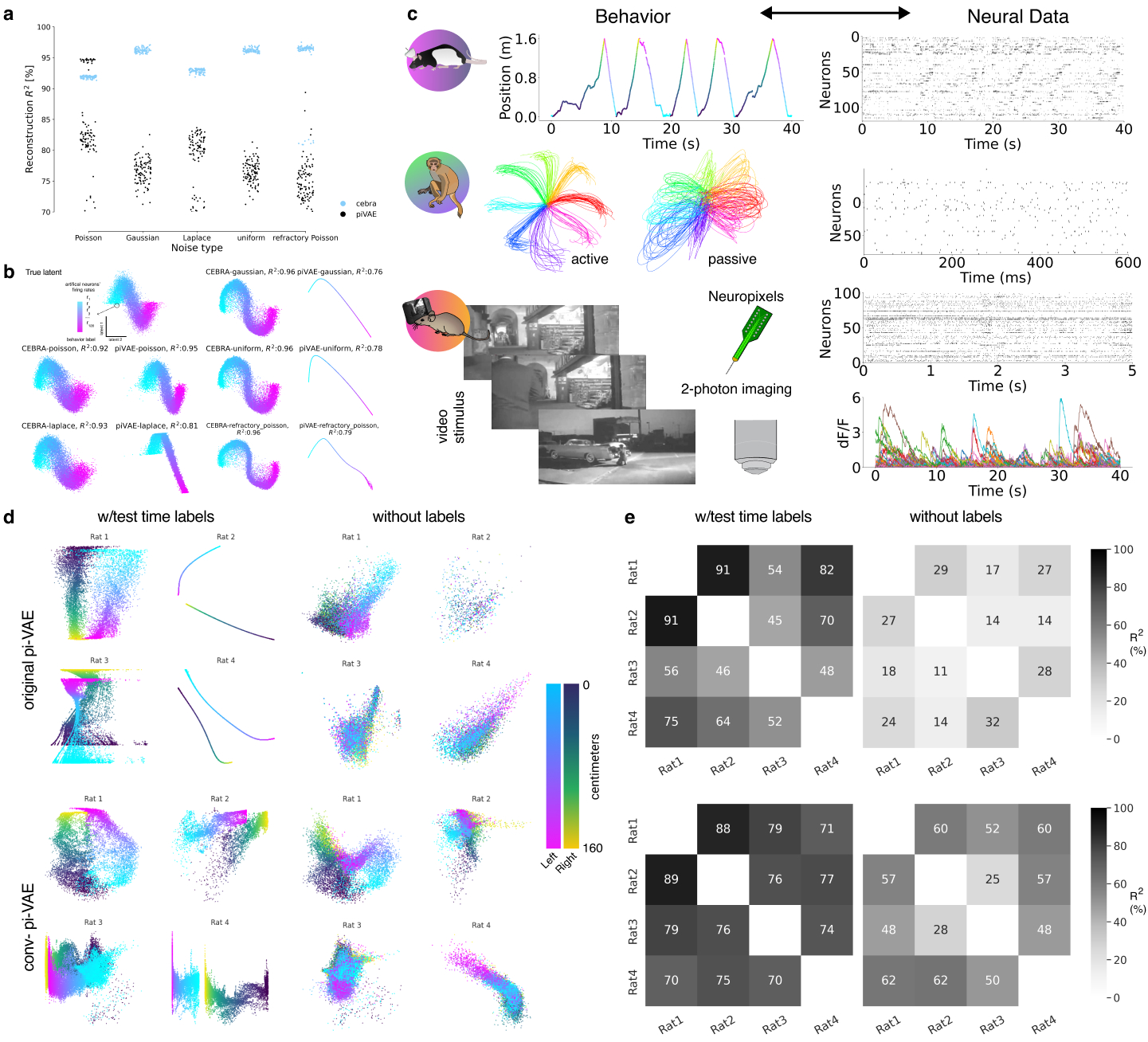}
\end{center}
\caption{{\bf Overview of datasets, synthetic data, \& original pi-VAE implementation vs. modified conv-pi-VAE.} 
{\bf ab)}: We generated synthetic datasets similar to Fig.~\ref{fig:CEBRAvsAll}b with additional variations in the noise distributions in the generative process. We benchmarked the reconstruction score of the true latent using CEBRA and pi-VAE (100 seeds) on the generated synthetic datasets. CEBRA showed higher and less variable reconstruction scores than pi-VAE in all noise types.
{\bf (b)} Example visualization of the reconstructed latents from CEBRA and pi-VAE on different synthetic dataset types. 
{\bf (c)}: we benchmarked and demonstrate the abilities of CEBRA on four datasets. Rat-based electrophysiology data from~\citet{grosmark2016diversity}, where the animal transversed a 1.6m linear track ``leftwards'' or ``rightwards''. Two mouse-based datasets: one 2-photon calcium imaging passively viewing dataset from~\citet{de2020large}, and one with the same stimulus but recorded with Neuropixels~\cite{Siegle2021SurveyOS}. A monkey-based electrophysiology dataset of center out reaching from~\citet{chowdhury2020area}, and processed to trial data as in~\citet{pei2021neural}.
{\bf (d)}: Conv-pi-VAE showed improved performance, both with labels (Wilcoxon signed-rank test, p=0.0341) and without labels Wilcoxon signed-rank test, p=0.0005). Example runs/embeddings the consistency across rats, with {\bf (e)}:  consistency across rats, from target to source, as computed in Fig.~\ref{fig:CEBRAvsAll}.
}
\label{fig:CEBRAintroData}
\end{figure*}

\begin{figure*}
\begin{center}
\includegraphics[width=.8\textwidth]{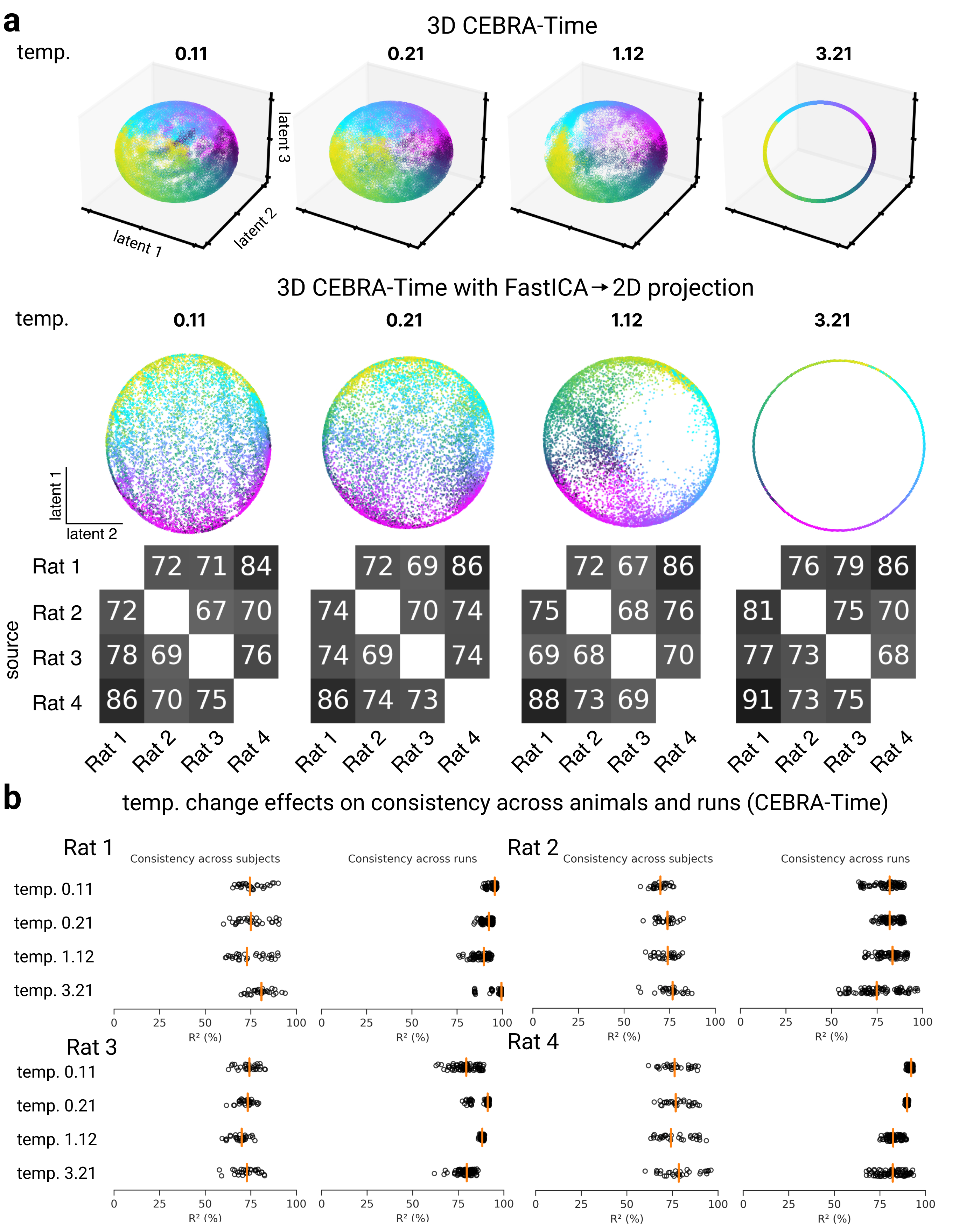}
\end{center}
\caption{{\bf Hyperparameter changes on visualization and consistency.} {\bf (a)}: Temperature has the largest effect on visualization (vs. consistency) of the embedding as shown by a range from 0.1 to 3.21 (highest consistency for Rat 1), as can be appreciated in 3D (top) and post FastICA into a 2D embedding (middle). Bottom row shows the corresponding change on mean consistency, and in\textbf{ b,} the variance can be noted. Orange line denotes the median and black dots are individual runs (subject consistency: 10 runs with 3 comparisons per rat; run consistency: 10 runs, each compared to 9 remaining runs).
}
\label{fig:CEBRAvsAlltemp}
\end{figure*}

\begin{figure*}
\begin{center}
\includegraphics[width=.75\textwidth]{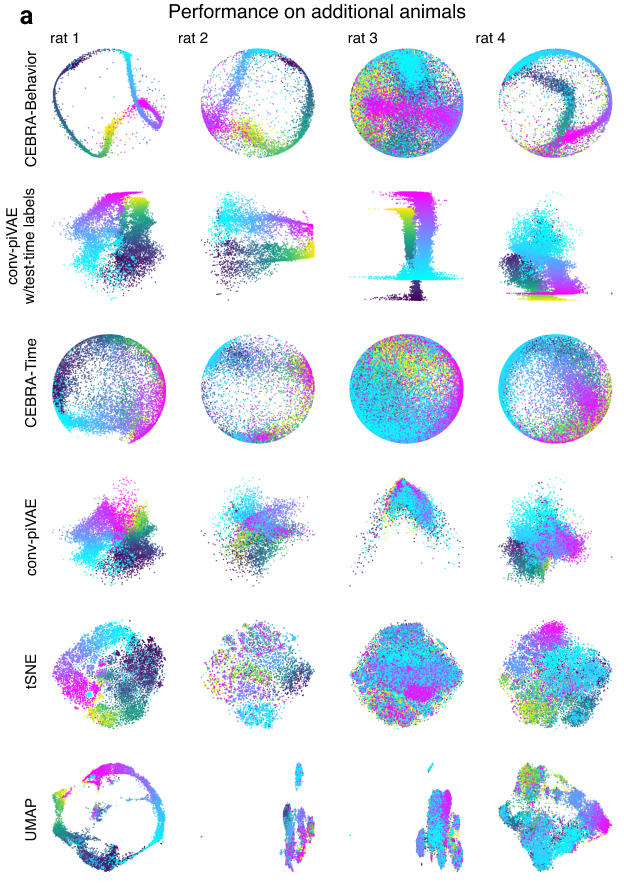}
\end{center}
\caption{{\bf \cebra \space produced consistent, highly decodable embeddings} {\bf (a)}: Additional rat data shown for all algorithms we benchmarked (see Methods). For \cebra-Behavior, we used temperature 1, time offset 10, batch size 512 and 10k training steps. For \cebra-Time, we used temperature 2.25, time offset 10, batch size 512 and 4k training steps. For UMAP, we used the cosine metric and $min\_dist$ of 0.99 and $n\_neighbors$ of 31. For tSNE we used cosine metric and $perplexity$ of 29. For conv-pi-VAE, we trained 1000 epochs with learning rate $2.5\times 10^{-4}$. 
\cebra \space was trained with output latent 3D (the minimum) and all other methods were trained with a 2D latent.}
\label{fig:CEBRAvsAllSUPPL}
\end{figure*}

\begin{figure*}
\begin{center}
\includegraphics[width=.6\textwidth]{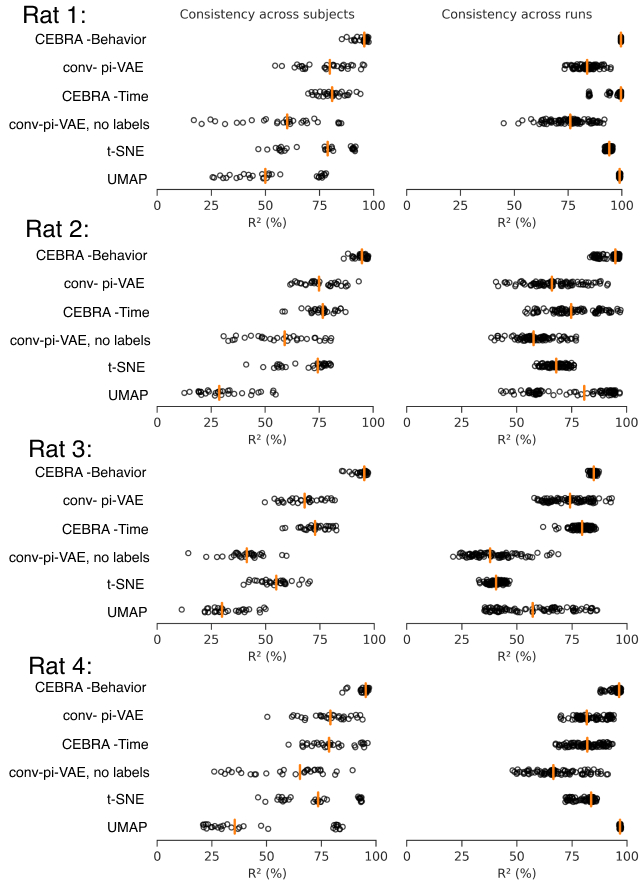}
\end{center}
\caption{{\bf Additional metrics used for benchmarking consistency} {\bf (a)}:  Comparisons of all algorithms along different metrics for Rats 1, 2, 3, 4. The orange line is median across n=10 runs, black circles denote individual runs. Each run is the average over three non-overlapping test splits.
}
\label{fig:CEBRAvsAllSUPPLrats}
\end{figure*}

\begin{figure*}
\begin{center}
\includegraphics[width=\textwidth]{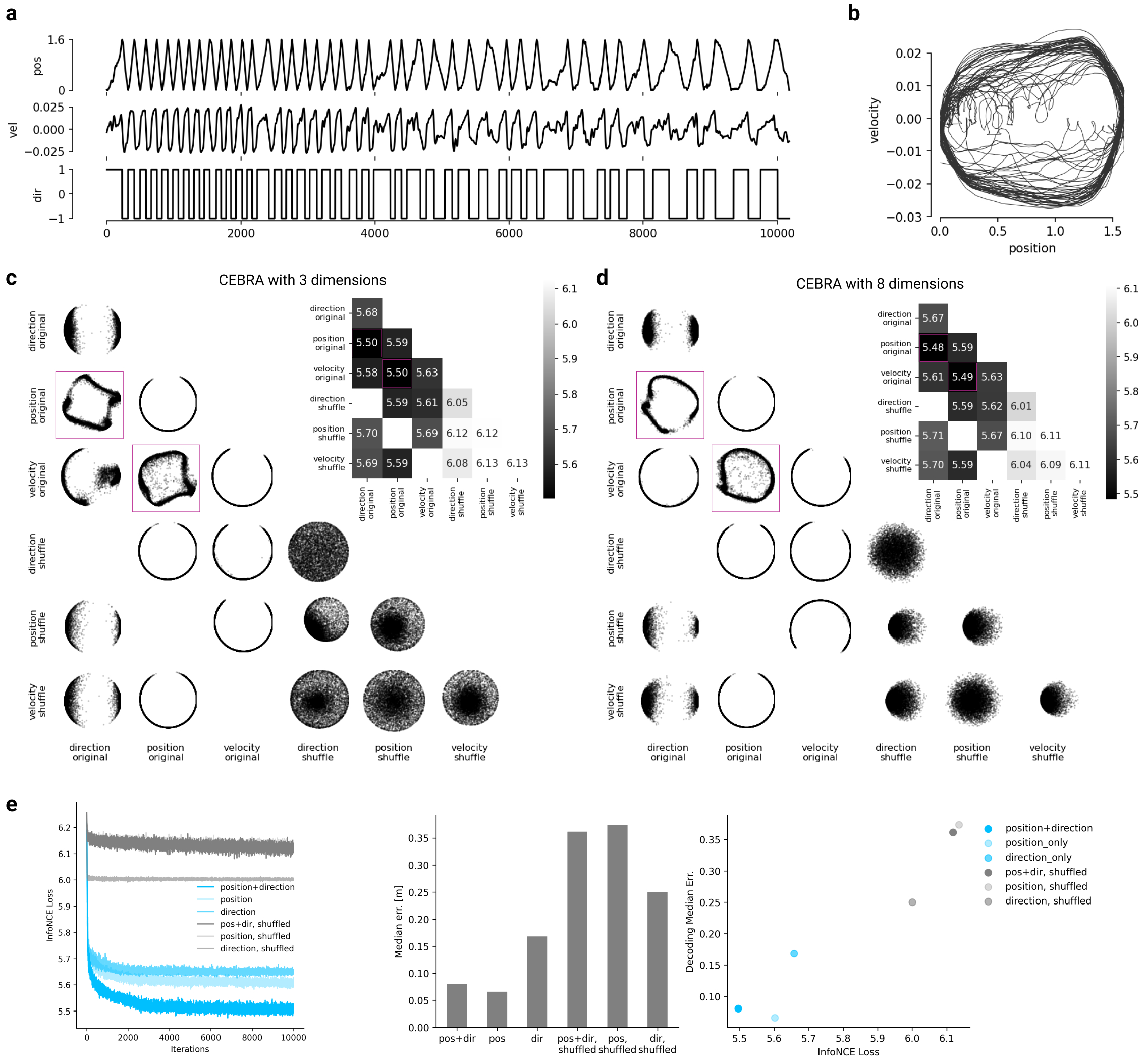}
\end{center}
\caption{{\bf Hypothesis testing with CEBRA}
    {\bf (a)}: Example data from a hippocampus recording session (Rat 1). We tested possible relationships between three experimental variables (rat location, velocity, movement direction) and the neural recordings (120 neurons, not shown).
    {\bf (b)}: Relationship between velocity and position.
    {\bf (c)}: We trained CEBRA with three-dimensional outputs on every single experimental variable (main diagonal) and every combination of two variables. All variables are treated as ``continuous'' in this experiment. We compared original to shuffled variables (shuffling is done by permuting all samples over the time dimension) as a control. We projected the original three dimensional space onto the first principal components. We show the minimum value of the InfoNCE loss on the trained embedding for all combinations in the confusion matrix (lower number is better).
    Either velocity or direction, paired with position information is needed for maximum structure in the embedding (highlighted, colored), yielding lowest InfoNCE error. 
    {\bf (d)}: Using an eight-dimensional CEBRA embedding did not qualitatively alter the results. We again report the first two principal components as well as InfoNCE training error upon convergence, and find non-trivial embeddings with lowest training error for combinations of direction/velocity and position.
    {\bf (e)}: The InfoNCE metric can serve as the goodness of fit metric, both for hypothesis testing and identifying decodable embeddings. We trained CEBRA in discovery-driven mode with 32 latent dimensions (empirically the best setup for decoding). We compared the InfoNCE loss (left, middle) between various hypotheses. Low InfoNCE was correlated with low decoding error (right).
    }
\label{fig:SFigure3_hypoTesting}
\end{figure*}

\begin{figure*}
\begin{center}
\includegraphics[width=\textwidth]{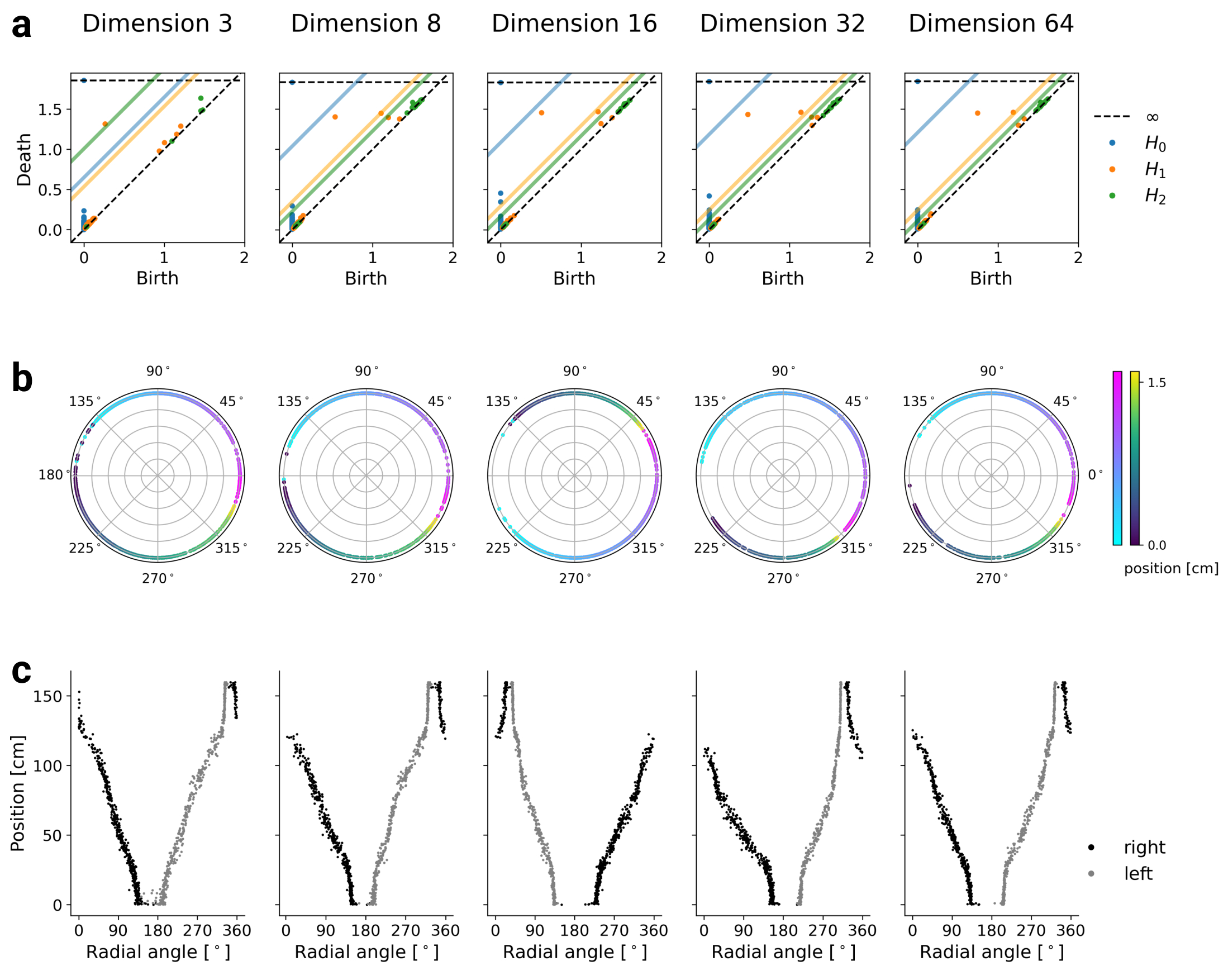}
\end{center}
\caption{{\bf Persistence across dimensions} {\bf (a)}: For each dimension of \cebra-Behavior embedding from the rat hippocampus dataset Betti numbers were computed by applying persistent co-homology. The colored dots are lifespans observed in hypothesis based \cebra-Behavior. To rule out noisy lifespans, we set a threshold (colored diagonal lines) as maximum lifespan based on 500 seeds of shuffled-CEBRA embedding for each dimension.
{\bf (b)}: The topology preserving circular coordinates using the first co-cycle from persistent co-homology analysis on the CEBRA embedding of each dimension is shown (see Methods). The colors indicate position and direction of the rat at the corresponding CEBRA embedding points. 
{\bf (c)}: The radial angle of each embedding point obtained from {\bf (b)} and the corresponding position and direction of the rat. 
}
\label{fig:SFigure3_persistence}
\end{figure*}

\begin{figure*}
\begin{center}
\includegraphics[width=\textwidth]{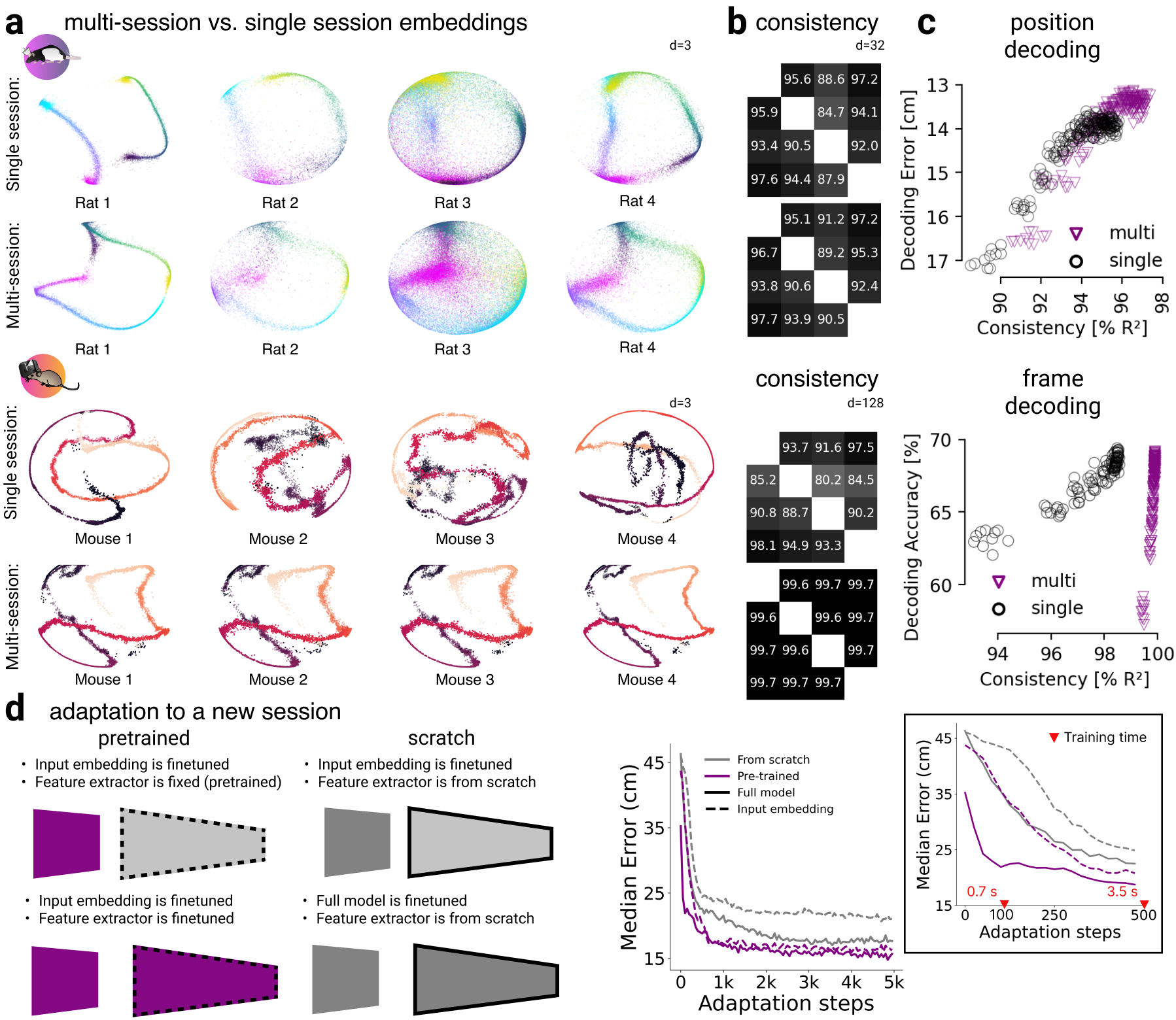}
\end{center}
\caption{{\bf Multi-session training and rapid decoding} {\bf (a)}: Top: hippocampus dataset, single animal vs. multi-animal training shows an increase in consistency across animals. Bottom: same for Allen dataset, 4 mice. {\bf (b)}: consistency matrix single vs. multi-session training for hippocampus (32D embedding) and Allen data (128D embedding) respectively.  Consistency is reported at the point in training where the average position decoding error is less than 14 cm (corresponds to 7 cm error for rat 1), and a decoding accuracy of 60\% on the Allen dataset.
\textbf{(c)}: Comparison of decoding metrics for single or multi-session training at various consistency levels (averaged across all 12 comparisons). Models were trained for 5,000 (single) or 10,000 (multi-session) steps with a 0.003 learning rate; batch size was 7,200 samples per session. Multi-session training requires longer training or higher learning rates to obtain the same accuracy due to the 4-fold larger batch size, but converges to same decoding accuracy. We plot points at intervals of 500 steps (n=10 seeds); training progresses from lower right to upper left corner within both plots.
{\bf (d)}: We demonstrate that we could also adapt to an unseen dataset; here, 3 rats were used for pretraining, and rat \#4 was used as a held-out test. The grey lines indicate models trained from scratch (random initialization). We also tested fine-tuning only the input embedding (first layer) or the full model, as the diagram, left, describes. We measured the average time (mean $\pm$ STD) to adapt 100 steps (0.65 $\pm$ 0.13 sec) and 500 steps (3.07 $\pm$ 0.61 sec) on 40 repeated experiments.
} 
\label{fig:multiSession}
\end{figure*}

\begin{figure*}
\begin{center}
\includegraphics[width=\textwidth]{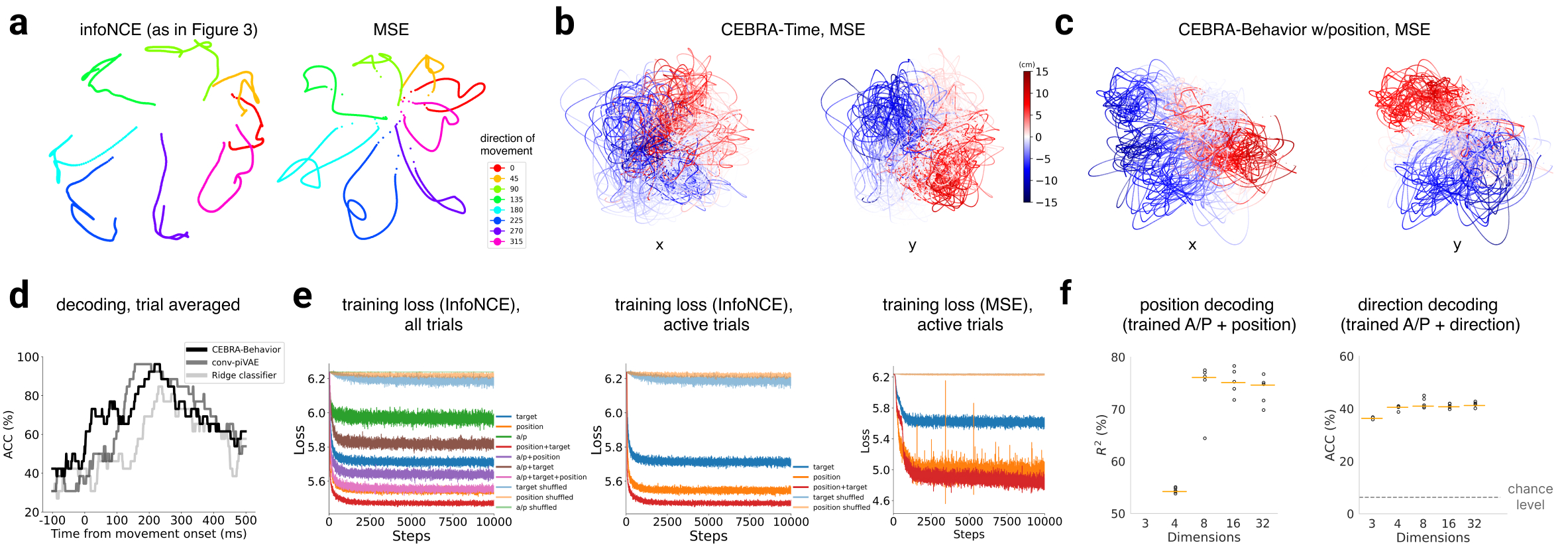}
\end{center}
\caption{{\bf Somatosensory cortex decoding from primate recordings} 
{\bf (a)}: We compare \cebra-Behavior with the cosine similarity and embeddings on the sphere reproduced from Fig.~\ref{fig:Reaching}b (left) against \cebra-Behavior trained with the MSE loss and unnormalized embeddings.
The embeddings of trials (n=364) of each direction were post-hoc averaged.
{\bf (b)}: \cebra-Behavior trained with x,y position of the hand. Left panel is color-coded to changes in x position and right panel is color-coded to changes in y position.
{\bf (c)}: \cebra-Time without any external behavior variables. As in \textbf{b}, left and right are color-coded to x and y position, respectively.
{\bf (d)}: Decoding performance of on target direction using \cebra-Behavior, conv-pi-VAE and a linear classifier. \cebra-Behavior shows significantly higher decoding performance than the linear classifier (one-way ANOVA, F(2,75)=3.37, p<0.05 with Post Hoc Tukey HSD p<0.05). 
{\bf (e)}: Loss (InfoNCE) vs. training iteration for \cebra-Behavior with position, direction, active or passive, and position+direction labels (and shuffled labels) for all trials (left) or only active trials (right), or active trials with a MSE loss.
{\bf (f)}: Additional decoding performance results on position and direction-trained CEBRA models with all trial types. For each case, we trained and evaluated 5 seeds represented by black dots and the orange line represents the median.
}
\label{SupplReaching}
\end{figure*}

\begin{figure*}
\begin{center}
\includegraphics[width=.8\textwidth]{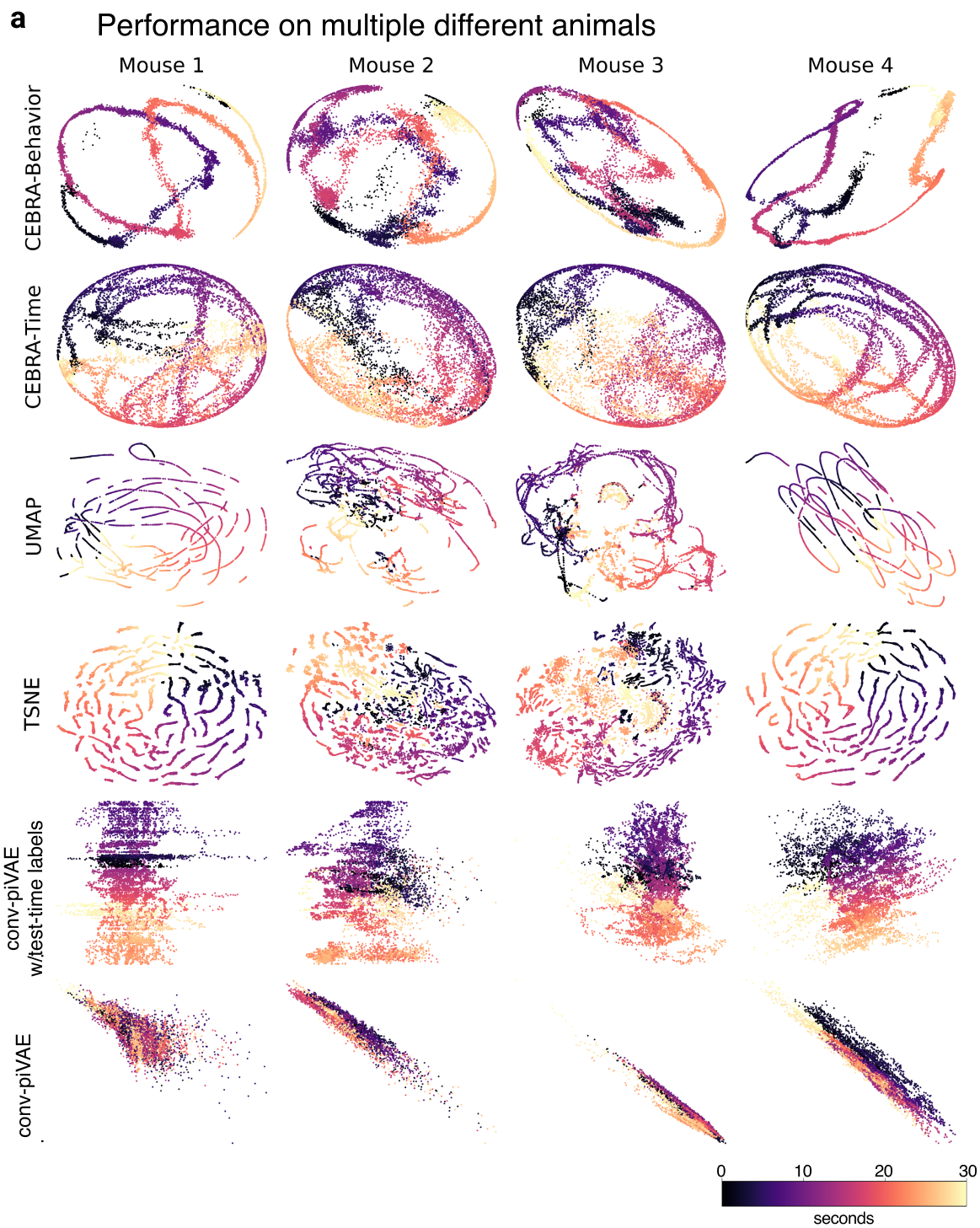}
\end{center}
\caption{{\bf \cebra \space produces consistent, highly decodable embeddings} {\bf (a)}: Additional 4 sessions with the most neurons in the Allen visual dataset calcium recording shown for all algorithms we benchmarked (see Methods). For \cebra-Behavior and \cebra-Time, we used temperature 1, time offset 10, batch size 128 and 10k training steps. For UMAP, we used a cosine metric and $n\_neighbors$ 15 and $min\_dist$ 0.1. For tSNE, we used a cosine metric and $perplexity$ 30. For conv-pi-VAE, we trained with 600 epochs, a batch size of 200 and a learning rate $5\times 10^{-4}$. All methods used 10 time bins input. \cebra \space was trained with 3D latent and all other methods were obtained with an equivalent 2D latent dimension.}
\label{fig:SUPPLAllenDataFigureMice}
\end{figure*}

\begin{figure*}
\begin{center}
\includegraphics[width=.87\textwidth]{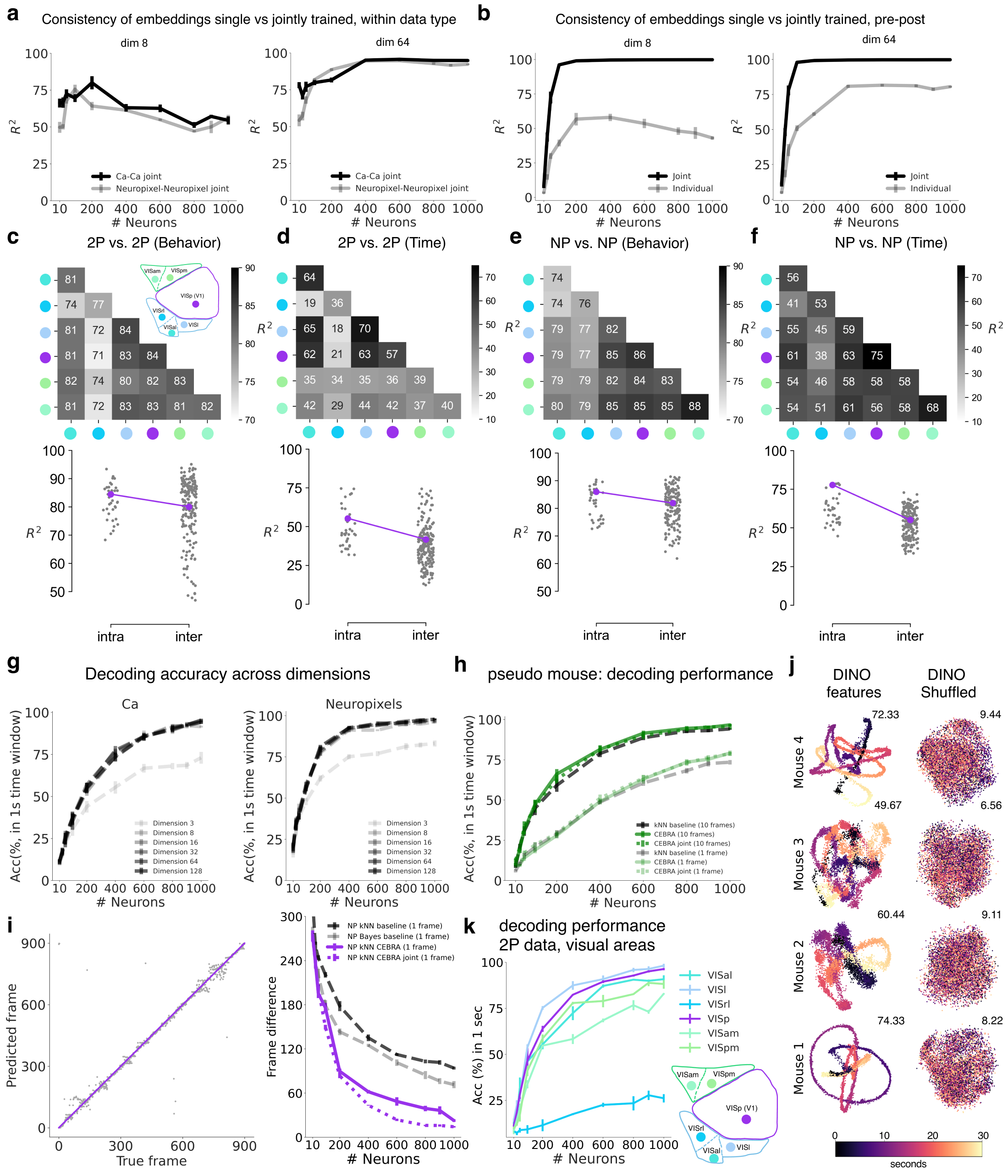}
\end{center}
\caption{{\bf Spikes and calcium signaling reveal similar embeddings}
{\bf (a)}: Consistency between the single modality embedding and jointly trained embedding from \cebra. In higher dimensions, the embedding from single recording modality and the jointly trained embedding became highly consistent with more neurons.
{\bf (b)}: Consistency of embeddings from two recording modalities, when a single modality was trained independently and/or jointly trained. The consistency significantly improved with joint training. In higher dimensions, the consistency between single modality embeddings improved as well, which shows that \cebra \space can find 'common latents' in two different recording methods (that is theoretically meant to have same information) even without joint training (yet, joint training improves consistency). This data is also presented in Fig.~\ref{fig:AllenDataFigure}e, h, but here plotted together to show improvement with joint training. 
{\bf (c-f)}: Consistency across modalities and areas for CEBRA-Behavior and -Time (as computed in Fig.~\ref{fig:AllenDataFigure}i-k). The purple dots indicate mean of intra-V1 scores and inter-V1 scores (inter-V1 vs intra-V1 Welch's t-test; 2P (Behavior): T(10.6)=1.52, p=0.081, 2P (Time): T(44.3)=4.26 ,p=0.0005, NP (Behavior): T(11.6)=2.83, p=0.0085, NP (Time): T(8.9)=15.51, p<0.00001)
{\bf (g)}: CEBRA + kNN decoding performance (see Methods) of \cebra \space embeddings of different output embedding dimensions, from calcium (2P) data or Neuropixels, as denoted.%
{\bf (h)}: Decoding accuracy measured by considering predicted frame being within 1 sec difference to true frame as correct prediction using CEBRA (2P only), jointly trained (2P+NP), or a baseline population vector kNN decoder (using the time window 33 ms (single frame), or 330 ms (10 frame receptive field)).
{\bf (i)}: Single frame performance and quantification using CEBRA 1 frame receptive field (NP data), or baseline models.
{\bf (j}: As a control experiment we shuffled DINO features: CEBRA-Behavior used the DINO features as behavior labels and CEBRA-Shuffled used the shuffled DINO features. We shuffled the frame order of DINO features within a repeat. Same shuffled order was use for all repeats. Color code is frame number from the movie. The prediction is considered as true if the predicted frame is within 1 sec from the true frame, and the accuracy (\%) is noted next to the embedding. For Mice ID 1-4: 337, 353, 397, 475 neurons were recorded, respectively.
{\bf (k)}: Decoding performance from 2P data from different visual cortical areas from different layers (2/3, 4, 5/6), as denoted, using a 10 frame window \cebra-Behavior model using DINO features with 128 output dimension.}
\label{fig:SUPPLAllenDataFigure}
\end{figure*} 

\end{multicols}

\clearpage
\setlength{\parskip}{\medskipamount}
\captionsetup[table]{name=Supplementary Table}

\section*{\huge Supplementary Information}

\newparagraph{Suppl. Video 1:} ``SupplVideo\_1.MP4''
Corresponding to Fig. 2d. CEBRA-Behavior trained with position+direction on Rat 1. Video is in 2X real-time.
\newline

\newparagraph{Suppl. Video 2:} ``SupplVideo\_2.MP4''
Corresponding to Fig. 5b. The left panels show example calcium traces from 2-photon imaging (top) and spikes from Neuropixels recording (bottom) of primary visual cortex while the video is shown to a mouse. The center panel shows an embedding space constructed by jointly training a CEBRA-Behavior model with 2-photon and Neuropixels recordings using DINO frame features as labels. The trace is embedding of held-out test repeat from Neuropixels recording. The colormap indicates frame number of the 30 second long video (30 Hz). The last panels show true video (top) and the predicted frame sequence (bottom) using kNN decoder on CEBRA-Behavior embedding from the test set. Video is in real-time.

\section*{Supplementary Note 1}

\subsection*{On identifiability and consistency}

When learning (non-linear) representations of a dataset, it is highly desirable that embedding algorithms generate \emph{consistent} embedding spaces. Multiple runs of the algorithm on the same data, multiple runs of the algorithm on data produced in the same way, etc., should generate embedding spaces with a meaningful relation to each other. %
This ``meaningful relation'' between algorithm runs can be formalized using tools from identifiability in non-linear independent component analysis (ICA). Suppose we are given two models $\ff'$ and $\ff^*$ trained on the same dataset, and the performance of these models matches in the sense that they represent the same probability distribution $p' = p^*$. Identifiability then entails that both models are the same up to some known class of transformations (e.g., linear or affine transformations, rotations, permutations and sign-flips, etc.).

For example, one option for parameterizing the distributions is as $p'(\yy | \xx, \yy_1 \dots \yy_n) = \exp(\ffx(\xx)^\top \ffy(\yy)) / \sum_i \exp(\ffx(\xx)^\top \ffy(\yy_i))$ and respectively for $\tilde p'$ defined respectively with $\tffx$ and $\tffy$. %
\citet{Roeder2020} show that if the two distributions match contrastive learning models produce consistent embedding spaces, and it is possible to find a linear mapping $\LL$ between the feature spaces, i.e., $\LL \ffx(\xx) = \tffx(\xx)$ for all $\xx$ in the dataset.
Other theoretical work has shown that contrastive learning with auxiliary variables is identifiable for bijective neural networks using the noise contrastive estimation (NCE) loss~\cite{Hyvrinen2019NonlinearIU}, and that with an InfoNCE loss this bijectivity assumption can be removed for certain distributions~\cite{Zimmermann2021ContrastiveLI}.
We will adapt the underlying proofs to our setup in Suppl. Note 2, and give a high-level outline below.

We will consider two important points in both the context of discovery and hypothesis driven training of CEBRA models. Firstly, when applying discovery-driven CEBRA, will two models estimated on comparable experimental data agree in their inferred representation? Second, under which assumptions about the data will we be able able to discover the \emph{true} latent distribution?

\medskip

\newparagraph{Consistency:}
For consistency across embedding spaces, we require a dataset with a sufficient amount of variability in time.
Intuitively, to estimate a $d$ dimensional embedding that is consistent across runs, points sampled from the embedding via the negative distribution $q$ need to vary in at least $d$ directions for each possible reference sample in the dataset.
Interestingly, consistency is mostly independent from the data generating process (i.e., the data modality of the recording) and merely requires a sufficiently varying dataset, as well as a choice of feature encoder that passes this variability on to the embedding space.

For example, consider the reaching dataset in Fig. 3 where we showed embeddings that vary in two dimensions (the direction and distance from the center). In this case, the sampling process needs to be designed such that for each reference point we can draw from the dataset, the embedding of the negative samples will vary in at least two directions. This is clearly the case for our training setup: The neurons encode both position and direction information, this information is transformed by the feature encoder, and the resulting embedding varies in at least two directions when the negative distribution samples uniformly across the dataset.

For the first property, we can leverage previous results on the consistency of contrastive learning models over multiple runs \citep{Roeder2020}. Consider the case where we train multiple CEBRA models on data originating from the same data distribution, and consider that we can train these models to full convergence. It is then guaranteed that the embedding spaces will agree up to a linear indeterminacy. In other words, it will always be possible to transform one embedding space into the other by applying a linear transformation.
Linear consistency of representations is interesting when we consider linear downstream processing of the inferred embedding space, as is common in neuroscience~\cite{Urai2022LargescaleNR}. Such a downstream algorithm (e.g., a linear regression or general linear model) will yield the same performance across different CEBRA models.
\medskip

\newparagraph{Recovering the ground-truth latents:}
Note that this notion of consistency only makes a statement about the \emph{inferred} latent representation (and identifiability) across multiple runs of the algorithm, but not yet about the relation between this latent representation and the \emph{true} underlying latent variables that generated the data.
This is the second property mentioned above, and requires additional assumptions about the data generating process to resolve the ambiguity of what a ``latent'' underlying a given dataset actually entails.
The assumptions concern the injectivity of the data generating process and the positive distribution $p$. While for discovery driven training, $p$ is an empirical property of the dataset, hypothesis driven training allows to precisely define $p$ based on the observed auxiliary variables. The same theory applies to both cases.
Importantly, as for consistency, note that these results are independent from the actual modality of the data and other properties of the signal space we consider. The assumptions are all with respect to the underlying latent distribution.

For the analyses in this paper, the theory for linear identifiability of the underlying latents applies: 
For time-contrastive training, it is required that the underlying latent distribution is uniform (e.g., there is no inherent bias in the experimental data), and that latents of nearby time steps vary according to a distribution of the form $p(\vv | \uu) = \exp(\phi(\uu, \vv))$, where $\uu$ and $\vv$ underlying the signal variables $\xx$ and $\yy$. If these conditions are met, the true underlying latents are recovered up to a linear transformation.
For hypothesis testing where the user actually specifies the distribution $p$, this requirement can be easily validated and met.
\medskip

\newparagraph{Relationship between consistency, identifiability, and the sampling mechanism:}
The CEBRA software package allows for other choices of similarity measures (potentially learnable), which allows to derive guarantees also for these cases. In the most general case, we pick $\phi$ as a trainable neural network that factorizes into individual components,
$\phi(\yy, \xx) := \sum_{i}^{d} \phi_i(y_i, \xx)$
where each $\phi_i$ is an individually trained neural network.
For sufficiently variable distributions, this allows to recover the underlying latents up to permutations and point-wise non-linear, bijective transformations.

Likewise, it is possible to modify the encoding networks $\ff$ and $\ff'$ for $\xx$ and $\yy$, respectively. While our experiments used one network $\ff = \ff'$ with $\xx$ and $\yy$ representing neural data, it is well possible to encode different aspects of the dataset and/or to break the symmetry between the two encoding networks and to train a separate network for each of $\ff$ and $\ff'$.
For example, neural data $\yy$ could be encoded using $\ff'$, and behavior $\xx$ could be encoded using $\ff$. It would also be possible to use a composition of neural and behavioral data for $\xx$.
In these cases, if $\ff$ and $\ff'$ are parameterized as two individual neural networks and $\phi$ is defined as the dot-product as before, if $\yy | \xx$ follows a conditionally exponential distribution, we are able to recover all sufficient statistics of this distribution up to a linear transformation.
\medskip

\newparagraph{Examples}:
As an example of the aforementioned results, let us consider the rat hippocampus dataset used in Fig.~1 and 2.
The auxiliary information in this dataset is the position, velocity, and direction of the rat on the linear track. 

We can apply different sampling schemes for investigating this dataset. For example, we can apply discovery-driven, time-contrastive learning.
In this setup, we sample time steps uniformly from the dataset to arrive at our reference samples. Given a time offset $\Delta$ (that informs the algorithm about the time-scale of interest), we obtain positive samples. The resulting batch will be composed of samples $\ss_t$ for the reference, $\ss(t + \Delta)$ for the positive, and $\ss(t_i)$ with uniformly sampled time steps $t_1,\dots,t_n$ for the negative samples. This corresponds to an approximation of the distribution $p(\uu_{t+\Delta} | \uu_t)$ of how the latents vary over the course of time.
If sufficient variation is present in the dataset along $d$ latent directions, CEBRA models will become, after training, consistent across runs. 
If additionally the \emph{true} distribution $p(\uu_{t+\Delta} | \uu_t)$ follows, e.g., a Gaussian distribution, CEBRA will identify the ground truth latents.

In comparison, our so-called hypothesis-driven, behavior-label guided contrastive learning approach would leverage the continuous position information as well as the movement direction of the rat. In the Methods, we denoted the continuous variable as $\cc_t$ and the discrete variables as $k_t$.
To arrive at a behavior contrastive embedding using this auxiliary information, we would build a set of differences.
If the variables are independent (or should reflect this in the embedding), we build one set $D = \{\cc_{t + \Delta} - \cc_t\}_{t = 1}^{T}$. For a reference sample at time step $t$, we sample $\dd \sim D$ uniformly, and apply this difference to the position $\cc_t$ at step $t$. We then pick the point closest to $\cc_t + \dd$ \emph{and} matching the discrete variable $k_t$ as the positive sample.

A lot of variations of this sampling process are possible to embed desirable properties and test hypotheses about the dataset. For instance, consider the primate reaching dataset: Here, $\cc_t$ could be selected as the x/y position in space, and the discrete label $k_t$ could denote the reaching direction. However, the 2D differences $\cc_{t + \Delta} - \cc_t$ will depend on $k_t$: A reaching direction towards the left will have most variance in negative x-direction $[-1, 0]^\top$, a reaching direction towards the top will have most variance in positive y-direction $[0, 1]^\top$.
One way to work around this issue is to consider a polar representation of the position and direction and apply the scheme outlined above. Another alternative is to build the set of differences conditional on the direction $k_t$, i.e., $D(k) = \{\cc_{t + \Delta} - \cc_t\}_{t: k_t = k}$.
The sampling process is almost analogous to the rat hippocampus example above: We would sample a time step $t$, look up the discrete variable $k_t$, but then only sample from $D(k_t)$ to reflect the conditioning on the direction.

Finally, e.g., for very complex movements, it is simple to adapt additional pre-processing schemes. These could involve other deep learning algorithms like DINO used for pre-processing video data in the Allen dataset (to convert pixel data without a meaningful metric into an embedding space with desirable distance properties); they could also involve simpler processing, such as computing the principal component analysis of a higher dimensional dataset, and using the behavior data in this space.

Many other variations are possible. While the most common use cases are reflected in the CEBRA software toolbox and high-level API and readily usable, more customized use cases can be easily added by the user thanks to a straightforward extension mechanism.

\subsection*{Improving pi-VAE}
    Zhou et al~\cite{zhou2020learning} demonstrate that pi-VAE outperforms LFADS~\cite{Pandarinath2018InferringSN}, demixed-PCA~\cite{Kobak2016DemixedPC}, UMAP~\cite{mcinnes2018umap}, PCA, and pfLDS~\cite{Gao2016LinearDN} using the rat and/or primate datasets (Extended Data Fig. 1a). We improved the performance of pi-VAE by modifying the encoder, which allows for longer time inputs (Extended Data Fig.1), and this improved version is used throughout, unless noted.

\subsection*{Utilizing CEBRA across contexts}
    Within our framework, we assume that independent latent variables are combined by a non-linear bijective mixing function to produce neural activity. The latent variables are assumed to change over time, or be correlated to the observed auxiliary variables used to train CEBRA. No additional special structure, or implicit generative models during training are needed.

    CEBRA allows for minimizing the impact of selected features on the embedding, while testing the role of others. For example, suppose you have neural data from four different animals, each from the hippocampus while the animal navigated a linear track. You hypothesize that the hippocampus encodes a continuous mapping of space along the track. In this scenario the animal ID is not important, but the spatial location of the animal is. Here, the user can specify to obtain an embedding that is invariant to the animal ID, but should incorporate the position information. Another amendable scenario is a hypothesis-free, discovery-driven approach (akin to unsupervised clustering). Here too, CEBRA can be used, with only time as the input (Fig. 1). Collectively, CEBRA can be used for both visualization of data and latent-space based embedding of neural activity for downstream tasks like decoding.

    The flexibility in choosing different auxiliary variables during data analysis allows users to leverage the same algorithm for a variety of applications on a given dataset: Discovery-driven analysis by purely self-supervised learning with time-contrastive learning, hypothesis-driven analysis by comparing embedding quality derived from different behavioral variables, or replacing supervised decoding algorithms, e.g., in brain-machine-interface contexts.

\newpage
\section*{Supplementary Note 2}

Here we provide theoretical results for consistency and identifiability of models trained within the CEBRA framework. We proceed by showing properties of the InfoNCE loss (Prop.~\ref{prop:infonce-minimizer}), and use them as the basis for showing that encoders trained on this loss function will become bijective under mild assumptions (Prop.~\ref{prop:bijective}). We then revisit existing theory on contrastive learning, and show that CEBRA falls into a category of models for which we can obtain theoretical guarantees on both consistency (Prop~\ref{prop:consistency}) across different model runs and identifiability of the ground truth latent distribution for both the discovery-driven (time-contrastive) learning mode (Prop.~\ref{prop:discovery-driven-identifiability}) and the hypothesis-driven mode (Prop.~\ref{prop:hypothesis-driven-identifiability}).
Our results leverage theory by \citet{Hyvrinen2019NonlinearIU, wang2020understanding, Roeder2020, Zimmermann2021ContrastiveLI}.

It should be noted that while consistency results between model runs do not require strong assumptions about the underlying data generating process, understanding the relation between the embedding space given by CEBRA and the underlying \emph{ground truth data generating process} naturally requires such assumptions. However, compared to assumptions in generative models (e.g., VAEs), these assumptions concern the relationship between the ground-truth latent variables, rather than making statements about the signal space.

\subsection*{Notation, data generation, and learning algorithm}

We will use the notation presented in the Methods Section.
We additionally introduce the latents $\uu$ and $\vv$ underlying the samples $\xx$ and $\yy$.
We will interchangeably use the distributions $\marginal$, $p$ and $q$ for either the latents $\uu$ and $\vv$ or their respective samples $\xx$ and $\yy$ depending on their arguments (we will show after Proposition~\ref{prop:infonce-minimizer} that this treatment is also formally correct due to the training setup considered here).
Definitions, propositions and theorems adapted from other works are cited and denoted by upper-case letters and are otherwise adapted to our notation.

\begin{defi}[Data generating process and encoder]
    \label{def:learning-setup}
    Let $\uu \in \Ru, \vv \in \Rv$ denote latents corresponding to the samples $\xx \in \Rx$ and $\yy \in \Ry$ in the respective signal space which are generated according to two differentiable and injective mixing functions $\ggx: \Ru \mapsto \Rx$ and $\ggy: \Rv \mapsto \Ry$,
    \begin{equation}
        \xx = \ggx(\uu), \quad \yy = \ggy(\vv)
    \end{equation}
    and there exist optimal differentiable encoders $\ffx: \Rx \mapsto \RE$ and
    $\ffy: \Ry \mapsto \RE$ such that
    \begin{equation}
        \fx(\ggx(\uu))_i = u_i
        \quad
        \fy(\ggy(\vv))_j = v_j.
    \end{equation}
    We will refer to the composition of the data generators and the encoders as $\hhx = \ffx \circ \ggx$ and $\hhy = \ffy \circ \ggy$. 
    In setups where two models are trained on potentially different mixing functions, we denote the second data generator, feature encoder, and the composition of both as $\thhx = \tffx \circ \tggx$ and $\thhy = \tffy \circ \tggy$. 
\end{defi}

We consider a marginal distribution $\marginal(\cdot)$, the positive sample conditional distribution $p(\cdot | \cdot)$ and the negative conditional distribution $q(\cdot | \cdot)$.
Reference samples $\uu$ from the (true) latent space are mapped to signal space by the injective function $\gg$, positive/negative samples $\vv$ from the (potentially different latent space) are mapped to (a possibly different) signal space $\yy$ by the injective function $\gg'$.
The encoder $\ff$ is applied to $\xx$ and the encoder $\ff'$ is applied to $\yy$ to recover the respective latents underlying $\xx$ and $\yy$. The similarity measure is denoted as $\phi$ with $\ff(\xx)$ and $\ff'(\yy)$ as its arguments.
Note that $\phi$ does not need to be a fixed function and can also be parameterized by a learnable neural network.
As in the Methods, we use the shortcuts $\psi(\xx, \yy) = \phi(\ffx(\xx), \ffy(\yy))$ and additionally introduce $\psi(\uu, \vv) := \phi(\hhx(\uu), \hhy(\vv))$ without additional subscripts, as the desired shortcut will be clear from the context and its arguments.

Note that this is a very general setup. We typically would assume that the number of dimensions in the (shared) latent space is the same, $\dimu = \dimv$, and could further assume that the number of dimensions $\dime$ of the embedding space is also matching.

We recall the contrastive objective in the limit of unlimited samples from the main text:
\begin{defi}[Generalized InfoNCE objective]
    \label{def:infonce-asymptotic}
    In the limit of unlimited negative samples, the InfoNCE objective is a functional
    \begin{equation}
        \Linf = \int \marginal(\xx)
            \left[
            \log \int q(\yy | \xx) e^{\psi(\xx,\yy)} d\yy
            - \int p(\yy | \xx) \psi(\xx,\yy) d\yy
            \right] d\xx,
    \end{equation}
    depending on the positive sample conditional density $p(\yy | \xx)$, the negative sample density $q(\yy | \xx)$, the marginal density $\marginal(\xx)$ and the embedding similarity $\psi$ as defined above.
\end{defi}
We call this objective ``generalized'' as it extends the original definition of the InfoNCE loss used in the literature. \citet{oord2018representation} introduced an objective where the marginal (there: prior) $\marginal$ and the negative sample distribution $q$ matched, which influences the types of functions that can be learned.
The discussion of the objective by \citet{wang2020understanding} makes stronger assumptions about the nature of the conditional distribution $p$ for the positive pair, and only considers uniform choices for the marginal $\marginal$ and negative conditional $q$.

Overall, in CEBRA, the key difference to prior uses of the InfoNCE objective is the ability to control the properties of the embedding space through $p$ and $q$ and $\phi$, and leverage this to retrieve the discovery-driven, hypothesis-driven, and hybrid modes as demonstrated in the main text.
In fact, for hypothesis-driven training, the distributions used for sampling do not even need to be connected to the underlying data generating process---instead, varying $p$ and testing to enforce various neighbourhood relations on the neural data is used as a tool to discover meaningful relations between the auxiliary variables (e.g., behavior) and signal (e.g., neural activity).
In the same manner, $p$ and $q$ can be selected such that a particular factor is sampled uniformly, to enforce invariance (e.g. across a subject, or a modality variable).

The loss optimized in practice acts on a limited number of negative samples in each mini-batch:
\begin{defi}[Generalized InfoNCE objective with limited batch size]
    \label{def:infonce-batches}
    For a fixed number of negative samples $n$, the InfoNCE objective is the functional
    \begin{equation*}
      \Ln = 
      \mathop{\mathbb{E}}\limits_{{\substack{
            \xx \sim \marginal(\xx),\ 
            \yy_+ \sim p(\yy|\xx)\\
            \yy_1,\dots,\yy_n \sim q(\yy|\xx)\\
        }}}
        \left[ - \psi(\xx, \yy_+) + \log \sum_{i=1}^{n} e^{\psi(\xx, \yy_i)}
        \right],
    \end{equation*}
    depending on the positive sample conditional density $p(\yy | \xx)$, the negative sample density $q(\yy | \xx)$, the marginal density $\marginal(\xx)$ and the embedding similarity $\psi$ as defined above.
\end{defi}

Both losses can be related due to Theorem 1 by \citet{wang2020understanding}. In the limit of unlimited samples $n \to \infty$, we obtain for the batch size $n$:
\begin{equation}
    \Linf = \lim_{n \to \infty} \left( \mathcal{L}[\psi]_{n} - \log n \right).
\end{equation}
For a sufficiently large batch size, we can leverage the quantity $\mathcal{L}[\psi]_{n} - \log n$ as a goodness of fit measure (as outlined in the Methods) that estimates the distance from a ``default'' embedding.
When comparing models with equal batch size $n$, note that the InfoNCE loss can also directly serve as this metric.

\subsection*{Minimizers of the generalized InfoNCE loss}
\label{sec:infonce-minimizers}

In this section, we show that optimizing the generalized InfoNCE objective from Def.~\ref{def:infonce-asymptotic} yields the unique minimizer $\psi(\xx, \yy) = C(\xx) + \log p(\yy | \xx) / q(\yy | \xx)$ or equivalently $\psi(\uu, \vv) = C'(\uu) + \log p(\vv | \uu) / q(\vv | \uu)$ up to an arbitrary constant function $C'(\uu)$ than can depend on the latents of the reference latents. In this regard, the InfoNCE loss is more flexible than the standard noise constrastive estimation (NCE) loss which has a similar minimizer, but is limited to $C(\xx) = C'(\uu) = 0$.
The minimum loss value is the negative Kullbach-Leibler divergence between the positive and negative distributions. To obtain non-trivial solutions, it is therefore important that $p$ and $q$ differ in a non-trivial way (which is the case for both time-contrastive and behavior-contrastive sampling outlined in the context of CEBRA). 

\begin{prop}
    \label{prop:infonce-minimizer}
    Let $p(\cdot | \cdot)$ be the conditional distribution of the positive samples, $q(\cdot | \cdot)$ the conditional distribution of the negative samples and $\marginal(\cdot)$ the marginal distribution of the reference samples. 
    The generalized InfoNCE objective (Def.~\ref{def:infonce-asymptotic}) is convex in $\psi$ with the unique minimizer
    \begin{equation}
        \psi^*(\xx,\yy) = \log{\frac{p(\yy | \xx )}{q(\yy | \xx)}} + C(\xx), \quad \text{with} \quad
        \Linfopt = -\DKL(p(\cdot|\cdot) \| q(\cdot|\cdot))
    \end{equation}
    on the support of $\marginal$, where $C: \Ru \to \R$ is an arbitrary mapping.
    \begin{proof}
        We rewrite the objective as 
        \begin{equation}
            \Linf = \int \marginal(\xx)
                \left[
                \log \int q(\yy | \xx) e^{\psi(\xx,\yy)} d\yy
                - \int p(\yy | \xx) \psi(\xx,\yy) d\yy
                \right] d\xx,
        \end{equation}
        and we can compute the first-order functional derivative (using the method discussed in Cahill 2014\footnote{\url{http://quantum.phys.unm.edu/523-14/ch15.pdf}})
        \begin{equation}
        \begin{aligned}
            \delta \Linf[{[h]}] =& \frac{d}{d\epsilon} \mathcal{L}[\psi + \epsilon h] \big|_{\epsilon = 0} \\
            =& \int \marginal(\xx)
                \left[ 
                \frac{1}{Z_{\psi}(\xx)} \int q(\yy | \xx) e^{\psi(\xx,\yy)} h(\xx, \yy) d\yy
                - \int p(\yy | \xx) h(\xx,\yy) d\yy
                \right] d\xx, \\
            &\textrm{with } Z_{\psi}(\xx) = \int q(\yy' | \xx) e^{\psi(\xx,\yy')} d\yy'.
        \end{aligned}   
        \end{equation}
        The first-order functional derivative vanishes for all functions $h(\xx, \yy)$ whenever
        \begin{equation}
            \marginal(\xx) \left[ \frac{1}{Z_\psi(\xx)} q(\yy | \xx) e^{\psi(\xx,\yy)} - p(\yy | \xx) \right] = 0.
        \end{equation}
        This is the case iff at any point $(\xx, \yy)$, either $\marginal (\xx) = 0$ or 
        \begin{equation}\label{eq:firstderivative}
            Z_\psi(\xx) = 
            \frac{q(\yy | \xx)}{p(\yy | \xx)}  e^{\psi(\xx,\yy)}.
        \end{equation}
        Since the left hand side of Eq.~\eqref{eq:firstderivative} is independent of $\yy$, the right hand side must be independent of $\yy$ as well. Hence all functions $\psi^*(\xx,\yy)$ which are solutions to Eq.~\eqref{eq:firstderivative} are of the form
        \begin{equation}
            \psi^*(\xx,\yy) = \log{\frac{p(\yy | \xx )}{q(\yy | \xx)}} + C(\xx), 
        \end{equation}
        where $C$ is an arbitrary function depending only on $\xx$ and not on $\yy$.
        Then by definition of $Z_\psi(\xx)$,
        \begin{equation}
            Z_\psi(\xx) = \int q(\yy' | \xx) e^{\psi(\xx,\yy')} d\yy' = e^{C(\xx)}
        \end{equation}
        which is consistent when inserted into Eq.~\eqref{eq:firstderivative}. 
        Therefore the minimizers of $\Linf$ form a convex connected set $\mathcal{M}$,
        \begin{equation}
            \mathcal{M} = \Bigg\{ \psi^*: \psi^*(\xx,\yy) = \begin{cases}
            \log{\frac{p(\yy | \xx )}{q(\yy | \xx)}} + C(\xx) & \text{ if } \marginal(\xx) \ne 0 \\
            f(\xx,\yy) &  \text{ if } \marginal(\xx) = 0 \\ \end{cases}\Bigg\} \Bigg\},
        \end{equation}
        where $C, f$ are arbitrary functions. It can be checked that all minima achieve the same value $L[\psi^*]$ given by
        \begin{equation}
        \begin{aligned}
            \Linfopt =  - \int \marginal(\xx)  \int p(\yy | \xx) \log{\frac{p(\yy | \xx )}{q(\yy | \xx)}} d\yy  = - \int \marginal(\xx) D_{\textrm{KL}} \left[ p( \cdot | \xx) || q (\cdot | \xx) \right] d\xx \leq 0.
        \end{aligned}
        \end{equation}
    It is left to show that the objective function is convex. The second-order functional derivative is given by
    \begin{equation}
    \begin{aligned}
        \delta^2 \Linf[{[h]}] =& \frac{d^2}{d\epsilon^2} \mathcal{L}[\psi + \epsilon h] \big|_{\epsilon = 0} \\
        =& \frac{d}{d\epsilon} \int p(\xx)
            \left[
            \frac{1}{Z_{\psi + \epsilon h}(\xx)} \int  q(\yy | \xx) e^{\psi(\xx,\yy)  + \epsilon h(\xx,\yy)} h(\xx, \yy) d\yy
            - \int p(\yy | \xx) h(\xx,\yy) d\yy
            \right] \bigg|_{\epsilon = 0} \\
        =& \int p(\xx) \left[ \mathbb{E}_{g(\yy|\xx)} \left[ h(\xx, \yy)^2 \right]  - \mathbb{E}_{g(\yy|\xx)} \left[ h(\xx, \yy) \right]^2
        \right], \\
        &\textrm{with } g(\yy | \xx) = \frac{1}{Z_{\psi}(\xx)} q(\yy | \xx) e^{\psi(\xx,\yy)}, \text{ and } \int g(\yy | \xx) d\yy = 1.
    \end{aligned}   
    \end{equation}
    Since $g(\yy | \xx)$ is a probability density function, we can apply Jensen's inequality to the convex function $A \to A^2$ for the random variable $A := h(\xx, \yy)$ to obtain
    \begin{equation}
        \begin{aligned}
            &\mathbb{E}_{g(\yy|\xx)} \left[ h(\xx, \yy)^2 \right]  - \left( \mathbb{E}_{g(\yy|\xx)} \left[ h(\xx, \yy) \right] \right)^2  \geq 0 \quad 
            \Rightarrow \delta^2 \Linf[{[h]}] = \int p(\xx) \left[ \mathbb{E}_{g(\yy|\xx)} \left[ h(\xx, \yy)^2 \right]  - \left( \mathbb{E}_{g(\yy|\xx)} \left[ h(\xx, \yy) \right] \right)^2 
        \right] \geq 0.
        \end{aligned}
    \end{equation}
    and it follows that the InfoNCE loss is convex in $\psi$.

To prove uniqueness of the minimum: 
    Note that since the mapping $A \to A^2$ is not affine, a necessary and sufficient condition for equality is $A$ to be constant, which holds iff $h(\xx, \yy) = h(\xx, \yy') := h(\xx)$ for all $\yy$. The objective is hence strictly convex for all variations involving a variation in $\yy$, and the second derivative vanishes for variations that only depend on $\xx$. Variations that only depend on $\xx$ are represented by the function $C$ which appeared in the set of minimizers $\mathcal{M}$.
    \end{proof}
\end{prop}

Note that the difference between the minimizer of the NCE loss and the InfoNCE loss is the additional constant function $C$ depending on the reference sample, which makes the loss function more flexible.
Another common formulation of the InfoNCE minimizer in the literature is given as
$\psi(\xx, \yy) = \log p(\xx, \yy) / (\marginal(\xx) \marginal(\yy))$ which is a special case of our more general solution if $C(\xx) = -\log \marginal(\xx)$, the negative distribution is chosen to be the marginal, $q = \marginal$, and the learning setup is symmetric.

Let us also confirm that our interchangeable use of the latents $(\uu, \vv)$ and signal variables $(\xx, \yy)$ is formally correct; due to the transformation theorem for any distribution
$p_\uu(\uu) = p_\xx(\gg(\uu))  \det \JJ_\gg(\uu)$ and respectively for $\vv, \ggy, \yy$. At the minimizer, we then arrive at
\begin{align}
    \log \frac{p(\yy | \xx)}{q(\yy | \xx)} + C(\xx)
    &= \log \frac{p(\yy | \xx) \marginal(\xx)}{q(\yy | \xx)\marginal(\xx)} + C(\xx) \\
    &= \log \frac{p(\vv | \uu) \marginal(\uu) \det \JJ_\gg(\uu)  \det \JJ_{\gg'}(\vv)}{q(\vv | \uu)\marginal(\uu) \det \JJ_\gg(\uu)  \det \JJ_{\gg'}(\vv)} + C'(\uu) \\
    &= \log \frac{p(\vv | \uu)}{q(\vv | \uu)} + C'(\uu), \quad \text{with } C(\gg(\uu)) = C'(\uu),
\end{align}
i.e., the minimizer can be equivalently written in terms of the latents and the signal variables.

\subsection*{Minimizers of the InfoNCE loss become bijective}
\label{sec:infonce-bijective}

A property that allows us to weaken some of the conditions given by \citet{Hyvrinen2019NonlinearIU}, \citet{Zimmermann2021ContrastiveLI} and \citet{Roeder2020} is the observation that the composition of data generating process and feature encoder becomes bijective for the optimal value of the generalized InfoNCE objective. We introduce the following:
\begin{defi}[Diversity condition for bijectivity]
    \label{def:div-condition-bijective}
    The sampling process composed of distributions $p$ and $q$ is sufficiently diverse if their log-likelihoods satisfy
    \begin{equation}
        \rank \left( \left[
            \frac{\partial^2 \log p(\vv | \uu)}{\partial u_i \partial v_j} 
        \right]_{i \in [\dimu], j\in[\dimu]}
        - \left[
            \frac{\partial^2 \log q(\vv | \uu)}{\partial u_i \partial v_j} 
        \right]_{i \in [\dimu], j\in[\dimu]} \right) = \dimu,
    \end{equation}
    for all $\uu$ in the support of the marginal distribution $\marginal$ and $\dimu = \dimv$.
\end{defi}
Def.~\ref{def:div-condition-bijective} is a mild condition on the distributions $p$ and $q$: Intuitively, the condition requires that for all samples $\uu$, we can sample positive samples $\vv$ that sufficiently vary in all $\dimu$ latent directions, which would be independent from the samples given by the negative distribution $q$.
Suppose $q$ is chosen to be uniform; then the condition is fulfilled for common choices like a Normal distribution with $\log p(\vv | \uu) = Z(\uu) - (\uu - \vv)^\top \boldsymbol\Sigma (\uu - \vv)$ (where $\rank (-\boldsymbol\Sigma) = d$) or a von Mises-Fisher distribution with $\log p(\vv | \uu) = Z(\uu) + \kappa \uu^\top \vv$ (where $\rank \kappa\II = d$). 

We make two additional observations: Firstly, for simple distributions $q$ that do not depend on $\uu$, the diversity assumption only affects the positive distribution $p$ as the second term vanishes. Secondly, if $p$ and $q$ are selected to train the network to become invariant to one factor $v_i$ with
    $p(\vv | \uu) = p(v_i) p(\vv_{\\i} | \uu)$ and
    $q(\vv | \uu) = p(v_i) q(\vv_{\\i} | \uu)$,
the distributions $p(v_i)$ will cancel out in the condition, and reduce the rank by one dimension (which is as intended, as the factor should be discarded during training).

From this diversity condition, we can derive bijectivity of the composition $\hhx = \ffx \circ \ggx$ of the data generating process and feature encoder:
\begin{prop}
    \label{prop:bijective}
    Assume that:
    \begin{enumerate}
        \item $\psi$ with $\psi(\uu, \vv) = \phi(\hhx(\uu), \hhy(\vv))$ is a minimizer of the InfoNCE objective (Def.~\ref{def:infonce-asymptotic}) in a learning setup as outlined in Def~\ref{def:learning-setup}.
        \item The distributions $p$ and $q$ satisfy the diversity condition for bijectivity (Def.~\ref{def:div-condition-bijective}).
    \end{enumerate}
    Then $\hhx$ and $\hhy$ are bijective on the support of $\marginal$.
        
    \begin{proof}
    By Proposition~\ref{prop:infonce-minimizer}, the minimizer of the InfoNCE loss on the support of $\marginal$ is
    \begin{equation}
        \label{eq:ref-minimizer}
        \psi(\uu, \vv)
        = \log \frac{p(\vv | \uu)}{q(\vv | \uu)} + C(\uu)
    \end{equation}
    For $\psi$, we compute the second derivatives and arrange them in matrix form as
    \begin{equation}
        \left[ \frac{\partial^2 \psi(\uu, \vv)}{\partial u_i \partial v_j} \right]_{i \in [d],j \in [d]} = \JJ^\top(\uu) \PP(\hhx(\uu), \hhy(\vv)) \JJ'(\vv)
    \end{equation}
    where we used the shorthand $\PP(\aa, \bb)_{ij} := \partial^2 \phi(\aa, \bb) / \partial \aa_i \bb_j$. $\JJ$ is the Jacobi matrix of $\hhx$, and $\JJ'$ is the Jacobi matrix of $\hhy$.
    For the right hand side of the previous equation, we note that
    \begin{equation}
        \rank(\JJ^\top(\uu) \PP(\hhx(\uu), \hhy(\vv)) \JJ'(\vv)) \le \min\{ \rank \JJ(\uu), \rank \JJ'(\vv), \rank \PP(\hhx(\uu), \hhy(\vv)) \},
    \end{equation}
    and the rank of the left hand side is given by inserting Eq.~\ref{eq:ref-minimizer} into the diversity assumption (2):
    \begin{equation*}
        \rank
            \left[ \frac{\partial^2 \psi(\uu, \vv)}{\partial u_i \partial v_j} \right]_{i \in [d],j \in [d]}
        = \rank \left( \left[
            \frac{\partial^2 (\log p(\vv | \uu) - \log q(\vv | \uu) + C(\uu))}{\partial u_i \partial v_j} 
        \right]_{i \in [d], j\in[d]} \right)
        = d.
    \end{equation*}
    Combining both results gives 
    \begin{equation}
        \rank(\JJ^\top(\uu) \PP(\hhx(\uu), \hhy(\vv)) \JJ'(\vv)) = d \le \min\{ \rank \JJ(\uu), \rank \JJ'(\vv), \rank \PP(\hhx(\uu), \hhy(\vv)) \},
    \end{equation}
    and implies
    \begin{align}
        \rank \JJ(\vv) = \rank \JJ'(\uu) = \rank \PP(\hhx(\uu), \hhy(\vv)) = d.
    \end{align}
    Then, both Jacobi matrices have full rank on the support of $\marginal$, hence $\hhx$ and $\hhy$ are bijective, concluding the proof.
    \end{proof}
\end{prop}

Notably, this result is independent of the particular choice of the (potentially learnable) similarity measure $\phi$. 
The similarity measure is implicitly constrained by the requirement that $\psi$ needs to match the log-likelihood ratio of $p$ and $q$ up to a constant.

\subsection*{CEBRA models are consistent}
\label{sec:consistency}

We proceed by showing that CEBRA models are \emph{consistent} under weak assumptions on the data distribution.
Consistency entails that the embedding spaces of two different models can be mapped onto each other by some known transformation.
In this subsection, we consider the class of \emph{linear} transformations and in the following subsection we will discuss alternative transformations.
We denote two independently trained CEBRA models as $\{\ffx, \ffy\}$ and $\{\tffx, \tffy\}$, and make statements about when linear transformations exist such that $\ffx = \LL \tffx$ and $\ffy = \MM \tffy$ for two full rank matrices $\LL$ and $\MM$.

We begin by recalling the Canonical Discriminative Form and Diversity Condition in \citet{Roeder2020}, adapted to our notation:

\begin{customdef}{A}[Canonical Discriminative Form, \citet{Roeder2020}]
    \label{def:canonical-discriminative-form}
    Given a data distribution $\marginal(\xx, \yy, S)$ with random variables $\xx$ and $\yy$ and a set $S$ containing the possible values of $\yy$ given $\xx$,
    \begin{equation}
        \marginal(\yy | \xx, S) > 0 \Longleftrightarrow \yy \in S,
    \end{equation}
    a generalized discriminative model family
    may be defined by its parameterization of the probability of
    the target variable $\yy$ conditioned on an observed variable $\xx$ and the set $S$ that contains not only the true target label $\yy$, but also a collection of distractors $\yy'$:
    \begin{equation}
        p_{\ff, \ff'}(\yy|\xx, S)
            = \frac{\exp(\ff(\xx)^\top \ff'(\yy))}
                   {\sum_{\yy' \in S} \exp(\ff(\xx)^\top \ff'(\yy'))}.
    \end{equation}
\end{customdef}

Note that the feature extractors $\ff$ and $\ff'$ could be two separate networks, as we already discussed in Suppl. Note 1.
\citet{Roeder2020} consider $\ff$ to be a ``data encoder'' and $\ff'$ to be a ``context encoder''. This is in contrast to the setup by \citet{Hyvrinen2019NonlinearIU} which we will revisit in the context of recovering the data generating factors, where $\ff(\xx)$ would play the role of an auxiliary variable while $\ff'$ is the feature encoder later used in downstream tasks and analysis.

In CEBRA, the choice and role of both functions can be configured, which is why we will discuss theoretical guarantees for both use cases. We proceed by re-stating two diversity conditions needed for $\ff$ or $\ff'$ to become consistent:

\begin{customdef}{B}[Diversity conditions for consistency, \citet{Roeder2020}]\label{def:diversity}
Let $Z(\xx, S) := -\log\sum_{\yy' \in S}\exp(\ff(\xx)^\top \ffy(\yy'))$.
Assume that for the encoders $(\ffx, \ffy, \tffx, \tffy)$ for which it holds that $p_{\ffx, \ffy} = p_{\tffx, \tffy}$
\begin{enumerate}
    \item for any given $\yy$, there exist $M + 1$ tuples $\{(\xx^{(i)}, S^{(i)})\}_{i = 1}^{M+1}$, such that $\marginal(\xx^{(i)}, \yy, S^{(i)}) > 0$, and such that the $((M + 1) \times (M+ 1))$ matrices $\MM$ and $\tilde \MM$ are invertible, where $\MM$ consists of columns $[-Z(\xx^{(i)}, S^{(i)}); \ffx(\xx^{(i)})]$, and $\tilde \MM$ consists of columns $[-Z(\xx^{(i)}, S^{(i)}); \tffx(\xx^{(i)})]$,
    \item for any given $\xx$, by repeated sampling $S \sim p_D(S|\xx)$ and picking two points $\yy_A, \yy_B \in S$, we can construct a set of $M$ distinct tuples ${(\yy_A^{(i)}, \yy_B^{(i)})}^M_{i=1}$ such that the matrices $\LL$ and $\tilde \LL$ are invertible, where $\LL$ consists of columns $(\ffy(\yy_A(i)) - \ffy(\yy_B(i)))$, and $\tilde \LL$ consists of columns $(\tffy(\yy_A(i)) - \tffy(\yy_B(i))$, $i \in {1, \dots , M}$.
\end{enumerate}
\end{customdef}

With the diversity condition in place, we recall Theorem 1 from \citet{Roeder2020}, adapted to our notation:

\begin{customprop}{A}[\citet{Roeder2020}]
    \label{ref:roeder-consistency}
    Under the diversity condition (Def.~\ref{def:diversity}), models following the canonical discriminative form (Def.~\ref{def:canonical-discriminative-form}) are linearly identifiable. That is, for any encoders $\{\ffx, \ffy\}$, and $\{\tffx, \tffy\}$ it holds that
    \begin{equation}
        p_{\ffx, \ffy} = p_{\tffx, \tffy} \quad\Longrightarrow\quad
            \ffx(\xx) = \LL \tffx(\xx), \quad
            \ffy(\yy) = \LL \tffy(\yy)
    \end{equation}
    for all samples ($\xx$, $\yy$) in the support of the data distribution.
    \begin{proof}
        See Theorem 1, \citet{Roeder2020}, where we replaced the equivalence condition on the right hand side by inserting its definition for clarity.
    \end{proof}
\end{customprop}

We can leverage this result as CEBRA falls into the class of models models described in Definition~\ref{def:canonical-discriminative-form}:

\begin{prop}[CEBRA models are consistent.]
    \label{prop:consistency}
    Assume that two CEBRA models are trained on data from the same latent data distribution, and denote the feature encoders of the trained models as $\ffx, \ffy$ and $\tffx, \tffy$.
    Further assume that for both models, the similarity measure $\phi$ is the dot-product similarity and $\psi$ minimizes the generalized InfoNCE loss.
    Finally assume that the diversity condition (Def.~\ref{def:diversity}) holds.
    Then the feature encoders $\ffx$, $\ffy$ are consistent up to a linear transformation, and
        $\ffx(\xx) = \mathbf{L} \tffx(\xx)$,
        $\ffy(\yy) = \mathbf{L'} \tffy(\yy)$,
    for linear transformations $\LL$, $\LL'$ for any pair of points $\xx, \yy$ in the support of the data distribution.
    
    \begin{proof}
        The generalized InfoNCE objective with limited samples (Def.~\ref{def:infonce-batches}) can be written as
        \begin{equation}
            \mathop{\mathbb{E}}\limits_{{\substack{
                \xx \sim p(\xx),\ 
                \yy_+ \sim p(\yy|\xx)\\
                \yy_1,\dots,\yy_n \sim q(\yy|\xx)\\
            }}} [\log p(\yy^+ | \xx, S) ], \quad
            \log p(\yy^+ | \xx, S)
            = 
            - \log \frac{\exp \psi(\xx, \yy_+)}{\sum_{i=1}^{n} \exp {\psi(\xx, \yy_i)}}
            =
            - \log \frac{\exp (\ffx(\xx)^\top \ffy(\yy_+) / \tau)}{\sum_{i=1}^{n} \exp (\ffx(\xx)^\top \ffy(\yy_i) / \tau)}
        \end{equation}
        which matches the canonical discriminative form (Def.~\ref{def:canonical-discriminative-form}) with encoders $\ffx(\cdot)$ and $\ffy(\cdot)/\tau$.
        The composition of the set $S := \{\yy_i\}_{i = 1}^{N}$ is given by the distribution $\marginal(\xx)q(\yy | \xx)$ and $S$ fulfills Def.~\ref{def:diversity} by assumption.
        At the minimizer, the values of the loss functions match, from which it follows that $p_{\ffx, \ffy} = p_{\tffx, \tffy}$.
        Hence, we apply Theorem~\ref{ref:roeder-consistency} to find that $\ffx$ and $\ffy$ are consistent up to a linear transform, concluding the proof.
    \end{proof}
\end{prop}

It is worth noting that this result also \emph{holds for datasets with limited samples}, i.e., the objective in Def.~\ref{def:infonce-batches}, as long as the dataset fulfills the diversity condition (Def.~\ref{def:diversity}). Checking the diversity condition as well as matching distributions for the two CEBRA models is possible in practice using only the dataset and the trained model.

Both diversity conditions from \citet{Roeder2020} depend on the variability of the ground truth latent distribution (and the presence of this variability after mapping the latents to signal space), and on the properties of the encoders $\ff$ and $\ff'$.
While \citet{Roeder2020} already discuss that in practice, a randomized neural network will fulfill the diversity conditions, with our diversity criterion for bijectivity (Def.~\ref{def:div-condition-bijective}) and the bijectivity of $\ff$ and $\ff'$ that follows, we can strengthen this argument and ensure that both conditions hold upon convergence for minimizers of the generalized InfoNCE objective:

\begin{prop}
    Assume that the encoders $(\ffx,\ffy,\tffx,\tffy)$ and hence also the compositions of data generator and encoders $(\hhx,\hhy,\thhx,\thhy)$ minimize the InfoNCE objective.
    Assume that upon convergence, the diversity condition for bijectivity (Def.~\ref{def:div-condition-bijective}) holds.
    Then, $\hh(\uu) = \AA \hh(\uu)$ and $\hhy(\vv) = \BB \thhy(\vv)$ for all latents $\uu, \vv$ in the support of the data distribution for two full-rank matrices $\AA$, $\BB$.
    \begin{proof}
        Both models share the minimizer
        \begin{equation}
            \hhx(\uu)^\top \hhy(\vv) = \thhx(\uu)^\top \thhy(\vv) = \log \frac{p(\vv | \uu)}{q(\vv | \uu)} + C(\uu)
        \end{equation}
        and we have
        \begin{align}
            \hhx(\uu)^\top \hhy(\vv) &= \thhx(\uu)^\top \thhy(\vv)\\
            \intertext{taking derivatives with respect to $\vv$, and then to $\uu$, we arrive at}
            \JJ(\uu)^\top \JJ'(\vv)  &= \tilde \JJ(\uu)^\top \tilde \JJ'(\vv)\\
            \intertext{where all Jacobian matrices have full rank due to Prop.~\ref{prop:bijective}. We can hence derive}
             \JJ'(\vv) = (\JJ(\uu)^{-\top} \tilde \JJ(\uu)^\top) \tilde \JJ'(\vv)
             &\quad
             \JJ(\uu)^\top  = \tilde \JJ(\uu)^\top (\tilde \JJ'(\vv) \JJ'(\vv)^{-1})\\
              \JJ'(\vv) = \AA(\uu) \tilde \JJ'(\vv)
             &\quad
             \JJ(\uu)^\top  = \tilde \JJ(\uu)^\top \BB(\vv)\\
            \intertext{for some full rank matrices $\AA(\uu)$ and $\BB(\vv)$. Because the left hand side do not depend on the argument of $\AA$ and $\BB$, both matrices need to be constant, leaving}
              \JJ'(\vv) = \AA \tilde \JJ'(\vv)
             &\quad
             \JJ(\uu)^\top  = \tilde \JJ(\uu)^\top \BB\\
             \intertext{from which it follows that}
             \hhx(\uu) = \BB^{-1} \thhx(\uu) 
             &\quad
             \hhy(\vv) = \AA \thhy(\vv). \label{eq:prop3-final-result}
        \end{align}
        concluding the proof.
    \end{proof}
\end{prop}

Note that the previous proposition applies even when the data generating functions (i.e., the datasets) between the model run differ, and merely the latent distributions match. In this case, the same latent $\uu_0$ can be mapped to different points $\xx_0$ and $\tilde \xx_0$ and the resulting embedding points would still satisfy $\ffx(\uu_0) = \AA \tffx(\uu_0)$. For this reason, the proposition is written w.r.t. the composition $\thhx$ of data generating process and encoder.
If the data generating processes match, it is clear that Eq.~\ref{eq:prop3-final-result} can be equivalently written as $\ffx(\xx) = \BB^{-1} \tffx(\xx)$, $\ffy(\yy) = \AA \tffy(\yy)$, which matches the statement by \citet{Roeder2020} (but for our modified diversity condition).

For symmetric encoders, we can give the following result:
\begin{prop}
    Assume that the encoders $\ffx=\ffy, \tffx=\tffy$ are shared, and $-\phi$ is a norm, $\phi(\aa, \bb) = - \| \aa - \bb \|$. Assume that the model minimizes the InfoNCE loss and assume that the co-domain of $\ffx$, $\ffy$ is a normed space over $\Ru$. Then $\hhx = \LL \thhx$.
    \begin{proof}
        We write the data in terms of the underlying latents $\xx = \ggx(\uu)$ and $\yy = \ggy(\vv)$. At the minimizer of the InfoNCE loss it then holds that
        \begin{equation}
            \| \hhx(\uu) - \hhx(\vv) \| + C(\uu) = \| \thhx(\uu) - \thhx(\vv) \| + \tilde C(\uu).
        \end{equation}
        for all points in the dataset. Inserting $\vv = \uu$ gives $C(\uu) = \tilde C(\uu)$. Because $\hhx, \thhx$ bijective, we define points $\aa = \thhx(\uu)$ and $\bb = \thhx(\vv)$, and it holds
        \begin{equation}
            \| \hhx(\thhx^{-1}(\aa)) - \hhx(\thhx^{-1}(\bb)) \| = \| \aa - \bb \|.
        \end{equation}
        Due to the Mazur–Ulam theorem, the map $\hhx \circ \thhx^{-1}$ is then affine, concluding the proof.
    \end{proof}
\end{prop}

Note that all results in this section are also independent of the mixing functions $\ggx$ and $\ggy$. This means that \emph{consistency can be guaranteed irrespective of the exact data modality and generative process}, as long as the \emph{underlying} latent distribution matches. This is given in well-controlled experiments (as one assumes when e.g., repeating experiments).

\subsection*{CEBRA models recover the ground-truth latents}
\label{sec:ground-truth-model}

In contrast to the identifiability results in the previous section (which made a connection between two models trained on the same or similar data distribution), in this section we will make a connection between the \emph{ground truth model} and the model trained on the data. To make this connection, assumptions are needed about the ground truth model.

As introduced in Def.~\ref{def:learning-setup}, we will denote the ground truth model(s) as $\ggx$ and $\ggy$, and data from these models is then encoded with $\ffx$ and $\ffy$, respectively. Our goal is to understand properties of their composition $\hhx = \ggx \circ \ffx$ and $\hhy = \ggy \circ \ffy$, and when this composition reduces to an affine or linear transformation.
Based on the property of the model (especially the similarity measure), other guarantees are possible. Towards the end of this subsection we will discuss CEBRA model setups where we obtain guarantees up to permutations and sign-flips and point-wise non-linear transformations by leveraging existing contrastive learning theory \cite{Hyvrinen2019NonlinearIU,Zimmermann2021ContrastiveLI}.
We base our theory on the identifiability proofs of contrastive learning given by \citet{Hyvrinen2019NonlinearIU} and \citet{Zimmermann2021ContrastiveLI}, and give new proofs to complete the theory for the most important usage modes in CEBRA.
Note that the theory also extends to settings not explicitly demonstrated in this paper, but integrated into the CEBRA software package (e.g., trainable similarity functions $\phi$).

One key property of CEBRA is its distinction of discovery-driven and hypothesis-driven training (or hybrid training, which is a combination of both where the feature encoders need to minimize both the time-contrastive and behavior-contrastive objectives).
For time- and behavior-contrastive learning, similar theory applies. A key difference is that in time-contrastive (discovery-driven) learning, an underlying distribution of the latents is inherent to the dataset and ``given'' by the temporal variation in the data. If it is desirable to recover the ground truth latents, the similarity measure needs to be suitable to allow full InfoNCE minimization, e.g., by model selection on basis of the InfoNCE or goodness of fit metric. Note that a cosine similarity measure is already quite flexible for a wide range of distributions and a good default, and also note that InfoNCE minimization is known to be empirically robust to minor violations between the model assumptions and ground-truth conditional distribution \citep{Zimmermann2021ContrastiveLI}.
For hypothesis-driven learning, CEBRA will always be able to find a suitable learning setup that recovers the auxiliary variables: The positive and negative sample distributions can be chosen and selected such that the propositions in this section will hold. Even if empirical distributions are used (e.g., the ``time delta'' distribution outlined in the Methods), it is possible to check that this distribution matches the similarity measure of the model.

Because we make statements about the relationship between continual and discrete context variables ($\cc_t$ and $k_t$) and the signal space ($\ss_t$) for each time-point $t$, we will again use this notation from the Methods; we use the variable names introduced in Def.~\ref{def:learning-setup} to denote the underlying ground-truth latents, and the samples fed to the model (which can, but do not necessarily need to, equal to the signal).
We start our discussion with discovery-driven training of CEBRA using time information:

\begin{prop}[Discovery-driven CEBRA]
    \label{prop:discovery-driven-identifiability}
    Assume the learning setup in Def.~\ref{def:learning-setup}, and that the ground-truth latents $\uu_1,\dots,\uu_T$ for each time point follow a uniform marginal distribution and the change between subsequent time steps is given by the conditional distribution of the form
    \begin{equation}
        p(\uu_{t + \Delta t} | \uu_{t}) = \frac{1}{Z(\uu_{t})}  \exp \delta (\uu_{t + \Delta t}, \uu_{t})
    \end{equation}
    where $\delta$ is either a (scaled) dot product (and $\uu_t \in \mathcal{S}^{n-1} \subset \R^\dimu$ lies on the $(n-1)$-sphere $\mathcal{S}^{n-1}$) or an arbitrary semi-metric (and $\uu_t \in \mathcal{U} \subset \Ru$ lies in a convex body $\mathcal{U}$).
    Assume that the data generating process $\gg$ with $\ss_t = \gg(\uu_t)$ is injective.
    Assume we train a symmetric CEBRA model with encoder $\ffx=\ffy$ and the similarity measure including a fixed temperature $\tau > 0$ is set to or sufficiently flexible such that $\phi = \delta$ for all arguments.
    Then $\hh = \hh' = \gg \circ \ff$ is affine.
    \begin{proof}
        For $\delta$ being the dot product, the result follows from the proof of Theorem 2 in \citet{Zimmermann2021ContrastiveLI}.
        For $\delta$ being a semi-metric, the result follows from the proof of Theorem 5 in \citet{Zimmermann2021ContrastiveLI}. 
    \end{proof}
\end{prop}

We will next consider the hypothesis-driven mode in CEBRA. Here, we either choose a parametric or non-parametric positive distribution $p$ to shape the embedding space. Our goal is find an embedding space reflecting the auxiliary variable, in case there is a meaningful relationship between the signal and this variable.

Naturally, the actual signal will depend on additional latents that we do not record as auxiliary information. The full data generating process can be written as
\begin{equation}
    \ss_t = \gg(\cc_t, k_t, \zz_t)
\end{equation}
where $\zz_t$ are additional latent sources not observed during training. Since $\gg$ is an injective function, it follows that $\ss_t = \ss_{t'}$ implies $\cc_t = \cc_{t'}$, $k_t = k_{t'}$, $\zz_t = \zz_{t'}$. Applying hypothesis-driven learning in this setup will force an embedding space representing $\cc$ and $k$, but not $\zz$. We can denote the set $G(\cc, k) := \{\zz \in Z: \gg(\cc, k, \zz)\}$ which contains all points in signal space corresponding to a particular set of auxiliary variables $\cc, k$. Because $\gg$ is injective, there exists a function $\tilde \gg$, which will retrieve the original auxiliary variables: $\tilde \gg(\gg(\cc, k, \zz)) = (\cc, k)^\top$.

\begin{prop}[Hypothesis-driven CEBRA]
    \label{prop:hypothesis-driven-identifiability}
    Assume a partially observable data generating process, where
    \begin{equation}
        \yy = \ggy(\cc, \vv)
    \end{equation}
    with $\ggy$ injective and with $\cc$ as an observable context variable, and $\vv$ as a latent.
    As in Prop.~\ref{prop:discovery-driven-identifiability} assume that $\cc$ lies on a hypersphere if $\phi$ is the dot-product similarity, and on a convex body if $-\phi$ is a semi-metric.
    Then the minimizer of the InfoNCE loss trained with a distribution $p(\cc | \tilde \cc) = \exp(\phi(\cc, \tilde \cc)) / Z(\tilde \cc)$ and a uniform marginal and negative distribution $q = \marginal$ will yield a $\hh$ that recovers $\cc$ up to an affine transformation.
    \begin{proof}
        $\hh$ is the composition $\gg \circ \ff$. 
        Per assumption, the similarity function $\phi$ is sufficiently flexible such that
        \begin{equation}
            \label{eq:hyp-eq-minimizer}
            \phi(\hh(\cc, \zz), \hh(\tilde \cc, \zz)) = \log p(\cc, \tilde \cc) + C(\tilde \cc) \quad \forall \zz.
        \end{equation}
        Because $\gg$ is injective, $\gg(\cc, \zz) = \gg(\cc', \zz')$ implies that $\cc = \cc'$ and $\zz = \zz'$ and there exists a function $\tilde \gg$ such that $\tilde \gg (\gg(\cc, \zz)) = \cc$ for all $\zz$.
        It remains to show that $\ff = \LL \tilde \gg$ at the minimizer for a full-rank matrix $\LL$.
        By comparing arguments on the left hand side and right and side, and inserting the form of $p$, solutions of Eq.~\ref{eq:hyp-eq-minimizer} need to simplify to
        \begin{equation}
            \phi(\hh(\cc), \hh(\tilde \cc)) = \phi(\cc, \tilde \cc) + C(\tilde \cc)
        \end{equation}
        By symmetry of $\phi$, $C(\tilde \cc) = 0$ vanishes.
        Then the result follows again from Theorem 2 in \citet{Zimmermann2021ContrastiveLI} for $\phi$ being the dot-product similarity, and from Theorem 5 in \citet{Zimmermann2021ContrastiveLI} for $-\phi$ being a semi-metric.
    \end{proof}
\end{prop}

With our results from Prop.~\ref{prop:infonce-minimizer} and Prop.~\ref{prop:bijective}, it is also possible for us to extend the results for distributions fulfilling the diversity condition for bijectivity (Def.~\ref{def:div-condition-bijective}). The following proposition enables a statement about distribution where the data generating process and/or feature encoder differs between the reference and positive/negative pairs.
An example used in the main text is multi-session training, or training across data modalities.
We first need to strengthen the assumption about the positive and negative distributions from Def.~\ref{def:div-condition-bijective}.
Again assume that the latent dimensions $\dimu = \dimv$ match.

\newcommand{\dimuv}{\dimu}
\newcommand{\Ruv}{\Ru}
\begin{defi}[Strong assumption of diversity for bijectivity]
    \label{def:constant-second-derivative}
    The sampling process with distributions $p$ and $q$ satisfies
    \begin{equation}
        \left[
            \frac{\partial^2 \log p(\vv | \uu)}{\partial u_i \partial v_j} 
        \right]_{i \in [\dimu], j\in[\dimuv]}
        - \left[
            \frac{\partial^2 \log q(\vv | \uu)}{\partial u_i \partial v_j} 
        \right]_{i \in [\dimu], j\in[\dimuv]} = \LL
    \end{equation}
    for some full-rank matrix $\LL \in \R^{\dimu \times \dimuv}$ for all $\uu \in \Ru$ in the support of the marginal distribution $\marginal$.
\end{defi}

Note that Def.~\ref{def:constant-second-derivative} is a special case of the diversity condition for bijectivity in Def.~\ref{def:div-condition-bijective}.
The stronger condition still covers important distributions like a Gaussian conditional, even with an additional mean and covariance $p(\vv | \uu) = \exp(- (\uu - \vv - \mu)^\top \boldsymbol{\Sigma}^{-1}(\uu - \vv - \mu) )$.
Likewise, it would cover van Mises-Fisher (vMF) distributions, even with a rotated reference sample and 
$p(\vv | \uu) = \exp(\kappa \uu^\top \QQ \vv)$.
We will cover these selected special cases using the stronger diversity assumption below.

\begin{prop}
\label{prop:dot-product-identifiability}
Consider a data generating process with uniform marginal and positive/negative distributions satisfying Def.~\ref{def:constant-second-derivative}.
Assume the data is generated by mixing functions $\ggx: \Ru \mapsto \Rx$ and $\ggy \in \Ruv \mapsto \Ry$ and encoded into a shared $\dime$-dimensional embedding space with two separate encoders $\ffx: \Rx \mapsto \RE$ and $\ffy: \Ry \mapsto \RE$ and assume that $\phi$ is the dot-product.
Then there are $\dimu$ dimensions in $\hhx$, and $\dimuv$ dimensions in $\hhy$ which represent the latents up to an affine transform.
\begin{proof}
Let us again denote the compositions $\hhx = \ggx \circ \ffx: \Ru \mapsto \RE$ and $\hhy = \ggy \circ \ffy: \Ruv \mapsto \RE$ and let us denote the Jacobian matrices as $\JJ: \Ru \mapsto \R^{\dime \times \dimu}$ and $\JJ': \Ruv \mapsto \R^{\dime \times \dimuv}$, respectively.  
Without loss of generality (the indices can be arbitrarily permuted), let us split each $\hhx$ into two parts, with $\hhx_1 := [h_1,\dots,h_{\dimu}]^\top$ and $\hhx_2 := [h_{\dimu+1},\dots,h_{\dime}]^\top$ (in case $\dime > \dimu$), and respectively for $\hhy$. 
At the minimizer of the InfoNCE loss, we get the condition 
\begin{align}
    \hhx(\uu)^\top \hhy(\vv) / \tau &= \log \frac{p(\vv | \uu)}{q(\vv | \uu)} + C(\uu). \label{eq:minimizer-prop-6-condition}
    \intertext{We take the derivative w.r.t. $\vv$ on both sides, with gives}
    \JJ'(\vv)^\top \hhx(\uu) / \tau &= \frac{\partial}{\partial \vv} \left[ \log \frac{p(\vv | \uu)}{q(\vv | \uu)} \right],
    \intertext{with $\JJ'$ denoting the Jacobian matrix of $\hhy$.
        We now take the derivative w.r.t. $\uu$ on both sides, with gives}
    \JJ(\uu)^\top \JJ'(\vv) / \tau &= \frac{\partial^2}{\partial \uu \partial \vv} \left[ \log \frac{p(\vv | \uu)}{q(\vv | \uu)} \right].
    \intertext{and by assumption we can insert Def.~\ref{def:constant-second-derivative} and re-arrange to}
    \JJ(\uu) \JJ'(\vv)^\top &= \tau \LL 
\end{align}
for some full-rank matrix $\LL \in \R^{\dimu \times \dimuv}$.
Note that when $\dime > \dimu$
this condition is underconstrained. Without loss of generality, let us split the Jacobian matrices into two parts:
\begin{equation}
    \JJ(\uu) := [\JJ_1(\uu); \mathbf{0}]
    \quad
    \JJ'(\vv) := [\JJ'_1(\vv); \JJ'_2(\vv)]
\end{equation}
where $\JJ_1: \Ru \mapsto \R^{\dimu \times \dimu}$,
$\JJ'_1: \Ruv \mapsto \R^{\dimuv \times \dimuv}$,
$\JJ'_2: \Ruv \mapsto \R^{(\dime - \dimuv) \times \dimuv}$.
This allows to re-write the condition as
\begin{equation}
    \JJ_1(\uu) \JJ_1'(\vv)^\top = \tau \LL 
\end{equation}
which is no longer underconstrained. This equation is valid for any $\uu, \vv$ in the support of the marginal and the positive/negative distribution, which is why the left hand side cannot depend on $\uu$ and $\vv$, leaving the final form of the Jacobians:
\begin{equation}
    \JJ(\uu) := [\JJ_1; \mathbf{0}]
    \quad
    \JJ'(\vv) := [\JJ'_1; \aa(\vv)],
\end{equation}
where $\aa: \Ruv \mapsto \R^{(\dime-\dimuv) \times \dimuv}$ is an arbitrary function that ensures that Eq.~\ref{eq:minimizer-prop-6-condition} can hold.
It follows that $\hhx_1$ and $\hhy_1$ are affine transforms, $\hhx_2$ is a constant, and $\hhy_2$ will be a potentially non-linear transform to match the InfoNCE minimizer, concluding the proof.
\end{proof}
\end{prop}

We next discuss two special cases that are of interest to users of CEBRA and demonstrate that the previous result is more flexible than the results presented in \citet{Zimmermann2021ContrastiveLI}. Consider a setup where we have a constant ``offset'' between the latents vs. having a perfectly symmetric $p$. We are still able to recover the underlying latents for both vMF and Normal conditional distributions:

\begin{corol}
    The aforementioned result holds for vMF distributions with a constant offset between $\uu$ and $\vv$ on the hypersphere, parameterized by a rotation matrix $\QQ$, with $\log p(\vv | \uu) = \kappa \uu^\top \QQ \vv$. Assume $d = d' = k$ and assume the domain and co-domain of $\hh$, $\hh'$ is the unit sphere.
    \begin{proof}
        The distributions satisfies constancy of the second derivative, and the condition on the Jacobian matrices is
        \begin{align}
            \JJ_1 \JJ_1'^\top = (\tau \kappa) \QQ 
        \end{align}
        and since all vectors $\hh(\aa) = \JJ^\top \aa$ need to be normalized and $\QQ$ is orthogonal, we can consider any column $\qq_i$ and the corresponding vector $\aa_i$ of $\JJ_1$
        \begin{align}
            \JJ_1 \aa_i = (\tau \kappa) \qq_i
            \Rightarrow \| \JJ_1 \aa_i \|^2 =  (\tau \kappa)  \|\qq_i \|^2 
            \Rightarrow 1 =  (\tau \kappa)  1
        \end{align}
        and it follows $\tau = 1/\kappa$.
    \end{proof}
\end{corol}

\begin{corol}
    The aforementioned result holds for a Gaussian distribution with a constant offset between $\uu$ and $\vv$ parameterized, with $\log p(\vv | \uu) = \| \uu - \vv - \mu \|^2$. Assume $d = d' = k$.
    \begin{proof}
        We rewrite the log-conditional as 
        \begin{align}
            \log p(\vv | \uu) 
            &= - \| (\uu - \vv) - \bmu \|^2 \\
            &= - \| \uu - \vv \|^2 - \| \bmu \|^2 + 2 (\uu - \vv)^\top \bmu \\
            &= - \| \uu \|^2 - \| \vv \|^2 - \| \bmu \|^2 + 2 \uu^\top \bmu + 2 \vv^\top \bmu + 2 \uu^\top \vv \\
        \end{align}
        which gives, at the minimizer of the InfoNCE objective,
        \begin{equation}
            \hhx(\uu)^\top \hhy(\vv) / \tau
            = 2 \uu^\top \vv
                + (- \| \vv \|^2 + 2 \vv^\top \bmu)
                + ( - \| \uu \|^2 + 2 \uu^\top \bmu + C(\uu))
        \end{equation}
        and we recover
        \begin{align}
            \hhx_1(\uu) &= \JJ_1 \uu, \\
            \hhy_1(\vv) &= \JJ'_1 \vv, \\
            C(\uu) &= \| \uu \|^2 - 2 \uu^\top \bmu \\
            \hhy_2(\vv) &= - \| \vv \|^2 + 2\vv^\top \bmu.
        \end{align}
    \end{proof}
\end{corol}

While the above results extended the theory of \citet{Zimmermann2021ContrastiveLI} to training setups within the CEBRA library and used in the main text (training with dot-product and negative mean squared error as the similarity measure and sampling procedure), we can also leverage previous results from \citet{Hyvrinen2019NonlinearIU} to further extend the families of possible distributions.
 
Firstly, for conditional exponential family distributions within an ICA framework, the dot-product similarity can be used in conjunction with non-symmetric encoders $\ffx$, $\ffy$ (i.e., two separate networks) to recover the components up to a linear indeterminacy.
We recall Definition 1 from \citet{Hyvrinen2019NonlinearIU}:
\begin{customdef}{C}[Conditionally exponential distributions (Def. 1 from \citet{Hyvrinen2019NonlinearIU})]
    \label{monoexpdef}
    A random variable (independent component) $v_i$ is conditionally exponential of order $k$ given random vector $\xx$ if its conditional probability density function can be given in the form
    \begin{equation} \label{expdef}
        p(v_i|\xx)=\frac{Q_i(v_i)}{Z_i(\xx)}\exp\left[\sum_{j=1}^k \suffstat_{ij}(v_i)\lambda_{ij}(\xx)\right]
    \end{equation}
    almost everywhere in the support of $\xx$, with $\suffstat_{ij}$, $\lambda_{ij}$, $Q_i$, and $Z_i$  scalar-valued  functions. The sufficient statistics $\suffstat_{ij}$ are assumed linearly independent (over $j$, for each fixed $i$). 
\end{customdef}%

We observe a similar functional form as used in Prop.~\ref{prop:dot-product-identifiability}.
Values of 
$\suffstat_{ij}(v_i)$ would be represented by $\ffy$, and values in $\lambda_{ij}(\xx)$ would be represented by $\ffx$. As in the previous proposition, more dimensions in $\ffx$, $\ffy$ as latent variables are required ($\dime > \dimu$) for representing conditional distributions of the conditional exponential form.
In this setup, Theorem 3 from \citet{Hyvrinen2019NonlinearIU} implies in our context that the sufficient statistics of the latents can be recovered up to a linear transformation.

Secondly, for arbitrary conditional distributions, as long as variability conditions are satisfied within an ICA framework, contrastive learning can recover the underlying components up to a permutation and point-wise invertible transformation \citep{Hyvrinen2019NonlinearIU}. Using this mode applies when a similarity measure $\phi$ which gives 
\begin{equation}
    \psi(\uu, \vv) := \sum_{i = 1}^{\dime} \phi_i(\hy_i(\vv), \hhx(\uu)), \text{ or w.r.t. to the signal variables, }
    \psi(\xx, \yy)
    = \sum_{i = 1}^{\dime} \phi_i(\fy_i(\yy), \ffx(\xx)),
\end{equation}
is used within the CEBRA framework.
In this case, Theorem 1 in \citet{Hyvrinen2019NonlinearIU} applies and gives an identifiability guarantee of $\hhx$ to recover the ground truth latents $\vv$ up to permutations and component-wise non-linear transformations.

\clearpage
\newpage

\section*{Supplementary Tables}

\begin{table*}[h]
    \centering
     \caption{Consistency statistics related to Fig. 1. Data includes all rats (n=4).}
    \begin{tabular}{ c c c c }
    \toprule
     group 1 & group 2  & {P-value} & Reject \\ [0.5ex] 
    \midrule
    CEBRA-Behavior & CEBRA-Time & 0.0124 & \textbf{True} \\
    CEBRA-Behavior & conv-pi-VAE w/labels & 0.0128 & \textbf{True} \\
    CEBRA-Behavior & conv-pi-VAE without & 1.4e-10 & \textbf{True} \\
    CEBRA-Behavior & tSNE & 2.2e-05 & \textbf{True} \\
    CEBRA-Behavior & UMAP & 2.1e-14 & \textbf{True} \\
    CEBRA-Time & conv-pi-VAE w/labels & .99 & False \\
    CEBRA-Time & conv-pi-VAE without & 0.0001 & \textbf{True}\\
    CEBRA-Time & tSNE & 0.45 & False \\
    CEBRA-Time & UMAP & 1.02e-08 & \textbf{True}\\
    \bottomrule
    \end{tabular}
   
    \label{tab:consistency}
\end{table*}

\begin{table*}[h!]
    \centering
    \caption{Decoding Statistics related to Fig. 2. Data includes all rats (n=4); supervised grouping one way ANOVA F-55, p=4.7e-31; self- and unsupervised, one way ANOVA F=8, p=6.95e-05. Posthoc Tukey HSD Tests:}
    \begin{tabular}{ c c c c }
    \toprule
     group 1 & group 2  & {P-value} & Reject \\ [0.5ex] 
    \midrule
    CEBRA-Behavior & conv-pi-VAE (MC decoding) & 0.9 & False \\
    CEBRA-Behavior & conv-pi-VAE (kNN) & 0.001 & \textbf{True} \\
    CEBRA-Behavior & pi-VAE (MC decoding) & 0.001 & \textbf{True} \\
    CEBRA-Behavior & pi-VAE (kNN) & 0.001 & \textbf{True} \\
    \midrule
    CEBRA-Time & PCA & 0.0024 & \textbf{True} \\
    CEBRA-Time & tSNE & 0.0021 & \textbf{True} \\
    CEBRA-Time & UMAP & 0.0057 & \textbf{True} \\
    \bottomrule
    \end{tabular}
    \label{tab:decoding}
\end{table*}

\begin{table*}[h!]
\caption{\textbf{Related to Fig. 5. Posthoc Tukey HSD Test} Allen Neuropixels dataset, 10 Frame window (< 30 neurons=False):}
\vspace{-12pt}
\small
\begin{center}
\begin{tabular}{ c c c c c } 
\toprule
neuron no. & group 1 & group 2  & {P-value} & Reject \\ [0.5ex]
\midrule
30 & baseline-bayes & baseline-knn & 0.016  & \textbf{True}\\ 
30 & baseline-bayes & CEBRA & 0.1255 & False \\
30 & baseline-knn & CEBRA & 0.7083 & False \\
30 & baseline-bayes & CEBRA-joint &  0.0072 &  \textbf{True}\\
30 & baseline-knn & CEBRA-joint &  0.9784 &  False\\
30 & CEBRA & CEBRA-joint  & 0.4762  & False\\
\midrule
50 & baseline-bayes & baseline-knn & 0.0358 & \textbf{True}\\ 
50 & baseline-bayes & CEBRA & 0.324 & False \\
50 & baseline-knn & CEBRA & 0.5956 & False \\
50 & baseline-bayes & CEBRA-joint &  0.1296 &  False\\
50 & baseline-knn & CEBRA-joint &  0.8989 &  False\\
50 & CEBRA & CEBRA-joint  & 0.9379  & False\\
\midrule
100 & baseline-bayes & baseline-knn & 0.0 & \textbf{True}\\ 
100 & baseline-bayes & CEBRA & 0.2589 & False \\
100 & baseline-knn & CEBRA & 0.0002 & \textbf{True} \\
100 & baseline-bayes & CEBRA-joint &  0.372 &  False\\
100 & baseline-knn & CEBRA-joint &  0.0001 &  \textbf{True}\\
100 & CEBRA & CEBRA-joint  & 0.9941  & False\\
\midrule
200 & baseline-bayes & baseline-knn & 0.0 & \textbf{True}\\ 
200 & baseline-bayes & CEBRA & 0.9976 & False \\
200 & baseline-knn & CEBRA & 0.0 & \textbf{True} \\
200 & baseline-bayes & CEBRA-joint &  0.7999 &  False\\
200 & baseline-knn & CEBRA-joint &  0.0 &  \textbf{True}\\
200 & CEBRA & CEBRA-joint  & 0.6964  & False\\
\midrule
400 & baseline-bayes & baseline-knn & 0.0004 & \textbf{True}\\ 
400 & baseline-bayes & CEBRA & 0.2531 & False \\
400 & baseline-knn & CEBRA & 0.0 & \textbf{True} \\
400 & baseline-bayes & CEBRA-joint &  0.2166 &  False\\
400 & baseline-knn & CEBRA-joint &  0.0 &  \textbf{True}\\
400 & CEBRA & CEBRA-joint  & 0.9996  & False\\
\midrule
600 & baseline-bayes & baseline-knn & 0.0002 & \textbf{True}\\ 
600 & baseline-bayes & CEBRA & 0.2095 & False \\
600 & baseline-knn & CEBRA & 0.0 & \textbf{True} \\
600 & baseline-bayes & CEBRA-joint &  0.2884 &  False\\
600 & baseline-knn & CEBRA-joint &  0.0 &  \textbf{True}\\
600 & CEBRA & CEBRA-joint  & 0.9967  & False\\
\midrule
800 & baseline-bayes & baseline-knn & 0.0001 & \textbf{True}\\ 
800 & baseline-bayes & CEBRA & 0.3691 & False \\
800 & baseline-knn & CEBRA & 0.0 & \textbf{True} \\
800 & baseline-bayes & CEBRA-joint &  0.1668 &  False\\
800 & baseline-knn & CEBRA-joint &  0.0 &  \textbf{True}\\
800 & CEBRA & CEBRA-joint  & 0.9524  & False\\
\midrule
900 & baseline-bayes & baseline-knn & 0.0003 & \textbf{True}\\ 
900 & baseline-bayes & CEBRA & 0.3867 & False \\
900 & baseline-knn & CEBRA & 0.0 & \textbf{True} \\
900 & baseline-bayes & CEBRA-joint &  0.3522 &  False\\
900 & baseline-knn & CEBRA-joint &  0.0 &  \textbf{True}\\
900 & CEBRA & CEBRA-joint  & 0.9999  & False\\
\midrule
1000 & baseline-bayes & baseline-knn & 0.0018 & \textbf{True}\\ 
1000 & baseline-bayes & CEBRA & 0.4707 & False \\
1000 & baseline-knn & CEBRA & 0.0001 & \textbf{True} \\
1000 & baseline-bayes & CEBRA-joint &  0.3785 &  False\\
1000 & baseline-knn & CEBRA-joint &  0.0001 &  \textbf{True}\\
1000 & CEBRA & CEBRA-joint  & 0.9981  & False\\
\bottomrule
\end{tabular}
\label{table:dataSTATS_10frame}
\end{center}
\end{table*}

\begin{table*}[h!]
\caption{\textbf{Related to Fig. 5. Posthoc Tukey HSD Test} Allen Neuropixels dataset, 1 Frame window (below 50 neurons all False):}
\vspace{-12pt}
\small
\begin{center}
\begin{tabular}{c c c c c} 
\toprule
neuron no. & group 1 & group 2  & {P-value} & Reject \\ [0.5ex]
\midrule
50 & baseline-bayes & baseline-knn & 0.0233 & \textbf{True}\\ 
50 & baseline-bayes & CEBRA & 0.8918 & False \\
50 & baseline-knn & CEBRA & 0.0055 & True \\
50 & baseline-bayes & CEBRA-joint &  0.8238 & False\\
50 & baseline-knn & CEBRA-joint &  0.004 &  \textbf{True}\\
50 & CEBRA & CEBRA-joint  & 0.9987  & False\\
\midrule
100 & baseline-bayes & baseline-knn & 0.0 & \textbf{True}\\ 
100 & baseline-bayes & CEBRA & 0.0275 & \textbf{True} \\
100 & baseline-knn & CEBRA & 0.0132 & \textbf{True} \\
100 & baseline-bayes & CEBRA-joint &  0.9977 &  False\\
100 & baseline-knn & CEBRA-joint &  0.0 &  \textbf{True}\\
100 & CEBRA & CEBRA-joint  & 0.0395  & \textbf{True}\\
\midrule
200 & baseline-bayes & baseline-knn & 0.0005 & \textbf{True}\\ 
200 & baseline-bayes & CEBRA & 0.3703 & False \\
200 & baseline-knn & CEBRA & 0.0 & \textbf{True} \\
200 & baseline-bayes & CEBRA-joint &  0.0058 &  \textbf{True}\\
200 & baseline-knn & CEBRA-joint &  0.0 &  \textbf{True}\\
200 & CEBRA & CEBRA-joint  & 0.1478  & False\\
\midrule
400 & baseline-bayes & baseline-knn & 0.0044 & \textbf{True}\\ 
400 & baseline-bayes & CEBRA & 0.0336 & \textbf{True} \\
400 & baseline-knn & CEBRA & 0.0 & \textbf{True} \\
400 & baseline-bayes & CEBRA-joint &  0.0 &  \textbf{True}\\
400 & baseline-knn & CEBRA-joint &  0.0 &  \textbf{True}\\
400 & CEBRA & CEBRA-joint  & 0.0  & \textbf{True}\\
\midrule
600 & baseline-bayes & baseline-knn & 0.0125 & \textbf{True}\\ 
600 & baseline-bayes & CEBRA & 0.6063 & False \\
600 & baseline-knn & CEBRA & 0.001 & \textbf{True} \\
600 & baseline-bayes & CEBRA-joint &  0.0 &  \textbf{True}\\
600 & baseline-knn & CEBRA-joint &  0.0 &  \textbf{True}\\
600 & CEBRA & CEBRA-joint  & 0.0  & \textbf{True}\\
\midrule
800 & baseline-bayes & baseline-knn & 0.0006 & \textbf{True}\\ 
800 & baseline-bayes & CEBRA & 0.0008 & \textbf{True} \\
800 & baseline-knn & CEBRA & 0.0 & \textbf{True} \\
800 & baseline-bayes & CEBRA-joint &  0.0 &  \textbf{True}\\
800 & baseline-knn & CEBRA-joint &  0.0 &  \textbf{True}\\
800 & CEBRA & CEBRA-joint  & 0  & \textbf{True}\\
\midrule
900 & baseline-bayes & baseline-knn & 0.0004 & \textbf{True}\\ 
900 & baseline-bayes & CEBRA & 0.0048 & \textbf{True} \\
900 & baseline-knn & CEBRA & 0.0 & \textbf{True} \\
900 & baseline-bayes & CEBRA-joint &  0.0 &  \textbf{True}\\
900 & baseline-knn & CEBRA-joint &  0.0 &  \textbf{True}\\
900 & CEBRA & CEBRA-joint  & 0.0  & \textbf{True}\\
\midrule
1000 & baseline-bayes & baseline-knn & 0.0 & \textbf{True}\\ 
1000 & baseline-bayes & CEBRA & 0.0 & \textbf{True} \\
1000 & baseline-knn & CEBRA & 0.0 & \textbf{True} \\
1000 & baseline-bayes & CEBRA-joint &  0.0 &  \textbf{True}\\
1000 & baseline-knn & CEBRA-joint &  0.0 &  \textbf{True}\\
1000 & CEBRA & CEBRA-joint  & 0.0019  & \textbf{True}\\
\bottomrule
\end{tabular}
\label{table:dataSTATS_1frame}
\end{center}
\end{table*}

\begin{table*}[h!]
\caption{\textbf{Related to Fig. 5. Posthoc Tukey HSD Test} Allen Neuropixels dataset, scene classification with 1 Frame window:}
\vspace{-12pt}
\small
\begin{center}
\begin{tabular}{ c c c c c } 
\toprule
neuron no. & group 1 & group 2  & {P-value} & Reject \\ [0.5ex]
\midrule
50 & baseline-bayes & baseline-knn & 0.4575 & False\\ 
50 & baseline-bayes & CEBRA & 0.1986 & False \\
50 & baseline-knn & CEBRA & 0.9355 & False \\
50 & baseline-bayes & CEBRA-joint &  0.0 &  \textbf{True}\\
50 & baseline-knn & CEBRA-joint &  0.0 & \textbf{True}\\
50 & CEBRA & CEBRA-joint  & 0.0  & \textbf{True}\\
\midrule
100 & baseline-bayes & baseline-knn & 0.0207 & \textbf{True} \\ 
100 & baseline-bayes & CEBRA & 0.0084 & \textbf{True} \\
100 & baseline-knn & CEBRA & 0.9694 & False \\
100 & baseline-bayes & CEBRA-joint &  0.0 &  \textbf{True}\\
100 & baseline-knn & CEBRA-joint &  0.0 & \textbf{True}\\
100 & CEBRA & CEBRA-joint  & 0.0  & \textbf{True}\\
\midrule
200 & baseline-bayes & baseline-knn & 0.0106 & \textbf{True} \\ 
200 & baseline-bayes & CEBRA & 0.0001 & \textbf{True} \\
200 & baseline-knn & CEBRA & 0.1269 & False \\
200 & baseline-bayes & CEBRA-joint &  0.0 &  \textbf{True}\\
200 & baseline-knn & CEBRA-joint &  0.0 & \textbf{True}\\
200 & CEBRA & CEBRA-joint  & 0.0  & \textbf{True}\\
\midrule
400 & baseline-bayes & baseline-knn & 0.0047 & \textbf{True} \\ 
400 & baseline-bayes & CEBRA & 0.0001 & \textbf{True} \\
400 & baseline-knn & CEBRA & 0.239 & False \\
400 & baseline-bayes & CEBRA-joint &  0.0 &  \textbf{True}\\
400 & baseline-knn & CEBRA-joint &  0.0 & \textbf{True}\\
400 & CEBRA & CEBRA-joint  & 0.0  & \textbf{True}\\
\midrule
600 & baseline-bayes & baseline-knn & 0.0013 & \textbf{True} \\ 
600 & baseline-bayes & CEBRA & 0.0 & \textbf{True} \\
600 & baseline-knn & CEBRA & 0.0032 & \textbf{True} \\
600 & baseline-bayes & CEBRA-joint &  0.0 &  \textbf{True}\\
600 & baseline-knn & CEBRA-joint &  0.0 & \textbf{True}\\
600 & CEBRA & CEBRA-joint  & 0.0  & \textbf{True}\\
\midrule
800 & baseline-bayes & baseline-knn & 0.0 & \textbf{True} \\ 
800 & baseline-bayes & CEBRA & 0.0 & \textbf{True} \\
800 & baseline-knn & CEBRA & 0.0 & \textbf{True} \\
800 & baseline-bayes & CEBRA-joint &  0.0 &  \textbf{True}\\
800 & baseline-knn & CEBRA-joint &  0.0 & \textbf{True}\\
800 & CEBRA & CEBRA-joint  & 0.0  & \textbf{True}\\
\midrule
900 & baseline-bayes & baseline-knn & 0.0062 & \textbf{True} \\ 
900 & baseline-bayes & CEBRA & 0.0 & \textbf{True} \\
900 & baseline-knn & CEBRA & 0.0168 & \textbf{True} \\
900 & baseline-bayes & CEBRA-joint &  0.0 &  \textbf{True}\\
900 & baseline-knn & CEBRA-joint &  0.0 & \textbf{True}\\
900 & CEBRA & CEBRA-joint  & 0.0  & \textbf{True}\\
\midrule
1000 & baseline-bayes & baseline-knn & 0.0002 & \textbf{True} \\ 
1000 & baseline-bayes & CEBRA & 0.0 & \textbf{True} \\
1000 & baseline-knn & CEBRA & 0.0 & \textbf{True} \\
1000 & baseline-bayes & CEBRA-joint &  0.0 &  \textbf{True}\\
1000 & baseline-knn & CEBRA-joint &  0.0 & \textbf{True}\\
1000 & CEBRA & CEBRA-joint  & 0.0  & \textbf{True}\\
\bottomrule
\end{tabular}
\label{table:dataSTATS_SCENE}
\end{center}
\end{table*}

\begin{table*}[h!]
\caption{\textbf{Related to Fig. 5. Posthoc Tukey HSD Test} Allen Neuropixels dataset, Mean frame error, 10 frames}
\vspace{-12pt}
\small
\begin{center}
\begin{tabular}{ c c c c c } 
\toprule
neuron no. & group 1 & group 2  & {P-value} & Reject \\ [0.5ex]
\midrule
1000 & baseline-bayes & baseline-knn & 0.5277 & False \\ 
1000 & baseline-bayes & CEBRA & 0.0013 & \textbf{True} \\
1000 & baseline-knn & CEBRA & 0.0001 & \textbf{True} \\
1000 & baseline-bayes & CEBRA-joint &  0.0011 &  \textbf{True}\\
1000 & baseline-knn & CEBRA-joint &  0.0001 & \textbf{True}\\
1000 & CEBRA & CEBRA-joint  & 0.9996  & False \\
\bottomrule
\end{tabular}
\label{table:dataSTATS_FRAMEDIFF}
\end{center}
\end{table*}

\end{document}